%% file: arxiv.tex
\documentclass[10pt,twocolumn,letterpaper]{article}

\usepackage[pagenumbers]{cvpr} %

\input{preamble}

\input{math_commands}

\usepackage{graphicx}

\definecolor{cvprblue}{rgb}{0.21,0.49,0.74}
\usepackage[pagebackref,breaklinks,colorlinks,allcolors=cvprblue]{hyperref}

\AtEndPreamble{%
  \crefname{figure}{Fig.}{Figs.}%
  \Crefname{figure}{Figure}{Figures}%
  \crefname{equation}{Eq.}{Eqs.}%
  \Crefname{equation}{Eq.}{Eqs.}%
  \crefname{algorithm}{Alg.}{Algs.}%
  \Crefname{algorithm}{Alg.}{Algs.}%
  \crefname{appendix}{App.}{Apps.}%
  \Crefname{appendix}{App.}{Apps.}%
}

\input{macros}
\InputIfFileExists{local_config.tex}{}{} %

\def\ArxivVersion{}

\title{PreFLOP: Fast, Faithful Image and Video Previews via Spectral Downsampling}
\author{Asaf Livne \qquad Shimon Malnick \qquad Shai Avidan\\
Tel Aviv University}

\begin{document}
\twocolumn[{\renewcommand\twocolumn[1][]{#1}\maketitle\input{figures/teaser/teaser}}]

\input{sections/abstract}

\input{sections/introduction}

\input{sections/related}
\input{sections/preliminaries}

\input{sections/method}

\input{sections/experiments}

\input{sections/discussion}

\input{sections/conclusion}

{
    \small
    \bibliographystyle{ieeenat_fullname}
    \bibliography{references}
}

\input{sections/supplementary}

\end{document}

%% file: preamble.tex
\usepackage{cuted} %

\usepackage{currfile} %

\usepackage{caption} %

\usepackage{microtype}
\usepackage{adjustbox}
\usepackage{multirow}
\usepackage{makecell}
\usepackage{tikz}
\usetikzlibrary{arrows.meta,backgrounds,calc,fit, positioning, shapes.geometric, decorations.pathmorphing, decorations.pathreplacing, external, spy, shapes.symbols}
\usepackage{pgfplots}\pgfplotsset{compat=1.18}\usepgfplotslibrary{groupplots} %
\usepackage{algorithm}
\usepackage{algpseudocode}
\usepackage{colortbl}
\usepackage{pifont}
\usepackage[normalem]{ulem}   %
\usepackage{wrapfig}        %
\usepackage{enumitem}

%% file: math_commands.tex
\usepackage{amsmath,amsfonts,bm}

\def\eqref#1{equation~\ref{#1}}

\def\1{\bm{1}}

\DeclareMathAlphabet{\mathsfit}{\encodingdefault}{\sfdefault}{m}{sl}
\SetMathAlphabet{\mathsfit}{bold}{\encodingdefault}{\sfdefault}{bx}{n}

%% file: macros.tex
\newcommand{\ignorethis}[1]{}

\newcommand{\Reals      }     {{\textrm{I\kern-0.18em R}}}

\newcommand{\change     } [1] {\mbox{{\footnotesize $\Delta$} \kern-3pt}#1}

\newcommand{\dn}[1]{\downarrow #1} %

\newlength{\w}
\newlength{\h}

\definecolor{darkred}{rgb}{0.7,0.1,0.1}
\definecolor{darkgreen}{rgb}{0.1,0.6,0.1}
\definecolor{cyan}{rgb}{0.7,0.0,0.7}
\definecolor{otherblue}{rgb}{0.1,0.4,0.8}
\definecolor{maroon}{rgb}{0.76,.13,.28}
\definecolor{burntorange}{rgb}{0.81,.33,0}
\definecolor{cyan_real}{rgb}{0.0,1.0,1.0}
\definecolor{purple_real}{rgb}{0.5,0.0,0.5}
\definecolor{csig}{HTML}{5B7DB1}%
\definecolor{cnoise}{HTML}{C4C4C4}%

\ifdefined\ShowNotes
  \newcommand{\colornote}[3]{{\color{#1}\textbf{#2} #3\normalfont}}
\else
  \newcommand{\colornote}[3]{}
\fi

\newcommand\todosilent[1]{}

\newcommand{\probP}{\text{I\kern-0.15em P}}

\ifdefined\RevisedVersion

\else

\fi

\newcommand{\filmstriph}{4.5}%
\newcommand{\filmholew}{3.2pt}%
\newcommand{\filmholeh}{2.2pt}%
\newcommand{\filmnholes}{15}%
\newcommand{\filmstripcol}{black!85}%
\newcommand{\filmsprocket}[1]{%
  \edef\filmstripw{\the\dimexpr#1\relax}%
  \expandafter\filmsprocketat\expandafter{\filmstripw}}
\newcommand{\filmsprocketat}[1]{%
  \begin{tikzpicture}[x=#1,y=1pt,baseline=0pt]
    \fill[\filmstripcol] (0,0) rectangle (1,\filmstriph);
    \foreach \i in {1,...,\filmnholes}{%
      \node[fill=white,inner sep=0pt,rounded corners=0.35pt,
            minimum width=\filmholew,minimum height=\filmholeh]
        at ({(\i-0.5)/\filmnholes},{\filmstriph/2}) {};}%
  \end{tikzpicture}}

\newcommand{\gridprompt}[1]{{\usefont{T1}{ppl}{m}{it}\strut
  {\usefont{T1}{ptm}{m}{n}\textquotedblleft}#1{\usefont{T1}{ptm}{m}{n}\textquotedblright}}}

\providecommand{\zoominsetfrac}{0.42}%
\providecommand{\zoomcolor}{zoomcolor@default}%
\definecolor{zoomcolor@default}{HTML}{F2A900}%
\providecommand{\zoommarks}{}%
\definecolor{markcolor}{HTML}{D81B60}%
\tikzset{defectmark/.style={markcolor,line width=0.6pt}}%
\newcommand{\marktile}[2]{\ifx\zoommarks\empty\includegraphics[width=#1]{#2}\else
  \begin{tikzpicture}
    \node[anchor=south west,inner sep=0] (img) at (0,0) {\includegraphics[width=#1]{#2}};
    \begin{scope}[x={(img.south east)},y={(img.north west)}]\zoommarks\end{scope}
  \end{tikzpicture}\fi}%
\makeatletter
\newlength{\zt@tilew}
\newlength{\zt@roih}
\newlength{\zt@insetw}
\newlength{\zt@inseth}
\newlength{\zt@bigw}
\newlength{\zt@margin}

\newlength{\zoompopx}
\newlength{\zoompopy}
\newlength{\zt@mx}
\newlength{\zt@my}
\newif\ifzt@pop

\newcommand{\zoomtile}[7]{%
  \edef\zt@tilewd{\the\dimexpr#1\relax}\setlength{\zt@tilew}{\zt@tilewd}%
  \begin{tikzpicture}
    \node[anchor=south west,inner sep=0] (img) at (0,0) {\includegraphics[width=\zt@tilew]{#2}};
    \ifdim\dimexpr\zoompopx+\zoompopy\relax>0pt
      \zt@poptrue \setlength{\zt@mx}{-\zoompopx}\setlength{\zt@my}{-\zoompopy}%
      \useasboundingbox (img.south west) rectangle (img.north east);
    \else
      \zt@popfalse \setlength{\zt@mx}{\zt@margin}\setlength{\zt@my}{\zt@margin}%
    \fi
    \begin{scope}[x={(img.south east)},y={(img.north west)}]
      \coordinate (roi-nw) at (#3,{1-(#4)});
      \coordinate (roi-ne) at ({(#3)+(#5)},{1-(#4)});
      \coordinate (roi-sw) at (#3,{1-(#4)-(#6)});
      \coordinate (roi-se) at ({(#3)+(#5)},{1-(#4)-(#6)});
    \end{scope}
    \pgfpointdiff{\pgfpointanchor{roi-sw}{center}}{\pgfpointanchor{roi-nw}{center}}%
    \edef\zt@roihd{\the\pgf@y}\setlength{\zt@roih}{\zt@roihd}%
    \pgfmathsetmacro{\ztmag}{\zoominsetfrac/(#5)}%
    \setlength{\zt@insetw}{\zoominsetfrac\zt@tilew}%
    \setlength{\zt@inseth}{\ztmag\zt@roih}%
    \setlength{\zt@bigw}{\ztmag\zt@tilew}%
    \edef\zt@corner{#7}%
    \def\zt@NW{north west}\def\zt@NE{north east}%
    \def\zt@SW{south west}\def\zt@SE{south east}%
    \ifx\zt@corner\zt@NW
      \coordinate (cw-nw) at ($(img.north west)+(\zt@mx,-\zt@my)$);
      \coordinate (cw-se) at ($(cw-nw)+(\zt@insetw,-\zt@inseth)$);
      \def\zt@roinear{roi-ne}\def\zt@cwnear{cw-se}%
    \else\ifx\zt@corner\zt@NE
      \coordinate (cw-ne) at ($(img.north east)+(-\zt@mx,-\zt@my)$);
      \coordinate (cw-nw) at ($(cw-ne)+(-\zt@insetw,0pt)$);
      \coordinate (cw-se) at ($(cw-nw)+(\zt@insetw,-\zt@inseth)$);
      \def\zt@roinear{roi-nw}\def\zt@cwnear{cw-sw}%
    \else\ifx\zt@corner\zt@SW
      \coordinate (cw-sw) at ($(img.south west)+(\zt@mx,\zt@my)$);
      \coordinate (cw-nw) at ($(cw-sw)+(0pt,\zt@inseth)$);
      \coordinate (cw-se) at ($(cw-nw)+(\zt@insetw,-\zt@inseth)$);
      \def\zt@roinear{roi-se}\def\zt@cwnear{cw-ne}%
    \else %
      \coordinate (cw-se) at ($(img.south east)+(-\zt@mx,\zt@my)$);
      \coordinate (cw-nw) at ($(cw-se)+(-\zt@insetw,\zt@inseth)$);
      \def\zt@roinear{roi-sw}\def\zt@cwnear{cw-nw}%
    \fi\fi\fi%
    \coordinate (cw-sw) at (cw-nw |- cw-se);
    \coordinate (cw-ne) at (cw-se |- cw-nw);
    \coordinate (roi-near) at (\zt@roinear);
    \coordinate (cw-near) at (\zt@cwnear);
    \coordinate (zt-vec) at ($(roi-nw)-(img.south west)$);
    \coordinate (zt-p2) at ($(cw-nw)-\ztmag*(zt-vec)$);
    \ifzt@pop\fill[white] (cw-nw) rectangle (cw-se);\fi
    \begin{scope}
      \clip (cw-nw) rectangle (cw-se);
      \node[anchor=south west,inner sep=0] at (zt-p2) {\includegraphics[width=\zt@bigw]{#2}};
    \end{scope}
    \draw[\zoomcolor,line width=0.5pt,opacity=0.7] (roi-near) -- (cw-near);
    \draw[white,line width=1.2pt] (roi-nw) rectangle (roi-se);
    \draw[\zoomcolor,line width=0.6pt] (roi-nw) rectangle (roi-se);
    \ifzt@pop %
      \draw[white,line width=1.6pt] (cw-nw) rectangle (cw-se);
      \draw[\zoomcolor,line width=0.9pt] (cw-nw) rectangle (cw-se);
    \else
      \draw[white,line width=1.2pt] (cw-nw) rectangle (cw-se);
      \draw[\zoomcolor,line width=0.6pt] (cw-nw) rectangle (cw-se);
    \fi
    \begin{scope}[x={(img.south east)},y={(img.north west)}]\zoommarks\end{scope}
  \end{tikzpicture}%
}
\makeatother

\algnewcommand{\IfThen}[2]{\State \algorithmicif\ #1\ \algorithmicthen\ #2}

%% file: figures/teaser/teaser.tex
\begin{center}
  \begin{minipage}[c]{0.49\textwidth}
  \begingroup
  \definecolor{teaseraccent}{HTML}{DD8452}%
  \definecolor{teaserfull}{HTML}{4C72B0}%
  \definecolor{teaserperson}{HTML}{8172B3}%
  \definecolor{teaseraction}{HTML}{4C72B0}%
  \newcommand{\teaserfullw}{0.340}%
  \newcommand{\teaserxl}{0.195}%
  \newcommand{\teaserxr}{0.515}%
  \newcommand{\teaseryt}{0.150}%
  \newcommand{\teaseryb}{-0.150}%
  \newcommand{\teaserfulltime}{13.872}%
  \newcommand{\teaserpreviewtime}{2.752}%
  \newcommand{\teasercontinuetime}{12.48}%
  \newcommand{\teasertimeleft}{0.150}%
  \newcommand{\teasertimeright}{0.855}%
  \pgfmathsetmacro{\teaserprevieww}{\teaserfullw/2}
  \pgfmathsetmacro{\teasersec}{(\teasertimeright-\teasertimeleft)/56}%
  \tikzset{
    teaser small/.style={font=\scriptsize\sffamily,align=center,inner sep=0pt},
    teaser tag/.style={fill=#1,text=white,font=\tiny\sffamily\bfseries,inner sep=1.2pt,
      rounded corners=1pt,anchor=north east},
    teaser res/.style={fill=white,fill opacity=0.8,text opacity=1,text=black,
      font=\tiny\ttfamily,inner sep=1.2pt,anchor=south west},
    teaser step/.style={-{Stealth[length=4.5pt,width=4pt]},teaseraccent,line width=1.1pt,
      shorten <=1.5pt,shorten >=1.5pt},
    teaser say/.style={font=\scriptsize\itshape,text=teaserperson,align=center,fill=white,inner sep=1pt},
    teaser bar/.style={rounded corners=1.2pt},
    teaser clock/.pic={\draw[line width=0.55pt] (0,0) circle (2.5pt);
      \draw[line width=0.5pt,line cap=round] (0,0) -- (0,1.7pt) (0,0) -- (1.2pt,0);}}
  \newcommand{\teaseraction}[1]{{\upshape\tiny\sffamily\bfseries\color{teaseraction}$\to$ #1}}
  \newcommand{\teaserimage}[5]{%
    \node[inner sep=0pt,outer sep=0pt] (#1) at (#2) {\includegraphics[width=#3\linewidth]{figures/teaser/#4}};
    \draw[#5,line width=0.8pt] (#1.south west) rectangle (#1.north east);}
  \ifdefined\tikzexternaldisable\tikzexternaldisable\fi
  \begin{tikzpicture}[x=\linewidth,y=\linewidth]
    \teaserimage{preview1}{\teaserxl,\teaseryt}{\teaserprevieww}{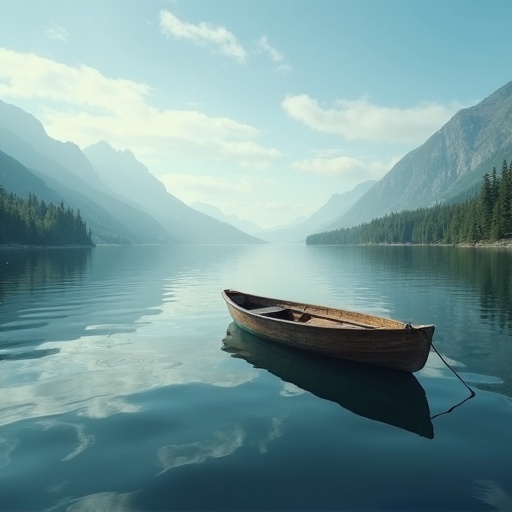}{teaseraccent}
    \teaserimage{preview2}{\teaserxl,\teaseryb}{\teaserprevieww}{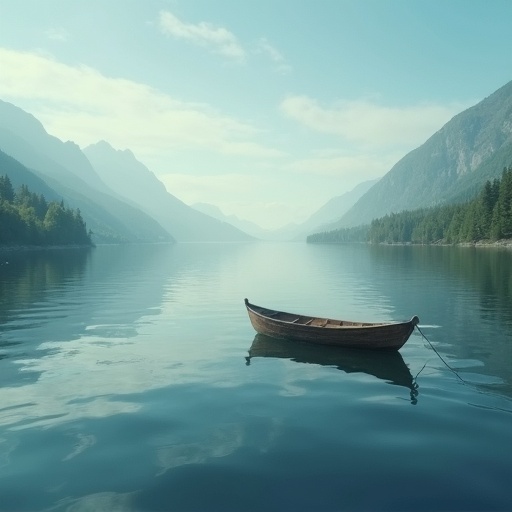}{teaseraccent}
    \teaserimage{preview3}{\teaserxr,\teaseryb}{\teaserprevieww}{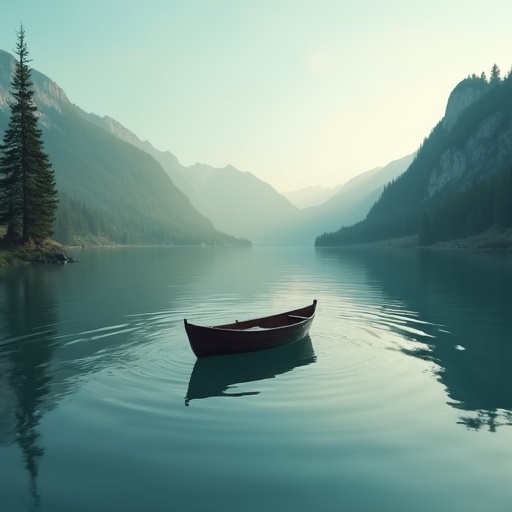}{teaseraccent}
    \teaserimage{preview4}{\teaserxr,\teaseryt}{\teaserprevieww}{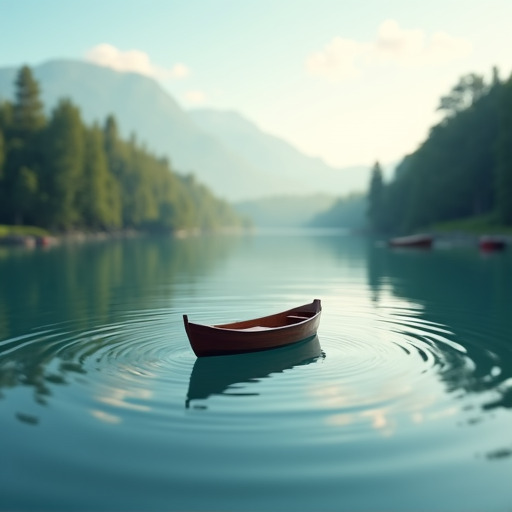}{teaseraccent}
    \foreach \i in {1,...,4}{%
      \node[teaser tag=teaseraccent] at ($(preview\i.north east)+(-0.006,-0.006)$) {\i};}
    \node[teaser res] at ($(preview1.south west)+(0.006,0.006)$) {512\,px};
    \draw[teaser step] (preview1.south) -- node[teaser say,fill=none,anchor=east,xshift=-3pt]
      {``the boat\\[-1pt]is too big''\\[2pt]\teaseraction{tiny}} (preview2.north);
    \draw[teaser step] (preview2.east) -- node[teaser say,fill=none,anchor=north,yshift=-3pt]
      {``another\\[-1pt]layout''\\[2pt]\teaseraction{seed}} (preview3.west);
    \draw[teaser step] (preview3.north) -- node[teaser say,fill=none,anchor=east,xshift=-3pt]
      {``too\\[-1pt]gloomy''\\[2pt]\teaseraction{CFG}} (preview4.south);
    \node[inner sep=0pt,anchor=south east] (full) at (1,{\teaseryb-\teaserprevieww/2})
      {\includegraphics[width=\teaserfullw\linewidth]{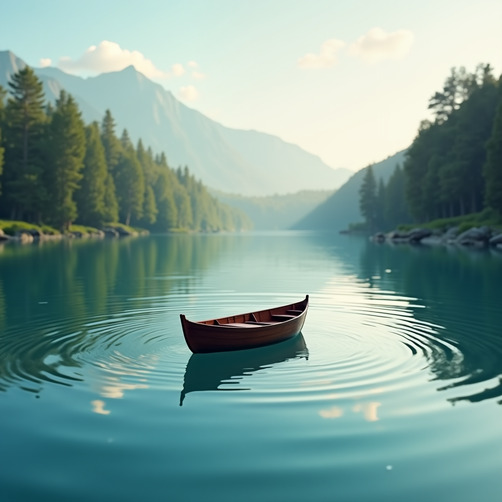}};
    \draw[teaserfull,line width=1.2pt] (full.south west) rectangle (full.north east);
    \node[teaser res] at ($(full.south west)+(0.008,0.008)$) {1024\,px};
    \draw[-{Stealth[length=5.5pt,width=5pt]},teaserfull,line width=1.4pt,rounded corners=4pt,
      shorten <=1.5pt,shorten >=1.5pt] (preview4.east) -| (full.north);
    \coordinate (teasercont) at ($(preview4.east)!0.5!(full.north |- preview4.east)$);
    \node[teaser say,fill=none,anchor=south] at ($(teasercont)+(0,0.006)$) {``that's it!''};
    \node[teaser say,fill=none,anchor=north] at ($(teasercont)+(0,-0.008)$) {\teaseraction{continue}};
    \def\yd{0.262}\def\xd{0.950}
    \tikzset{teaser d/.style={black!55,line width=0.6pt,dash pattern=on 2pt off 1.4pt}}
    \draw[teaser d,{Stealth[length=3.5pt,width=3pt]}-{Stealth[length=3.5pt,width=3pt]}]
      (preview4.north |- 0,0.237) -- (preview4.north |- 0,\yd) -- (\xd,\yd)
      -- (\xd,{\teaseryb-\teaserprevieww/2+\teaserfullw+0.003});
    \node[fill=white,font=\scriptsize,text=black!75,inner sep=1.2pt]
      at ({(\teaserxr+\xd)/2},\yd) {$d(\hat x_0, x_0)$};
    \begin{scope}[shift={(-0.008,0.105)}]
      \fill[teaserperson] (0.050,0.040) circle (0.021);
      \fill[teaserperson,rounded corners=2pt] (0.008,-0.040) -- (0.008,-0.015)
        .. controls (0.008,0.012) and (0.092,0.012) .. (0.092,-0.015) -- (0.092,-0.040) -- cycle;
      \foreach \k/\r in {1/0.005,2/0.007,3/0.009}{%
        \fill[teaserperson!35] ({0.050+0.006*\k},{0.058+0.022*\k}) circle (\r);}
    \end{scope}
    \node[draw=teaserperson!50,fill=teaserperson!8,rounded corners=2pt,inner sep=1.6pt,
      anchor=south west,font=\scriptsize] (prompt) at (0,0.255) {\gridprompt{a boat on a lake}};
    \node[font=\scriptsize\sffamily\bfseries,text=teaseraccent,anchor=center]
      at ($(prompt.center -| preview1.east)!0.5!(prompt.center -| preview4.west)$) {Preview loop};
    \def\ya{-0.317}\def\yf{-0.372}\def\yx{-0.409}\def\h{0.038}
    \node[teaser small,anchor=east,text=teaseraccent] at (\teasertimeleft-0.012,\ya) {Previews};
    \node[teaser small,anchor=east,text=teaserfull] at (\teasertimeleft-0.012,\yf) {Full};
    \foreach \i in {1,...,4}{%
      \fill[teaseraccent,teaser bar]
        ({\teasertimeleft+(\i-1)*\teaserpreviewtime*\teasersec+0.0015},\ya-\h/2)
        rectangle ++({\teaserpreviewtime*\teasersec-0.003},\h);
      \node[font=\tiny\sffamily\bfseries,text=white]
        at ({\teasertimeleft+(\i-0.5)*\teaserpreviewtime*\teasersec},\ya) {\i};
      \fill[teaserfull!35,teaser bar]
        ({\teasertimeleft+(\i-1)*\teaserfulltime*\teasersec+0.002},\yf-\h/2)
        rectangle ++({\teaserfulltime*\teasersec-0.004},\h);
      \node[teaser small,text=teaserfull]
        at ({\teasertimeleft+(\i-0.5)*\teaserfulltime*\teasersec},\yf) {\i};}
    \pgfmathsetmacro{\teaserstart}{\teasertimeleft+4*\teaserpreviewtime*\teasersec}
    \pgfmathsetmacro{\teaserend}{\teaserstart+\teasercontinuetime*\teasersec}
    \fill[teaserfull,teaser bar] ({\teaserstart+0.0015},\ya-\h/2) rectangle (\teaserend,\ya+\h/2);
    \node[teaser small,text=white] at ({(\teaserstart+\teaserend)/2},\ya) {4};
    \node[teaser small,anchor=west,text=teaseraccent,font=\scriptsize\sffamily\bfseries]
      (total) at (\teaserend+0.034,\ya) {23.5\,s};
    \pic[teaseraccent] at ($(total.west)+(-0.017,0)$) {teaser clock};
    \pgfmathsetmacro{\teaserend}{\teasertimeleft+4*\teaserfulltime*\teasersec}
    \node[teaser small,anchor=west,text=teaserfull,font=\scriptsize\sffamily\bfseries]
      (fulltotal) at (\teaserend+0.034,\yf) {55.5\,s};
    \pic[teaserfull] at ($(fulltotal.west)+(-0.017,0)$) {teaser clock};
    \draw[black!45] (\teasertimeleft,\yx) -- (\teasertimeright,\yx);
    \foreach \t in {0,10,...,50}{%
      \draw[black!45] ({\teasertimeleft+\t*\teasersec},\yx) -- ++(0,-0.008);
      \node[font=\tiny\sffamily,text=black!55,anchor=north]
        at ({\teasertimeleft+\t*\teasersec},\yx-0.009) {\t\,s};}
  \end{tikzpicture}\par
  \ifdefined\tikzexternalenable\tikzexternalenable\fi
  \endgroup
  \end{minipage}\hfill
  \begin{minipage}[c]{0.49\textwidth}
  \centering
  \definecolor{qptf}{HTML}{DD8452}\definecolor{qptr}{HTML}{55A868}\definecolor{qpsea}{HTML}{4C72B0}
  \definecolor{qpjte}{HTML}{8172B3}\definecolor{qpdss}{HTML}{C44E52}\definecolor{qpcln}{HTML}{CCB974}
  \definecolor{qpfew}{RGB}{74,74,74}
  \pgfplotsset{
    qp ours/.style={line width=1pt,mark size=1.5pt},
    qp base/.style={line width=1pt,mark size=1.5pt,dash pattern=on 3.6pt off 1.6pt,mark options={solid}},
    qp tf/.style={qp ours,qptf,mark=*},
    qp tr/.style={qp ours,qptr,mark=triangle*,mark size=1.9pt},
    qp sea/.style={qp base,qpsea,mark=diamond*,mark size=1.9pt},
    qp jte/.style={qp base,qpjte,mark=square*,mark size=1.3pt},
    qp dss/.style={qp base,qpdss,mark=diamond*,mark size=1.7pt},
    qp cln/.style={qp base,qpcln,mark=triangle*,mark size=1.9pt,every mark/.append style={rotate=180}},
    qp few/.style={qp base,qpfew,mark=+,mark size=2pt,mark options={solid,line width=1.1pt}}}
  \ifdefined\tikzexternaldisable\tikzexternaldisable\fi
  \begin{tikzpicture}
    \begin{groupplot}[group style={group size=2 by 1,horizontal sep=8pt,y descriptions at=edge left},
        width=0.385\linewidth,height=0.385\linewidth,scale only axis,
        xmin=2.4,xmax=11,ymin=0,ymax=0.75,y dir=reverse,
        xtick={3,5,8,10},xticklabels={3$\times$,5$\times$,8$\times$,10$\times$},
        ytick={0,0.25,0.5,0.75},yticklabels={0.00,0.25,0.50,0.75},
        ticklabel style={font=\scriptsize},tick style={black,line width=0.4pt},tick align=outside,
        major tick length=2pt,grid=major,grid style={black!10,line width=0.4pt},
        axis lines*=left,axis line style={line width=0.5pt,-},
        title style={font=\scriptsize,yshift=-3pt},clip=false,
        legend style={font=\tiny,draw=none,fill=none,legend columns=3,inner sep=0pt,
          /tikz/every even column/.append style={column sep=5pt}},
        legend image code/.code={\draw[mark repeat=2,mark phase=2] plot coordinates {(0cm,0cm) (0.22cm,0cm) (0.44cm,0cm)};},
        legend style={
          at={(1.043,-0.25)},anchor=north},
        legend cell align=left]
      \nextgroupplot[title={FLUX.1-dev},ylabel={Faithfulness (DreamSim) $\rightarrow$},
        ylabel style={font=\scriptsize}]
      \addplot[qp sea,forget plot] coordinates {(3.010,0.0495) (4.810,0.1420) (7.050,0.5640) (9.180,0.6689)};
      \addplot[qp jte,forget plot] coordinates {(3.210,0.1100) (5.100,0.2610) (8.080,0.5065)};
      \addplot[qp dss,forget plot] coordinates {(3.361,0.1303) (4.504,0.1703) (6.351,0.2175)};
      \addplot[qp cln,forget plot] coordinates {(2.941,0.1827) (4.809,0.2878) (7.046,0.4109) (9.182,0.5104)};
      \addplot[qp few,forget plot] coordinates {(2.941,0.1656) (4.809,0.2648) (7.046,0.3802) (9.182,0.5483)};
      \addplot[qp tr,forget plot] coordinates {(3.110,0.0458) (5.039,0.0783) (8.099,0.1161) (10.638,0.1277)};
      \addplot[qp tf,forget plot] coordinates {(3.110,0.0490) (5.040,0.0888) (8.100,0.1433) (10.640,0.1730)};
      \addlegendimage{qp tf}\addlegendentry{PreFLOP, training-free}
      \addlegendimage{qp jte}\addlegendentry{Jeong et al.\ + TaylorSeer}
      \addlegendimage{qp cln}\addlegendentry{Clean-output prediction}
      \addlegendimage{qp tr}\addlegendentry{PreFLOP, trained}
      \addlegendimage{qp dss}\addlegendentry{Downsampled seed}
      \addlegendimage{qp few}\addlegendentry{Fewer steps}
      \addlegendimage{qp sea}\addlegendentry{SeaCache}
      \nextgroupplot[title={SD3.5-Large}]
      \addplot[qp sea] coordinates {(2.950,0.0718) (4.840,0.1698) (7.130,0.4080) (9.320,0.7198)};
      \addplot[qp jte] coordinates {(3.090,0.1653) (4.900,0.2808) (7.050,0.5326)};
      \addplot[qp dss] coordinates {(3.604,0.2611) (4.992,0.3054) (7.316,0.3969)};
      \addplot[qp cln] coordinates {(2.952,0.1580) (4.843,0.3088) (7.125,0.4592) (9.321,0.5658)};
      \addplot[qp few] coordinates {(2.952,0.1590) (4.843,0.2422) (7.125,0.3449) (9.321,0.4879)};
      \addplot[qp tr] coordinates {(3.020,0.0470) (5.079,0.0928) (8.257,0.1394) (10.249,0.1633)};
      \addplot[qp tf] coordinates {(3.020,0.0686) (5.080,0.1352) (8.260,0.1974) (10.250,0.2296)};
    \end{groupplot}
    \node[font=\scriptsize,anchor=north] at ($(group c1r1.south east)!0.5!(group c2r1.south west)+(0,-12pt)$)
      {Speed-up $\rightarrow$};
  \end{tikzpicture}
  \ifdefined\tikzexternalenable\tikzexternalenable\fi
  \end{minipage}
  
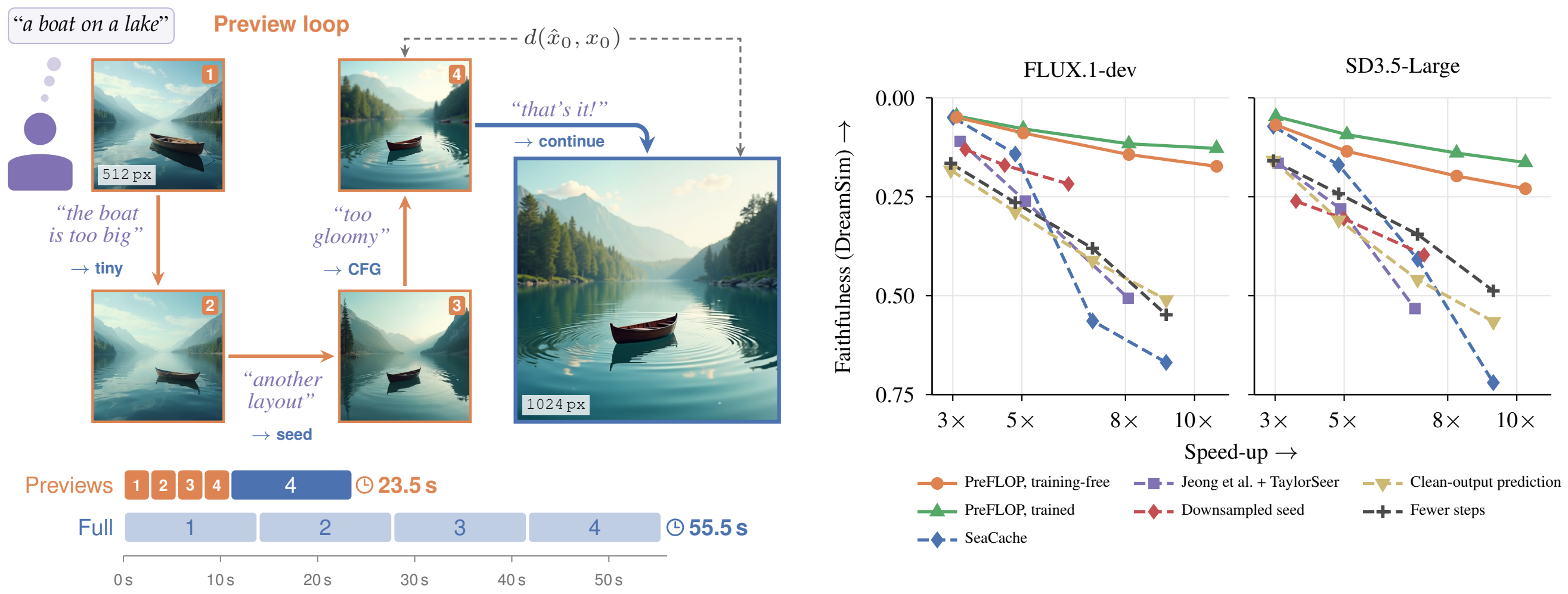
\captionof{figure}{\textbf{Previews shorten the creative loop.} Left: a user refines an idea through four low-resolution previews
  and continues only the chosen one, in less than half the generation time of four full runs. Right: our previews
  stay closer to the full image than the baselines as the speed-up grows.}
  \label{fig:teaser}
\end{center}

%% file: sections/abstract.tex
\begin{abstract}
Creating an image or video often means discovering, through successive iterations, how to express what one wants,
and what one wants in the first place. Users revise the prompt, seed, or settings in response to what they see, but
each revision requires another full generation to evaluate, making visual feedback expensive and the process slow.
Our goal is to provide this feedback through a preview: an image or video that closely resembles the full output at
a fraction of the generation cost. We present PreFLOP (Predictive Flow Previews), a training-free method that spectrally downsamples a
full-resolution intermediate latent and completes the preview at lower resolution. It retains the
low-frequency DCT coefficients and realigns the downsampled latent's noise level, while a schedule
search determines when to downsample and how to allocate the remaining steps across lower resolutions. Experiments
on FLUX.1-dev, SD3.5-Large, and Wan2.1 show that PreFLOP achieves higher fidelity to the full output than any
tested baseline, a result supported by a human evaluation. Seed rankings computed on previews better match those on
full outputs, under seven image scorers and six video measures.
Fine-tuning a low-resolution LoRA further improves performance. Finally, we formulate spectral consistency, a key property for
generation across resolutions, as three commutation conditions that can serve as training objectives.
\end{abstract}

%% file: sections/introduction.tex
\section{Introduction}
 \label{sec:intro}

Creating images and videos with text-conditioned flow models~\cite{esser2024sd3,bfl2024flux,wan2025wan} often
involves a cycle of generation, inspection and revision. Users respond to each output by changing the prompt, seed
or settings, whether to correct a failure, explore an alternative or refine what they
want~\cite{liu2022promptguidelines,xie2023promptlog}. This process can be lengthy: sessions in Midjourney and Stable Diffusion logs average 10--14
prompts~\cite{xie2023promptlog}. Each iteration typically requires a full generation, costing tens of seconds for
an image or tens of minutes for a video before the user can judge the result and decide what to change. Users would
therefore benefit from a preview: an image or video that closely resembles the full output
at a fraction of its cost. Such a preview would let them decide whether to proceed to the full run or make the next
change (\cref{fig:teaser}).

Several alternatives exist to generate previews.
Running fewer sampling steps or reusing computations through
caching~\cite{ma2024deepcache,liu2025teacache,chung2026seacache} produces a finished image or video more cheaply, but at
larger speed-ups both increasingly depart from the full output.
Alternatively, the model's
\emph{clean-output prediction} can be decoded partway through sampling, but early predictions are blurry and
incomplete.
Other preview methods align a few-step sampler with
the full output~\cite{wang2026consolver,jia2026faithfulpreviews}, but they require extensive training and have been
shown only on images. Closest to our work, Jeong et al.~\cite{jeong2026previews}
provide training-free low-resolution previews by subsampling an intermediate full-resolution latent, which aliases its high-frequency
noise and likely limits their fidelity.
Existing approaches thus lose fidelity, especially at larger speed-ups, or require extensive training.

To address this gap, we propose PreFLOP (Predictive Flow Previews), a training-free method whose finished previews
remain faithful to the full output at large speed-ups, for both images and videos. Sampling starts at full resolution,
where the first steps fix the composition of the full output. An intermediate state is then spectrally downsampled, as sampling theory prescribes~\cite{oppenheim2010dtsp}: its
noise-dominated, high-frequency band is discarded without aliasing into the low-frequency signal, and the remaining
steps run at lower resolution. Since downsampling attenuates
the noise but not the signal, we renormalize the state and relabel its time to match its new noise
level~\cite{xiao2026speed}.
A schedule search determines when to downsample, which resolutions to use and how to
allocate the remaining steps under a computational budget.

Across FLUX.1-dev, SD3.5-Large and Wan2.1, PreFLOP improves the fidelity-cost trade-off over the tested baselines at
larger speed-ups, and a fine-tuned low-resolution LoRA further improves fidelity. On average over seven image scorers
and six video measures, the previews also match the seed order of their full outputs better than any tested baseline. In a human evaluation, raters pick our previews as the
closest match to the full output, especially at larger speed-ups.
In summary, our contributions are:
\begin{itemize}[nosep,leftmargin=*]
  \item We introduce PreFLOP, which produces cheap, faithful previews of high-resolution outputs by downsampling an intermediate latent and
  completing sampling at lower resolution. It requires no training, while light fine-tuning further improves
  fidelity.
  \item We demonstrate improved trade-offs between fidelity and computational cost across image and video models
  at larger speed-ups, supported by a human evaluation. Moreover, reward models compare seeds more reliably on our previews than on any tested
  baseline.
  \item We formulate spectral consistency across resolutions as three commutation conditions, on the prior, the
  velocity field and the decoder, relate the limits of preview fidelity to them, and suggest how future models
  could be trained to satisfy them.
\end{itemize}

%% file: sections/related.tex
\section{Related Work}
\label{sec:related}

\input{figures/method_overview/method_overview}

Prior work lowers the cost of reaching a desired output in three main ways: it improves the quality of each sample,
reduces its cost, or estimates its outcome before sampling completes.

\paragraph{Sample quality.}
Some methods increase the likelihood that a single sample is the desired output. Prompt optimization rewrites the user's
prompt into one the model follows better~\cite{hao2023promptist}. Noise optimization improves the initial noise, by
optimizing it directly~\cite{guo2024initno}, learning to transform it~\cite{zhou2025goldennoise}, or reusing initial
noise that yields better images across prompts~\cite{xu2025goodseed}. Guidance methods~\cite{ho2022cfg,ahn2024pag,karras2024autoguidance,fan2025cfgzero} raise sample quality and
prompt adherence. These methods aim to improve the quality of each sample, whereas previews
reduce the cost of evaluating it.

\paragraph{Sampling cost.}
Acceleration methods make every generation cheaper while aiming to preserve its quality. Some shorten sampling by
distillation~\cite{song2023consistency,luo2023lcm,yin2024dmd}, reuse computations across neighboring steps, as in
SeaCache~\cite{chung2026seacache} and others~\cite{ma2024deepcache,liu2025teacache,malnick2026softcache}, or hand some
steps to a smaller model~\cite{pan2025tstitch,cheng2025srdiffusion,munir2026track}.
Others run part of the trajectory at a lower
resolution~\cite{xiao2026speed,jeong2025mixedres,tian2025bottleneck,zheng2026mrflow}.
These methods move toward full resolution during sampling, whereas our method moves in the opposite direction.
Since they produce the final
output, all these methods trade speed-up against its quality. Our method confines this trade-off to the preview,
allowing larger speed-ups than these methods can afford, while leaving the final output exactly as the model normally
generates it.

\paragraph{Early evaluation and previews.}
A sample can also be evaluated before sampling completes, either by trained readouts of its intermediate
state~\cite{zhao2026latsearch,cui2026diffusionprobe,kwon2026earlyfailure} or by previewing its final output. The standard
preview is the model's clean-output prediction, which search methods score to select or steer
candidates~\cite{guimaraes2026psp,rawal2026flashbon,subramaniam2026previewdiff}, but it is blurry early in sampling.
Wang et al.~\cite{wang2026consolver} and Jia et al.~\cite{jia2026faithfulpreviews} instead obtain previews by training
corrections to a few-step sampler at full resolution. As in our method, Jeong et
al.~\cite{jeong2026previews} preview at a lower resolution without training, by subsampling an intermediate full-resolution latent and steering the
remaining steps toward stored velocities.

%% file: figures/method_overview/method_overview.tex
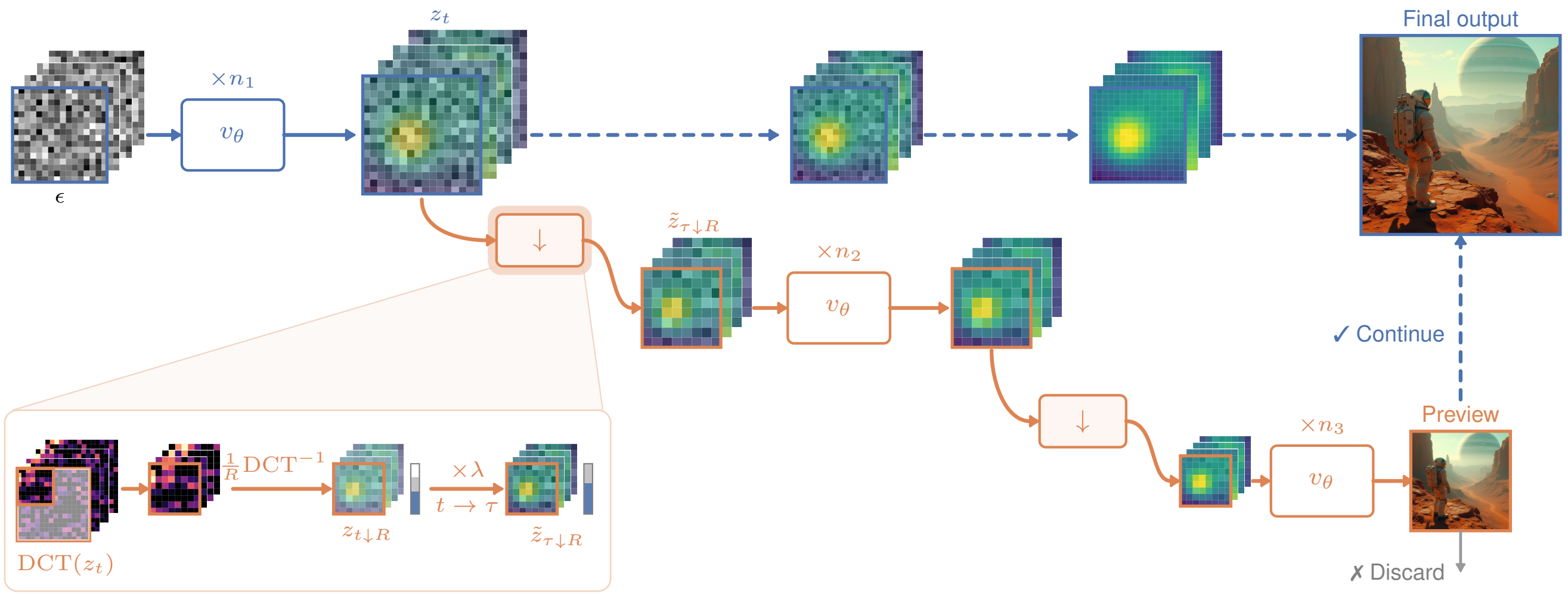
\begin{figure*}[t]
  \centering
  \begingroup
  \definecolor{cfull}{HTML}{4C72B0}%
  \definecolor{clow}{HTML}{DD8452}%
  \tikzset{
    c label/.style={font=\footnotesize\sffamily,align=center,inner sep=0pt,outer sep=0pt},
    c small/.style={c label,font=\scriptsize\sffamily},
    c arrow/.style={line width=1.2pt,line cap=round,line join=round,
      -{Latex[round,length=4pt,width=5pt]}}}
  \newcommand{\ctile}[5]{%
    \node[inner sep=0pt,outer sep=0pt] (#1) at (#2) {\includegraphics[width=#3\linewidth,height=#3\linewidth]{#5}};
    \draw[#4,line width=1pt] (#1.south west) rectangle (#1.north east);}
  \newcommand{\bdir}{figures/method_overview/assets}
  \newcommand{\bd}{0.13}%
  \newcommand{\bstack}[5]{%
    \foreach \bi in {3,2,1,0}{%
      \node[inner sep=0pt,draw=white,line width=0.5pt] at ({#1+\bi*\bd*#3},{#2+\bi*\bd*#3})
        {\includegraphics[width=#3\linewidth]{\bdir/#4_c\bi.png}};}
    \draw[#5] ({#1-#3/2},{#2-#3/2}) rectangle ({#1+#3/2},{#2+#3/2});}
  \newcommand{\bmix}[5]{%
    \fill[csig] (#1,#2) rectangle ({#1+0.0055},{#2+#4*#3});
    \fill[cnoise] (#1,{#2+#4*#3}) rectangle ({#1+0.0055},{#2+(#4+#5)*#3});
    \draw[black!50,line width=0.4pt] (#1,#2) rectangle ({#1+0.0055},{#2+#3});}
  \tikzset{b frame full/.style={cfull,line width=1pt},b frame low/.style={clow,line width=1pt},
    b frame plain/.style={black!40,line width=0.6pt}}
  \ifdefined\tikzexternaldisable\tikzexternaldisable\fi
  \begin{tikzpicture}[x=\linewidth,y=\linewidth]
    \tikzset{
      c model/.style={c label,draw,rounded corners=2pt,minimum width=0.065\linewidth,
        minimum height=0.045\linewidth,fill=white,line width=0.8pt},
      c handover/.style={c label,draw=clow,text=clow,fill=clow!7,rounded corners=2pt,
        minimum width=0.055\linewidth,minimum height=0.033\linewidth,line width=0.8pt}}
    \bstack{0.040}{0.290}{0.060}{eps}{b frame full}
    \node[c small,anchor=north] at (0.040,0.253) {$\epsilon$};
    \node[c model,draw=cfull,text=cfull] (model1) at (0.150,0.290) {$v_\theta$};
    \node[c small,text=cfull,anchor=south] at (0.150,0.320) {$\times n_1$};
    \draw[c arrow,cfull] (0.096,0.290) -- (model1.west);
    \draw[c arrow,cfull] (model1.east) -- (0.228,0.290);
    \bstack{0.270}{0.290}{0.075}{zt}{b frame full}
    \node[c small,anchor=south,text=cfull] at (0.282,0.361) {$z_t$};

    \draw[c arrow,cfull,dashed] (0.340,0.290) -- (0.497,0.290);
    \bstack{0.535}{0.290}{0.060}{full0}{b frame full}
    \draw[c arrow,cfull,dashed] (0.590,0.290) -- (0.687,0.290);
    \bstack{0.725}{0.290}{0.060}{full2}{b frame full}
    \draw[c arrow,cfull,dashed] (0.780,0.290) -- (0.864,0.290);
    \ctile{full}{0.930,0.290}{0.1265}{cfull}{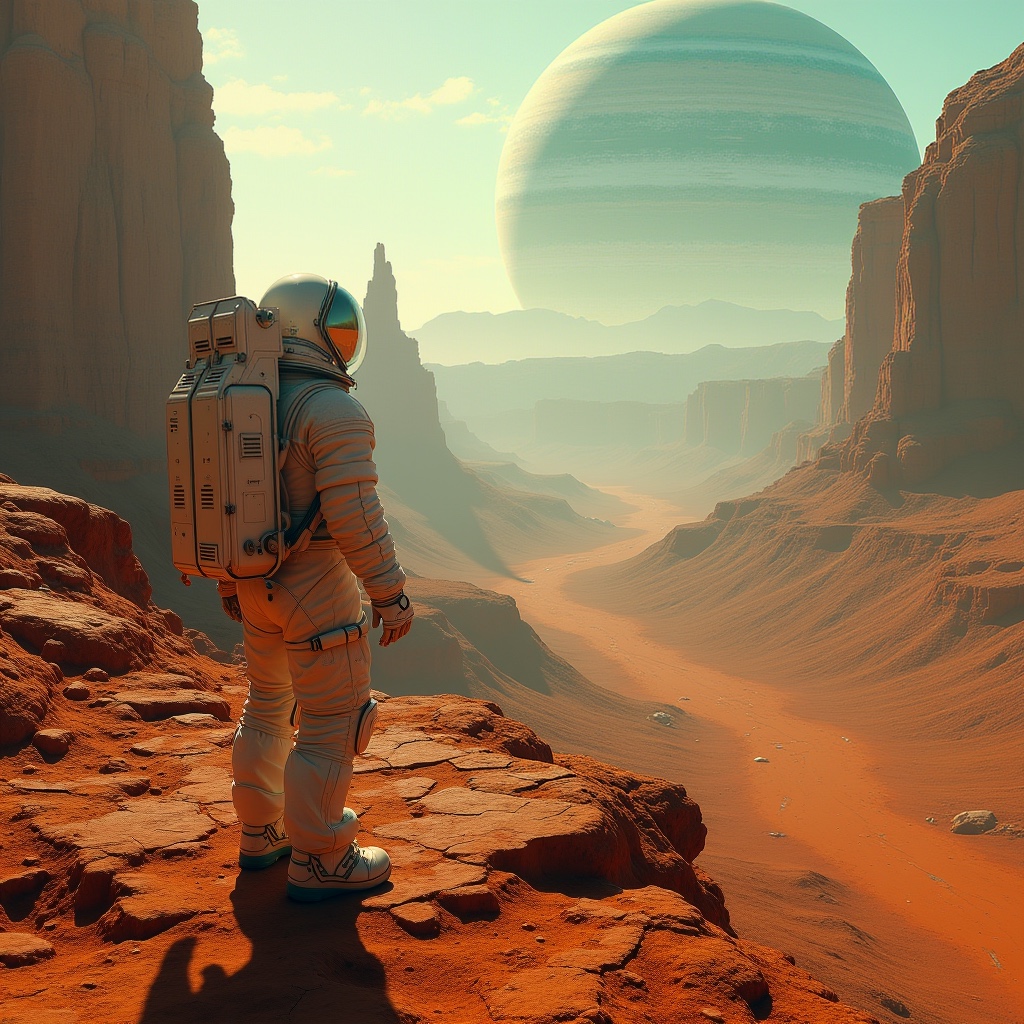}
    \node[c small,anchor=south,text=cfull] at ([yshift=1.5pt]full.north) {Final output};

    \node[c handover] (handover1) at (0.345,0.223) {$\downarrow$};
    \draw[c arrow,clow] (0.270,0.249) to[out=-90,in=180] (handover1.west);
    \draw[c arrow,clow] (handover1.east) to[out=0,in=180] (0.410,0.180);
    \bstack{0.435}{0.180}{0.050}{aligned}{b frame low}
    \node[c small,anchor=south,text=clow] at (0.443,0.227) {$\tilde z_{\tau\dn{R}}$};
    \node[c model,draw=clow,text=clow] (model2) at (0.535,0.180) {$v_\theta$};
    \node[c small,text=clow,anchor=south] at (0.535,0.210) {$\times n_2$};
    \draw[c arrow,clow] (0.481,0.180) -- (model2.west);
    \draw[c arrow,clow] (model2.east) -- (0.605,0.180);
    \bstack{0.632}{0.180}{0.050}{small0}{b frame low}

    \node[c handover] (handover2) at (0.690,0.108) {$\downarrow$};
    \draw[c arrow,clow] (0.632,0.153) to[out=-90,in=180] (handover2.west);
    \draw[c arrow,clow] (handover2.east) to[out=0,in=180] (0.752,0.070);
    \bstack{0.768}{0.070}{0.032}{aligned2}{b frame low}
    \node[c model,draw=clow,text=clow] (model3) at (0.842,0.070) {$v_\theta$};
    \node[c small,text=clow,anchor=south] at (0.842,0.100) {$\times n_3$};
    \draw[c arrow,clow] (0.798,0.070) -- (model3.west);
    \draw[c arrow,clow] (model3.east) -- (0.900,0.070);
    \ctile{prev}{0.930,0.070}{0.06325}{clow}{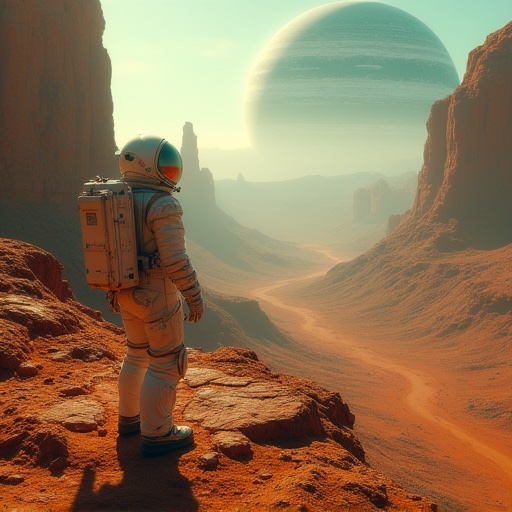}
    \node[c small,anchor=south,text=clow] (prevlab) at ([yshift=1.5pt]prev.north) {\vphantom{p}Preview};
    \draw[c arrow,cfull,dashed] ([yshift=2pt]prevlab.north) -- (full.south);
    \node[c small,anchor=east,text=cfull] at (0.920,0.164) {\ding{51} Continue};
    \draw[black!40,line width=0.9pt,line cap=round,-{Latex[round,length=3.5pt,width=4pt]}]
      (prev.south) -- (0.930,0.012);
    \node[c small,anchor=east,text=black!50] at (0.920,0.012) {\ding{55} Discard};

    \begin{pgfonlayer}{background}
      \fill[clow,opacity=0.05] (handover1.south west) -- (0.010,0.115) -- (0.385,0.115)
        -- (handover1.south east) -- cycle;
      \draw[clow!30,line width=2.5pt,rounded corners=3pt]
        ([shift={(-1.5pt,-1.5pt)}]handover1.south west) rectangle ([shift={(1.5pt,1.5pt)}]handover1.north east);
    \end{pgfonlayer}
    \draw[clow!35,line width=0.4pt] (handover1.south west) -- (0.010,0.115)
      (handover1.south east) -- (0.385,0.115);
    \draw[clow!35,fill=white,rounded corners=4pt,line width=0.5pt]
      (0.005,0.000) rectangle (0.390,0.115);

    \bstack{0.036}{0.055}{0.046}{spec}{b frame low}
    \fill[white,opacity=0.55,even odd rule] (0.013,0.032) rectangle (0.059,0.078)
      (0.013,0.078) rectangle (0.036,0.055);
    \draw[clow,line width=1pt] (0.013,0.055) rectangle (0.036,0.078);
    \draw[c arrow,clow] (0.080,0.065) -- (0.097,0.065);
    \bstack{0.113}{0.065}{0.032}{speccut}{b frame low}
    \node[c small,anchor=north,text=clow] at (0.044,0.024) {$\mathrm{DCT}(z_t)$};
    \draw[c arrow,clow] (0.150,0.065) -- (0.213,0.065)
      node[midway,above=3pt,xshift=-3pt,c small] {$\frac{1}{R}\mathrm{DCT}^{-1}$};
    \bstack{0.230}{0.065}{0.032}{zlr}{b frame low}
    \bmix{0.263}{0.049}{0.032}{0.45}{0.275}
    \node[c small,anchor=north,text=clow] at (0.235,0.041) {$z_{t\dn{R}}$};
    \draw[c arrow,clow] (0.276,0.065) -- (0.322,0.065)
      node[midway,above=3pt,c small] {$\times\lambda$}
      node[midway,below=3pt,c small] {$t\to\tau$};
    \bstack{0.340}{0.065}{0.032}{aligned}{b frame low}
    \bmix{0.373}{0.049}{0.032}{0.621}{0.379}
    \node[c small,anchor=north,text=clow] at (0.356,0.041) {$\tilde z_{\tau\dn{R}}$};

  \end{tikzpicture}\par
  \ifdefined\tikzexternalenable\tikzexternalenable\fi
  \endgroup
  \caption{\textbf{Previews from the model's own trajectory.} A generation runs
  $n_1$ steps at full resolution and saves $z_t$. Each downsampling $\downarrow$ keeps the
  low-frequency DCT block, renormalizes with $\lambda$ and relabels the time to $\tau$
  (\crefrange{eq:lambda}{eq:handover}). The model then finishes the preview in $n_2$ and $n_3$ steps on smaller
  grids. A continued seed resumes from $z_t$ (dashed). Bars show the {\color{csig}signal} and
  {\color{cnoise!70!black}noise} weights.}
  \label{fig:method}
\end{figure*}

%% file: sections/preliminaries.tex
\section{Preliminaries}
\label{sec:prelim}

\paragraph{Flow matching.}
Flow matching~\cite{lipman2023flowmatching,liu2023rectifiedflow} learns a velocity field from linear interpolations
between noise and data. Let $z_0$ be a clean latent, the VAE encoding of a data sample, and $\epsilon$ a noise
sample of the same shape, drawn from the prior $p_1=\mathcal N(0,I)$. The path between them, parameterized by $t$, is given by
\begin{equation}
  z_t=(1-t)\,z_0+t\,\epsilon,\qquad t\in[0,1].
  \label{eq:path}
\end{equation}
A network $v_\theta(z_t,t,c)$ is trained to predict the velocity $\dot z_t=\mathrm{d}z_t/\mathrm{d}t=\epsilon-z_0$ of
this path for a prompt $c$; we denote its prediction as $v_t := v_\theta(z_t,t,c)$. At inference, sampling starts from
pure noise, $z_1=\epsilon$, and an ODE solver integrates the learned velocity $v_\theta$ backwards in time from $t=1$ to
$t=0$ in $N$ steps; the endpoint $z_0$ is the generated latent.
Solving \cref{eq:path} for $z_0$ with the predicted
velocity gives the model's current estimate of the clean latent, $\hat z_0=z_t-t\,v_t$, which converges
to $z_0$ as sampling progresses. Decoding it with the VAE gives the clean-output prediction.

\paragraph{Ideal downsampling.}
Sampling theory gives the ideal way to downsample a signal by a factor $R$ per side: remove the frequencies that the
smaller grid cannot represent with an ideal low-pass filter, then resample~\cite{oppenheim2010dtsp}. Dropping
samples without the filter lets those frequencies alias into the retained band. The orthonormal discrete cosine
transform (DCT)~\cite{ahmed1974dct} does both in one step: keeping the lowest $H/R\times W/R$ block of
coefficients of an $H\times W$ signal and inverting it on the smaller grid is ideal low-pass filtering followed by
resampling, exact for any $R$ with $H/R$ and $W/R$ integers. For a signal $x\in\mathbb R^{H\times W}$, we write this
downsampling as
\begin{equation}
  x_{\dn{R}}=D_R(x)=\frac{1}{R}\,\mathrm{DCT}^{-1}\bigl(\mathrm{DCT}(x)\bigl[{:}\tfrac{H}{R},\,{:}\tfrac{W}{R}\bigr]\bigr),
  \label{eq:cut}
\end{equation}
where the slice keeps the lowest $H/R\times W/R$ block of coefficients, the DCT acts on each channel, and the factor $1/R$ preserves the RMS amplitude of retained content on the
smaller grid, since the orthonormal transform preserves total energy.

%% file: sections/method.tex
\section{Method}
\label{sec:method}

Our pipeline is shown in \cref{fig:method}, where a generation runs its first $n_1$ steps at full resolution and
saves its state.
The state is then downsampled, and the preview is finished at one or more lower resolutions by the same model or a
smaller one. A continued generation resumes from the saved state.
We first motivate this design (\cref{sec:motivation}), then describe the downsampling and the relabeling of the
state (\cref{sec:downsampling}), the allocation of steps across scales (\cref{sec:schedules}), and fine-tuning
that brings the previews closer to the full output (\cref{sec:training}).

\subsection{Motivation}
\label{sec:motivation}

The composition and layout of a generated image are largely set by the low-frequency band of its latent, from the initial
noise onward~\cite{wang2023painters,wang2025seeds,Z_Cohen_2026}. \Cref{fig:motivation} illustrates this for the initial noise: images
generated from noise samples which share their low-frequency band are much more similar to each other than images from
samples that share their high-frequency band. Downsampling, which keeps the low band and discards the high one, should
therefore preserve the composition and layout. Ideally, running the model at a lower resolution from the downsampled noise would
produce the downsampled image. In practice it does not, because the model lacks \textbf{spectral consistency}: it does not treat a
given frequency band the same way at every resolution, a property we formalize in \cref{sec:discussion}. We therefore
run the first few steps at full resolution, which mitigates the resulting drift: the low frequencies of the state are
formed by the model at the resolution of the final image, and the downsampled state inherits them. The number of
full-resolution steps is thus a trade-off between preview fidelity and speed-up: each step at full resolution brings
the preview closer to the final image, but costs as much as in the full generation.

\input{figures/motivation/motivation}

\subsection{Downsampling}
\label{sec:downsampling}

Downsampling should return a state on the path of \cref{eq:path}, since the model is trained on such states. Let
$R>1$ divide both $H$ and $W$. The operator $D_R$ is linear, and therefore it downsamples the clean latent and the noise separately,
$D_R(z_t)=(1-t)\,D_R(z_0)+t\,D_R(\epsilon)$. Assuming the clean latent lies mostly in the retained band, which $D_R$
preserves exactly, it keeps its amplitude, and we define $z_{0\dn{R}}=D_R(z_0)$. White noise does not: its retained coefficients are standard Gaussians, so
$D_R(\epsilon)=\epsilon_{\dn{R}}/R$ with $\epsilon_{\dn{R}}\sim\mathcal N(0,I)$. Hence
\begin{equation}
  z_{t\dn{R}}=D_R(z_t)=(1-t)\,z_{0\dn{R}}+\frac{t}{R}\,\epsilon_{\dn{R}},
\end{equation}
whose weights do not sum to one for $t>0$, so the state is off the path of \cref{eq:path}. Following the timestep
alignment of SPEED~\cite{xiao2026speed}, we renormalize the state with $\lambda$ and relabel its time $\tau$:
\begin{align}
  \lambda(t,R)&=\frac{1}{(1-t)+t/R}\qquad \tau(t,R)=\frac{t}{R}\,\lambda(t,R) \label{eq:lambda}\\[3pt]
  \tilde z_{\tau\dn{R}}&=\lambda(t,R)\,z_{t\dn{R}}=(1-\tau)\,z_{0\dn{R}}+\tau\,\epsilon_{\dn{R}} \label{eq:handover}
\end{align}

$\tilde z_{\tau\dn{R}}$ now has the weights of \cref{eq:path} at $\tau$, with the downscaled clean latent and
noise from the same seed, and $\tau$ equals Esser et al.'s timestep shift~\cite{esser2024sd3} for a resolution ratio of $1/R$. Sampling continues from $(\tilde z_{\tau\dn{R}},\tau)$ at lower resolution.
Successive downsampling steps compose, allowing previews at several resolutions. Video latents are downsampled in the same way, in
space and frame by frame, with the same $\lambda(t,R)$ and $\tau(t,R)$.

\subsection{Schedule search}
\label{sec:schedules}

We define a schedule $\mathbf n=(n_1,\dots,n_S)$ by the number of steps the model takes at each of the
predetermined scales $1=s_1>\dots>s_S$. The first $n_1$ steps are the first $n_1$ steps of the model's native schedule
at full resolution, so the preview coincides with the full generation until the first downsampling. After each
downsampling, the steps left are uniformly respaced from $\tau$ to 0 as in the model's native schedule (see the
supplementary material).
Let $T(\mathbf n)$ be the runtime of a schedule, and $T_{\mathrm{full}}$ that of the full generation.

For a target speed-up $\rho$, we search on a calibration set of 32 images or 16 videos for
\begin{equation}
  \min_{\mathbf n:\ T(\mathbf n)\le T_{\mathrm{full}}/\rho}\ \mathbb E\bigl[d(\hat x_0(\mathbf n),x_0)\bigr],
  \label{eq:score}
\end{equation}
where $d(\cdot, \cdot)$ is a distance between the preview $\hat x_0(\mathbf n)$, the decoded output of schedule $\mathbf n$,
and the full output $x_0$ of the same prompt and noise.
We use DreamSim~\cite{fu2023dreamsim} for images and $-\mathrm{VJ\text{-}patch}$ for video, where VJ-patch is the
cosine similarity of corresponding V-JEPA~2 features~\cite{assran2025vjepa2} of the two videos.
We solve \cref{eq:score} for each model and $\rho$ by Bayesian optimization with a random-forest surrogate, following
SMAC3~\cite{lindauer2022smac3}.

\input{tables/overhead/overhead}

\subsection{Fine-tuning}
\label{sec:training}

The previews can be further improved by fine-tuning on the full outputs, with no use of external data. We fine-tune
the model with a LoRA~\cite{hu2022lora} update $\Delta\theta$ that is active only below full resolution, so the full-resolution steps remain
unchanged. We use a path loss for images and a commutator loss for video.

\paragraph{Path loss.}
For images, we use a learned downsampling: a map
\begin{equation}
  H_\psi(z_t)=\tilde z_{\tau\dn{R}}+r_\psi(z_t,\hat z_0,t)
  \label{eq:map}
\end{equation}
replaces \cref{eq:handover} at each change of scale, where $r_\psi$ is a small residual network that also sees the model's last
clean-latent prediction $\hat z_0$. Its last layer is zero-initialized, so $H_\psi$ starts as the downsampling of
\cref{eq:handover}.
We then run a preview schedule from the saved full-resolution state to its end and compute the loss between the decoded
preview $\hat x_0$ and the reference $x_0$ of \cref{sec:schedules},
\begin{equation}
  \mathcal L_{\mathrm{path}}(\psi,\Delta\theta)=\mathbb E\bigl[\mathrm{LPIPS}(\hat x_0,x_0)+\beta\|\hat x_0-x_0\|_1\bigr],
  \label{eq:loss-path}
\end{equation}
$\beta=0.1$, with VGG-based LPIPS~\cite{zhang2018lpips}. The gradient flows through every step below full resolution and the
decoder, and updates both the LoRA and the map. Each training example draws one of the searched schedules at random,
so one map and one LoRA serve every speed-up.

\paragraph{Commutator loss.}
Jeong et al.~\cite{jeong2026previews} measure the commutator between the velocity predicted at the downsampled state,
$v_\theta(z_{t\dn{}},t,c)$, and the derivative of the downsampled state with respect to $t$, which is
$\dot z_{t\dn{}}=D\bigl(v_\theta(z_t,t,c)\bigr)$ under their downsampling operator $D$. Our downsampling changes both the state and its time coordinate, so the corresponding target is the derivative of the aligned state with respect to $\tau$,
$\dot{\tilde z}_{\tau\dn{R}}=\mathrm d\tilde z_{\tau\dn{R}}/\mathrm d\tau$ (derived in the supplementary material):
\begin{equation}
  \dot{\tilde z}_{\tau\dn{R}}=(R-1)\,D_R(z_t)+\bigl[R-(R-1)\,t\bigr]\,D_R(v_t)
  \label{eq:vstar}
\end{equation}
Our generalized commutator, the difference between the predicted velocity and the target of \cref{eq:vstar}, measures
how the low-resolution path deviates from the downscaled full-resolution path. We train the LoRA update $\Delta\theta$ to
match this target on saved full-resolution states,
\begin{equation}
  \mathcal L_{\mathrm{comm}}(\Delta\theta)=\mathbb E_{c,t}\bigl\|v_{\theta+\Delta\theta}\bigl(\tilde z_{\tau\dn{R}},\tau,c\bigr)-\dot{\tilde z}_{\tau\dn{R}}\bigr\|^2.
  \label{eq:loss-comm}
\end{equation}
We use this loss for video, where backpropagating through a full rollout and the decoder is prohibitive.

%% file: figures/motivation/motivation.tex
\begin{figure}[t]
  \centering
  \definecolor{clow}{HTML}{DD8452}%
  \providecommand{\qpmask}[4]{\begin{scope}[shift={(#3,#4)},x=#1/8,y=#1/8]
    \pgfmathsetseed{7}%
    \foreach \i in {0,...,7}{\foreach \j in {0,...,7}{%
      \pgfmathsetmacro{\v}{round(15+70*rnd)}%
      \pgfmathtruncatemacro{\low}{(\i<4)&&(\j<4)}%
      \ifx#2t\pgfmathtruncatemacro{\sh}{\low}\else\pgfmathtruncatemacro{\sh}{1-\low}\fi
      \ifnum\sh=1 \fill[clow!\v!white]\else\fill[black!\v!white]\fi (\i,-\j) rectangle ++(1,-1);}}
    \ifx#2t\draw[clow,line width=0.9pt] (0,0) rectangle (4,-4);\else
      \draw[clow,line width=0.9pt] (4,0)--(8,0)--(8,-8)--(0,-8)--(0,-4)--(4,-4)--cycle;\fi
    \draw[black!60,line width=0.3pt] (0,0) rectangle (8,-8);
  \end{scope}}%
  \newlength{\qprowh}\setlength{\qprowh}{\dimexpr(\linewidth-1em-5pt)*384/(1554+192)\relax}%
  \begin{tikzpicture}[inner sep=0]
    \node[rotate=90,font=\scriptsize] at (0.4em,-0.5\qprowh) {new high freq.};
    \node[rotate=90,font=\scriptsize] at (0.4em,-1.5\qprowh-1.5pt) {new low freq.};
    \qpmask{0.5\qprowh}{t}{1em}{-0.25\qprowh}
    \qpmask{0.5\qprowh}{b}{1em}{-1.25\qprowh-1.5pt}
    \node[anchor=north west] at (1em+0.5\qprowh+5pt,0)
      {\includegraphics[height=\qprowh,trim=0 390 0 0,clip]{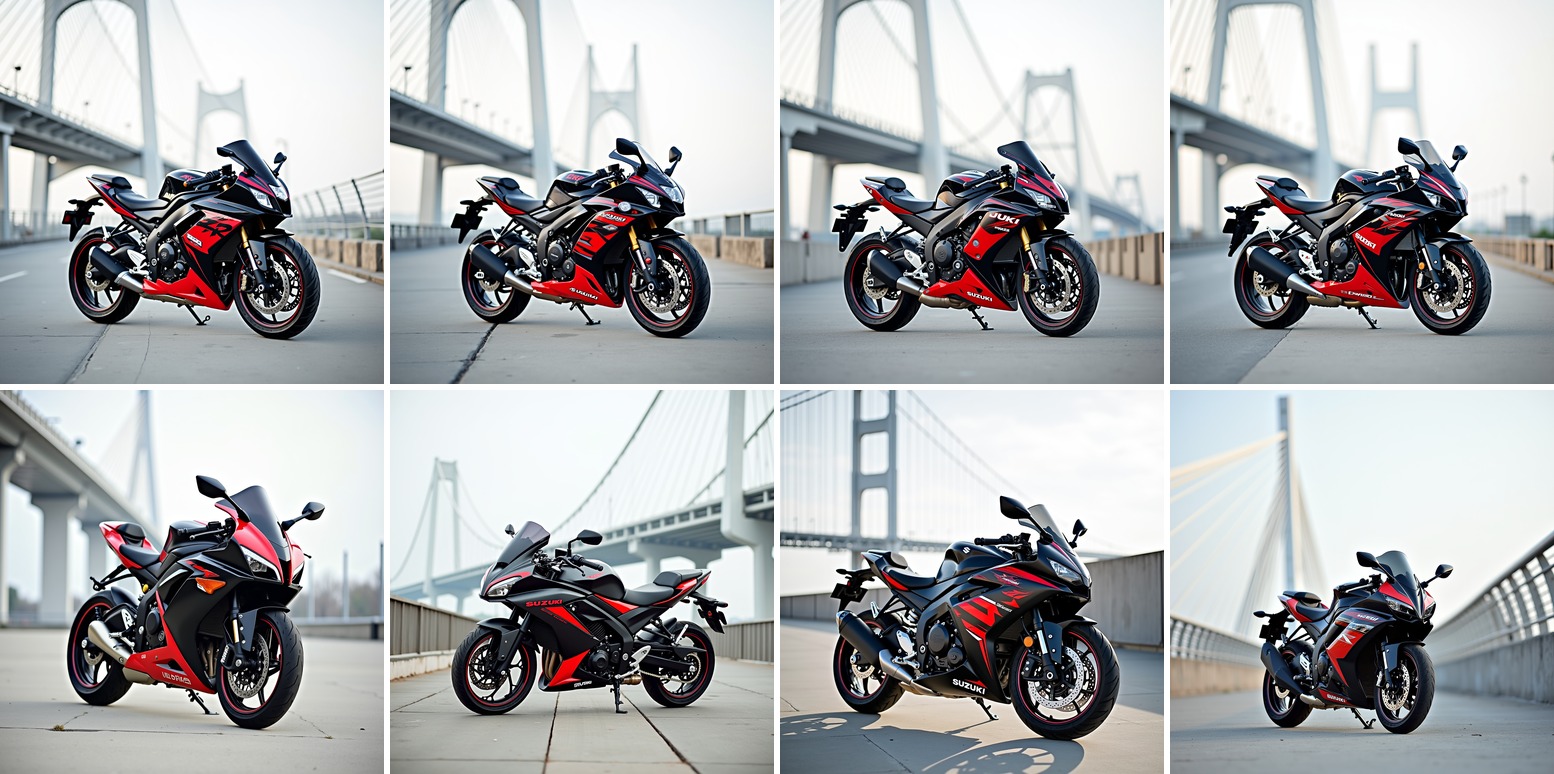}};
    \node[anchor=north west] at (1em+0.5\qprowh+5pt,-\qprowh-1.5pt)
      {\includegraphics[height=\qprowh,trim=0 0 0 390,clip]{figures/motivation/motivation.jpg}};
  \end{tikzpicture}\par\vspace{2pt}
  {\scriptsize \tikz[x=2.2pt,y=2.2pt,baseline=0pt]{\foreach \i/\v in {0/70,1/35,2/55}{\foreach \j/\w in {0/30,1/60,2/45}{%
    \pgfmathtruncatemacro{\c}{mod(\v+\w,70)+15}\fill[clow!\c!white] (\i,\j) rectangle ++(1,1);}}
    \draw[clow,line width=0.6pt] (0,0) rectangle (3,3);}~shared across the row\quad
  \tikz[x=2.2pt,y=2.2pt,baseline=0pt]{\foreach \i/\v in {0/70,1/35,2/55}{\foreach \j/\w in {0/30,1/60,2/45}{%
    \pgfmathtruncatemacro{\c}{mod(\v+\w,70)+15}\fill[black!\c!white] (\i,\j) rectangle ++(1,1);}}
    \draw[black!60,line width=0.3pt] (0,0) rectangle (3,3);}~random per image}
  \caption{\textbf{The low-frequency band of the noise decides the image structure.} Redrawing the high-frequency
  three quarters of the DCT coefficients barely changes the image (top), while redrawing only the lowest quarter changes
  its layout (bottom).}
  \label{fig:motivation}
\end{figure}

%% file: tables/overhead/overhead.tex
\begin{table*}[t]
  \centering
  \small
  \setlength{\tabcolsep}{3.4pt}%
  \begin{tabular}{@{}l ccccc@{\hspace{5pt}}!{\color{black!30}\vrule width 0.4pt}@{\hspace{5pt}}ccccc@{}}%
    \toprule
    & \multicolumn{5}{c}{FLUX.1-dev} & \multicolumn{5}{c}{SD3.5-Large} \\
    \cmidrule(l{0.5em}r{10.4pt}){2-6} \cmidrule(l{0pt}){7-11}%
    Method & \makecell{Speed\\$\uparrow$} & \makecell{ImageReward\\$\uparrow$} & \makecell{DreamSim\\$\downarrow$} & \makecell{DiffSim\\$\uparrow$} & \makecell{LPIPS\\$\downarrow$}
      & \makecell{Speed\\$\uparrow$} & \makecell{ImageReward\\$\uparrow$} & \makecell{DreamSim\\$\downarrow$} & \makecell{DiffSim\\$\uparrow$} & \makecell{LPIPS\\$\downarrow$} \\
    \midrule
    Full image (30) & 1.00$\times$ & 0.89 & -- & -- & -- & 1.00$\times$ & 0.89 & -- & -- & -- \\
    \midrule
    Fewer steps (10) & 2.94$\times$ & 0.71 & 0.166 & 0.813 & 0.404 & 2.95$\times$ & 0.72 & 0.159 & 0.805 & 0.415 \\
    Clean-output pred.\ (10) & 2.94$\times$ & 0.29 & 0.183 & 0.780 & 0.408 & 2.95$\times$ & 0.12 & 0.158 & 0.860 & 0.384 \\
     Downsampled seed & 3.36$\times$ & \textbf{0.88} & 0.130 & 0.856 & 0.387 & 3.60$\times$ & 0.53 & 0.261 & 0.691 & 0.603 \\
    Jeong + TaylorSeer & 3.21$\times$ & 0.79 & 0.110 & 0.890 & 0.338 & 3.09$\times$ & 0.71 & 0.165 & 0.816 & 0.453 \\
    SeaCache~\cite{chung2026seacache} & 3.01$\times$ & 0.84 & 0.050 & \underline{0.950} & \underline{0.179} & 2.95$\times$ & 0.71 & 0.072 & \underline{0.926} & \underline{0.238} \\
    PreFLOP & 3.11$\times$ & \underline{0.87} & \underline{0.049} & 0.945 & 0.205 & 3.02$\times$ & \underline{0.77} & \underline{0.069} & 0.923 & 0.246 \\
    PreFLOP, trained & 3.11$\times$ & \underline{0.87} & \textbf{0.046} & \textbf{0.957} & \textbf{0.178} & 3.02$\times$ & \textbf{0.85} & \textbf{0.047} & \textbf{0.953} & \textbf{0.185} \\
    \midrule
    Fewer steps (6) & 4.81$\times$ & 0.37 & 0.265 & 0.685 & 0.522 & 4.84$\times$ & 0.44 & 0.242 & 0.694 & 0.544 \\
    Clean-output pred.\ (6) & 4.81$\times$ & $-$0.38 & 0.288 & 0.722 & 0.502 & 4.84$\times$ & $-$0.65 & 0.309 & 0.723 & 0.533 \\
    Downsampled seed & 4.50$\times$ & \underline{0.85} & 0.170 & 0.809 & 0.435 & 4.99$\times$ & 0.37 & 0.305 & 0.643 & 0.626 \\
    Jeong + TaylorSeer & 5.10$\times$ & 0.43 & 0.261 & 0.691 & 0.545 & 4.90$\times$ & 0.49 & 0.281 & 0.646 & 0.651 \\
    SeaCache~\cite{chung2026seacache} & 4.81$\times$ & 0.65 & 0.142 & 0.816 & 0.378 & 4.84$\times$ & 0.38 & 0.170 & 0.823 & 0.428 \\
    PreFLOP & 5.04$\times$ & \underline{0.85} & \underline{0.089} & \underline{0.906} & \underline{0.296} & 5.08$\times$ & \underline{0.66} & \underline{0.135} & \underline{0.827} & \underline{0.410} \\
    PreFLOP, trained & 5.04$\times$ & \textbf{0.86} & \textbf{0.078} & \textbf{0.923} & \textbf{0.255} & 5.08$\times$ & \textbf{0.80} & \textbf{0.093} & \textbf{0.904} & \textbf{0.292} \\
    \bottomrule
  \end{tabular}
  \caption{\textbf{Image previews.} ImageReward and similarity to the full image (steps in parentheses). Best in bold,
  second-best underlined.}
  \label{tab:overhead}
\end{table*}

%% file: sections/experiments.tex
\section{Experiments}
\label{sec:experiments}

\paragraph{Setup.}
In our experiments, we quantify how close each method's preview is to the area-downscaled full output of the same
prompt and seed, and how well it serves downstream tasks such as seed selection. We measure cost as the speed-up over
the full generation, timing both with the VAE decode included.
We compare with five baselines, each at the cost of the preview it is compared
with: the model's decoded clean-output prediction; a complete run with fewer steps; a downsampled seed (the seed's
low frequencies, sampled with all steps at the preview size); a complete run accelerated by SeaCache~\cite{chung2026seacache}; and, for images, Jeong et al.'s
previews~\cite{jeong2026previews} with TaylorSeer~\cite{liu2025taylorseer}, as they combine them for larger speed-ups. We calibrate each baseline for every speed-up to maximize its
performance. Our previews are training-free unless marked as trained.
Implementation details, additional
experiments and visual examples are in the supplementary material.

\subsection{Images}
\label{sec:exp-images}

We preview FLUX.1-dev~\cite{bfl2024flux} and SD3.5-Large~\cite{esser2024sd3} images ($1024^2$, 30 steps) on 500
held-out prompts of the PixArt-30K evaluation set~\cite{chen2024pixartsigma}, at $768^2$ for 3$\times$ and $512^2$ from 5$\times$ on. We measure
similarity to the full image with DreamSim~\cite{fu2023dreamsim}, DiffSim~\cite{song2025diffsim} and
LPIPS~\cite{zhang2018lpips} on AlexNet features, and quality with ImageReward~\cite{xu2023imagereward}.

Our previews are the closest to the full image at every speed-up (\cref{tab:overhead,fig:examples}).
Beyond 5$\times$, SeaCache and Jeong et al.'s previews with TaylorSeer degrade sharply, whereas ours degrade slowly
up to 10$\times$ (\cref{fig:teaser}). Training brings our previews even closer to the full image. Although
neither the schedule search nor the training targets image quality, our SD3.5 previews score higher in ImageReward
than native low-resolution generation (the downsampled seed), while our FLUX previews retain its quality.

\input{figures/examples/examples}

\subsection{Videos}
\label{sec:exp-video}

\input{tables/video/video}

We preview Wan2.1-1.3B~\cite{wan2025wan} videos ($480$p, 81 frames, 50 steps) at $320$p for 5$\times$ and $240$p
for 10$\times$. We measure their similarity to the full video with patch
V-JEPA~2 features~\cite{assran2025vjepa2} (VJ-patch), patch VideoMAE
features~\cite{tong2022videomae} and LPIPS~\cite{zhang2018lpips}, and their quality with VideoReward~\cite{liu2025videoreward}, a learned
human-preference score that uses no reference (\cref{tab:video}). At both speed-ups our training-free previews are closer to the full video than
every baseline under every similarity measure, and training the LoRA with the commutator loss increases the similarity
further. VideoReward rates our previews above every baseline at 10$\times$, and below fewer steps and SeaCache at
5$\times$.

\input{tables/handoff/handoff}

\paragraph{Hand-over to a smaller model.}
A preview can also hand the state over to a smaller model of the same family~\cite{cheng2025srdiffusion}. We preview Wan2.1-14B videos with
Wan2.1-1.3B, which shares its VAE and text encoder and whose steps cost a small fraction of a 14B step at the same resolution.
At 8$\times$ and 15$\times$ these previews are closer to the full 14B video than every baseline
under every similarity measure (\cref{tab:handoff}). VideoReward rates them on par with the full video and above every baseline. These previews are
training-free.

\subsection{Seed selection by reward models}
\label{sec:exp-order}

\input{figures/order/order}
\input{figures/best8/best8}

Automated selection scores a preview of each seed with a reward model, and keeps the seeds that score highest.
Such a method therefore needs the reward model to rank a prompt's seeds by their previews as
it ranks them by their full images. We report the percentage of seed pairs that the two rankings order alike. For images, we use seven scorers of prompt
adherence~\cite{ghosh2023geneval,hessel2021clipscore,lin2024vqascore}, human
preference~\cite{xu2023imagereward,kirstain2023pickapic,wu2023hpsv2} and aesthetic quality~\cite{schuhmann2022laion5b}.
For videos, we use six VBench~\cite{huang2024vbench}
measures of consistency, quality and motion.
Averaged over the scorers, our previews preserve the seed order better than every baseline on FLUX.1-dev, SD3.5-Large
and Wan2.1-1.3B, especially at higher speed-ups (\cref{fig:order}).

Jia et al.~\cite{jia2026faithfulpreviews} evaluate previews by the seed they keep: given a prompt generated with eight
seeds, the one whose preview scores highest is kept, and the score of its full image is reported on a scale where 0 is
the average score of the eight and 100 their maximum. We follow their protocol, with FLUX.1-dev images at $512^2$ (28 steps) and
previews at $384^2$, on 512 randomly sampled COCO val2017 captions~\cite{lin2014coco} with eight seeds each.
Without any training, our previews are on par with
their trained model (\cref{fig:best8}). With light training, we score 10 to 20 points higher under every scorer, although our training set is 36 times smaller and our training takes about 6 H100 GPU hours, against an estimated 47
for theirs.

\subsection{Human evaluation}
\label{sec:exp-human}

\input{tables/human/human}

We complement the learned similarity metrics with human judgments from 25 volunteers per medium. Each trial presents a reference and four
candidates with the question ``Which image/video is most similar to the reference?'', and the rater selects a
candidate or answers ``Don't know''. Raters
perform two tasks. In the comparison task, the reference is the full output and the candidates are the different previews
at matched cost: our preview, the clean-output prediction or the fewer-step run, SeaCache, and, for images at
3$\times$ and 5$\times$, Jeong et al.\ with TaylorSeer. In the identification task, the reference is our preview and
the candidates are the full outputs of four seeds of the same prompt. Trials cover FLUX.1-dev and Wan2.1-1.3B, on
prompts and seeds drawn at random.

\Cref{tab:human} reports the results. In the comparison task, SeaCache is competitive at 3$\times$, and our preview is
selected far more often than any baseline as the speed-up grows. In the identification task, accuracy declines with the
speed-up, and ``Don't know'' answers become more frequent. Full results and implementation details are given in the supplementary material.

\subsection{Ablations}
\label{sec:exp-ablations}

\input{tables/ablations/ablations}

\paragraph{Downsampling operator.}
We compare downsampling operators at a single downsampling step from $1024^2$ to $512^2$ after 3 or 6 of 30 steps, on the 500
held-out prompts (\cref{tab:ablations}). Each row names a way of downsampling the state and the time it is given:
\begin{itemize}[nosep,leftmargin=*]
  \item \emph{DCT, $\tau$}: ours, DCT downsampling relabeled to time $\tau$.
  \item \emph{Mean-pooling, $\tau$}: averages each $2\times2$ block, which is identical to blockwise DCT downsampling,
  and is relabeled to $\tau$.
  \item \emph{DCT, $t$}: our downsampling, re-noised back to the downsampling time $t$ with the seed's own noise,
  $z_t^{\mathrm{ren}}=D_R(z_t)+t\,(1-1/R)\,\epsilon_{\dn{R}}$, instead of relabeled to $\tau$.
  \item \emph{Min.-commutator pixel, $t$}: keeps one latent pixel of every $2\times2$ block,
  choosing among the four pixel patterns the one whose commutator with the velocity field is smallest, as Jeong
  et al.~\cite{jeong2026previews} do.
  \item \emph{Random pixel, $t$}: keeps a random pixel of every $2\times2$ block, with no selection.
\end{itemize}
As \cref{tab:ablations}
shows, both relabeling to $\tau$ and low-pass filtering, by either operator, correlate with lower DreamSim and a lower velocity
commutator (\cref{eq:vstar}).

%% file: figures/examples/examples.tex
\input{figures/examples/speedups}
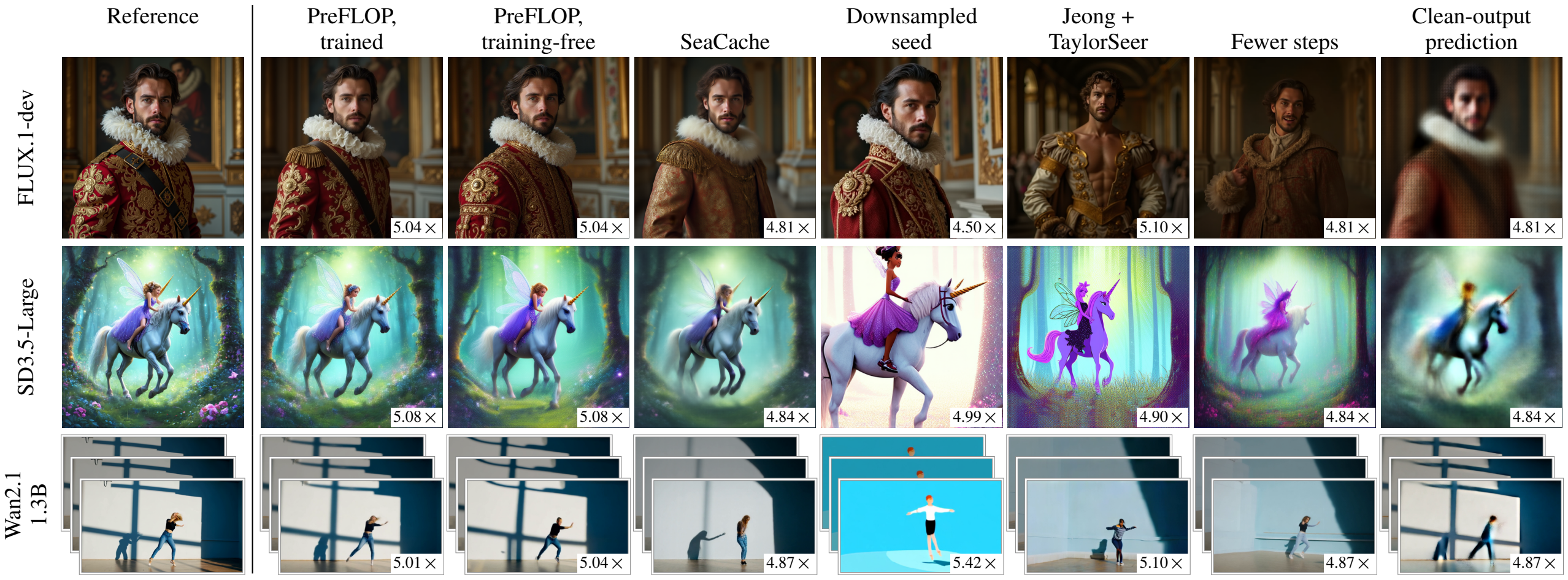
\begin{figure*}[t]
  \centering
  \newcommand{\gpanelw}{0.112}%
  \newcommand{\gcolsep}{0.8pt}%
  \newcommand{\grefsep}{2.5pt}%
  \newcommand{\gheadsep}{1.5pt}%
  \newcommand{\growsep}{2.5pt}%
  \setlength{\tabcolsep}{\gcolsep}\renewcommand{\arraystretch}{0}%
  \newcommand{\ghead}[2]{\vtop{\hbox to \gpanelw\linewidth{\hss\strut#1\hss}\hbox to \gpanelw\linewidth{\hss\strut#2\hss}}}%
  \newcommand{\gtile}[2]{\begin{tikzpicture}[baseline=(i.south)]
    \node[inner sep=0pt,outer sep=0pt] (i) {\includegraphics[width=\gpanelw\linewidth]{figures/examples/#1/#2.png}};
    \ifcsname exspd@#1@#2\endcsname\node[anchor=south east,fill=white,inner sep=0.8pt,font=\tiny] at (i.south east)
      {\csname exspd@#1@#2\endcsname};\fi
    \end{tikzpicture}}%
  \newcommand{\gvidaspect}{0.5769}%
  \newcommand{\gstackdx}{0.05}%
  \newcommand{\gstackdy}{0.12}%
  \newcommand{\gstackdim}{0.3}%
  \newcommand{\gclip}[2]{\begin{tikzpicture}[baseline=(f2.south),x=\gpanelw\linewidth,y=\gpanelw\linewidth]
    \pgfmathsetmacro{\gsh}{\gvidaspect*(1-2*\gstackdx)+2*\gstackdy}%
    \useasboundingbox (0,0) rectangle (1,\gsh);
    \pgfmathsetlengthmacro{\gfw}{(1-2*\gstackdx)*\gpanelw*\linewidth}%
    \foreach \k/\fr in {0/30,1/35,2/40}{%
      \node[anchor=north west,inner sep=0pt,outer sep=0pt] (f\k) at (\k*\gstackdx,\gsh-\k*\gstackdy)
        {\includegraphics[width=\gfw]{figures/examples/#1_frames/#2_f\fr.png}};
      \ifnum\k<2 \fill[black,opacity=\gstackdim] (f\k.south west) rectangle (f\k.north east);\fi
      \draw[black!45,line width=0.9pt] (f\k.south west) rectangle (f\k.north east);
      \draw[white,line width=0.5pt] (f\k.south west) rectangle (f\k.north east);}
    \ifcsname exspd@#1@#2\endcsname\node[anchor=south east,fill=white,inner sep=0.8pt,font=\tiny] at (f2.south east)
      {\csname exspd@#1@#2\endcsname};\fi
    \end{tikzpicture}}%
  \newcommand{\gnone}{}%
  \newcommand{\glabel}[3]{\begin{tikzpicture}[baseline={([yshift=-0.5*#3*\gpanelw*\linewidth]current bounding box.center)}]
    \node[rotate=90,inner sep=0pt] (l) {\renewcommand{\arraystretch}{1}\begin{tabular}{@{}c@{}}#2\end{tabular}};
    \ifx\relax#1\relax\else\node[anchor=south,inner sep=0pt] at ([yshift=2pt]l.east) {#1};\fi
    \end{tikzpicture}}%
  \ifdefined\tikzexternaldisable\tikzexternaldisable\fi
  {\scriptsize  %
  \begin{tabular}{@{}c@{\hspace{\grefsep}}c@{\hspace{\grefsep}}|@{\hspace{\grefsep}}ccccccc@{}}
    & \ghead{Reference}{} & \ghead{PreFLOP,}{trained} & \ghead{PreFLOP,}{training-free} & \ghead{}{SeaCache} & \ghead{Downsampled}{seed} & \ghead{Jeong +}{TaylorSeer} & \ghead{}{Fewer steps} & \ghead{Clean-output}{prediction} \\[\gheadsep]
    \glabel{}{FLUX.1-dev}{1.0000} & \gtile{flux_p0195}{ref} & \gtile{flux_p0195}{trained} & \gtile{flux_p0195}{free} & \gtile{flux_p0195}{seacache} & \gtile{flux_p0195}{k0} & \gtile{flux_p0195}{jts} & \gtile{flux_p0195}{fewer} & \gtile{flux_p0195}{x0} \\[\growsep]
    \glabel{}{SD3.5-Large}{1.0000} & \gtile{sd35_p0223}{ref} & \gtile{sd35_p0223}{trained} & \gtile{sd35_p0223}{free} & \gtile{sd35_p0223}{seacache} & \gtile{sd35_p0223}{k0} & \gtile{sd35_p0223}{jts} & \gtile{sd35_p0223}{fewer} & \gtile{sd35_p0223}{x0} \\[\growsep]
    \glabel{}{Wan2.1\\1.3B}{0.7592} & \gclip{wan_p00599}{ref} & \gclip{wan_p00599}{trained} & \gclip{wan_p00599}{free} & \gclip{wan_p00599}{seacache} & \gclip{wan_p00599}{k0} & \gclip{wan_p00599}{jts} & \gclip{wan_p00599}{fewer} & \gclip{wan_p00599}{x0}
  \end{tabular}}
  \ifdefined\tikzexternalenable\tikzexternalenable\fi
  \caption{\textbf{Previews at about 5$\times$.} One prompt per model, and one frame of the Wan2.1-1.3B video.}
  \label{fig:examples}
\end{figure*}

%% file: figures/examples/speedups.tex
\expandafter\def\csname exspd@flux_p0195@trained\endcsname{5.04$\times$}
\expandafter\def\csname exspd@flux_p0195@free\endcsname{5.04$\times$}
\expandafter\def\csname exspd@flux_p0195@seacache\endcsname{4.81$\times$}
\expandafter\def\csname exspd@flux_p0195@k0\endcsname{4.50$\times$}
\expandafter\def\csname exspd@flux_p0195@jts\endcsname{5.10$\times$}
\expandafter\def\csname exspd@flux_p0195@fewer\endcsname{4.81$\times$}
\expandafter\def\csname exspd@flux_p0195@x0\endcsname{4.81$\times$}
\expandafter\def\csname exspd@sd35_p0223@trained\endcsname{5.08$\times$}
\expandafter\def\csname exspd@sd35_p0223@free\endcsname{5.08$\times$}
\expandafter\def\csname exspd@sd35_p0223@seacache\endcsname{4.84$\times$}
\expandafter\def\csname exspd@sd35_p0223@k0\endcsname{4.99$\times$}
\expandafter\def\csname exspd@sd35_p0223@jts\endcsname{4.90$\times$}
\expandafter\def\csname exspd@sd35_p0223@fewer\endcsname{4.84$\times$}
\expandafter\def\csname exspd@sd35_p0223@x0\endcsname{4.84$\times$}
\expandafter\def\csname exspd@wan_p00599@trained\endcsname{5.01$\times$}
\expandafter\def\csname exspd@wan_p00599@free\endcsname{5.04$\times$}
\expandafter\def\csname exspd@wan_p00599@seacache\endcsname{4.87$\times$}
\expandafter\def\csname exspd@wan_p00599@k0\endcsname{5.42$\times$}
\expandafter\def\csname exspd@wan_p00599@jts\endcsname{5.10$\times$}
\expandafter\def\csname exspd@wan_p00599@fewer\endcsname{4.87$\times$}
\expandafter\def\csname exspd@wan_p00599@x0\endcsname{4.87$\times$}

%% file: tables/video/video.tex
\begin{table}[t]
  \centering
  \footnotesize
  \setlength{\tabcolsep}{2pt}
  \begin{tabular}{@{}l c cccc@{}}
    \toprule
    Method & \makecell{Speed\\$\uparrow$} & \makecell{VJ-patch\\$\uparrow$} & \makecell{VideoMAE\\$\uparrow$} & \makecell{LPIPS\\$\downarrow$} & \makecell{Video\\Reward $\uparrow$} \\
    \midrule
    Wan2.1-1.3B (50) & 1.00$\times$ & -- & -- & -- & $-0.55$ \\
    \midrule
    Fewer steps (9) & 4.87$\times$ & 0.448 & 0.824 & 0.495 & \underline{$-1.33$} \\
    Clean-output pred.\ (9) & 4.87$\times$ & 0.465 & 0.875 & 0.435 & $-2.86$ \\
    Downsampled seed & 4.76$\times$ & 0.403 & 0.755 & 0.570 & $-2.37$ \\
    SeaCache~\cite{chung2026seacache} & 4.87$\times$ & 0.529 & 0.900 & 0.321 & $\mathbf{-1.18}$ \\
    PreFLOP & 5.04$\times$ & \underline{0.543} & \underline{0.907} & \underline{0.256} & $-1.56$ \\
    PreFLOP, trained & 5.01$\times$ & \textbf{0.562} & \textbf{0.922} & \textbf{0.229} & $-1.49$ \\
    \midrule
    Fewer steps (4) & 9.23$\times$ & 0.391 & 0.793 & 0.636 & $-2.79$ \\
    Clean-output pred.\ (4) & 9.23$\times$ & 0.367 & 0.798 & 0.613 & $-3.47$ \\
    Downsampled seed & 9.40$\times$ & 0.381 & 0.733 & 0.622 & $-2.87$ \\
    SeaCache~\cite{chung2026seacache} & 9.23$\times$ & 0.368 & 0.804 & 0.645 & $-3.95$ \\
    PreFLOP & 10.02$\times$ & \underline{0.467} & \underline{0.840} & \underline{0.394} & \underline{$-1.86$} \\
    PreFLOP, trained & 10.02$\times$ & \textbf{0.496} & \textbf{0.868} & \textbf{0.343} & $\mathbf{-1.62}$ \\
    \bottomrule
  \end{tabular}
  \caption{\textbf{Video previews.} Similarity to the 50-step Wan2.1-1.3B video and VideoReward~\cite{liu2025videoreward}, 300 held-out videos.}
  \label{tab:video}
\end{table}

%% file: tables/handoff/handoff.tex
\begin{table}[t]
  \centering
  \footnotesize
  \setlength{\tabcolsep}{2pt}
  \begin{tabular}{@{}l c cccc@{}}
    \toprule
    Method & \makecell{Speed\\$\uparrow$} & \makecell{VJ-patch\\$\uparrow$} & \makecell{VideoMAE\\$\uparrow$} & \makecell{LPIPS\\$\downarrow$} & \makecell{Video\\Reward $\uparrow$} \\
    \midrule
    Wan2.1-14B (50) & 1.00$\times$ & -- & -- & -- & $0.00$ \\
    \midrule
    Fewer steps (6) & 8.14$\times$ & 0.410 & 0.817 & 0.570 & $-2.01$ \\
    Clean-output pred.\ (6) & 8.14$\times$ & \underline{0.485} & \underline{0.891} & 0.462 & $-1.99$ \\
    Downsampled seed & 7.74$\times$ & 0.449 & 0.827 & 0.469 & \underline{$-0.35$} \\
    SeaCache~\cite{chung2026seacache} & 8.14$\times$ & 0.473 & 0.883 & \underline{0.436} & $-1.74$ \\
    PreFLOP & 8.12$\times$ & \textbf{0.553} & \textbf{0.920} & \textbf{0.279} & $\mathbf{0.23}$ \\
    \midrule
    Fewer steps (3) & 15.86$\times$ & 0.362 & 0.789 & 0.665 & $-3.07$ \\
    Clean-output pred.\ (3) & 15.86$\times$ & 0.391 & \underline{0.810} & 0.589 & $-2.91$ \\
    Downsampled seed & 15.03$\times$ & \underline{0.431} & 0.807 & \underline{0.509} & \underline{$-0.60$} \\
    SeaCache~\cite{chung2026seacache} & 15.86$\times$ & 0.339 & 0.800 & 0.619 & $-3.71$ \\
    PreFLOP & 15.66$\times$ & \textbf{0.497} & \textbf{0.880} & \textbf{0.371} & $\mathbf{0.12}$ \\
    \bottomrule
  \end{tabular}
  \caption{\textbf{Large-to-small hand-over.} PreFLOP runs Wan2.1-14B at $720$p and $480$p, then Wan2.1-1.3B
  at $480$p. Similarity to the full 14B video and VideoReward~\cite{liu2025videoreward}, at $480$p on 100 held-out
  videos.}
  \label{tab:handoff}
\end{table}

%% file: figures/order/order.tex
\begin{figure}[t]
  \centering
  \includegraphics[width=\linewidth]{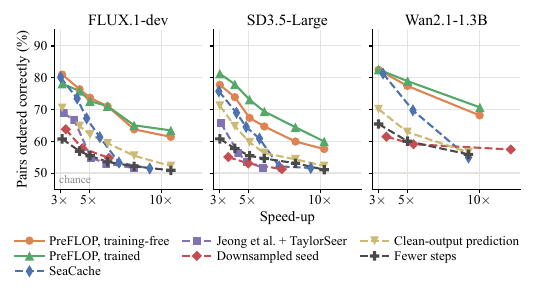}
  \caption{\textbf{Seed order against speed-up.} Percentage of seed pairs that a preview orders as their full outputs
  are ordered under the same scorer, averaged over the scorers (50 is chance).}
  \label{fig:order}
\end{figure}

%% file: figures/best8/best8.tex
\begin{figure}[t]
  \centering
  \includegraphics[width=\linewidth]{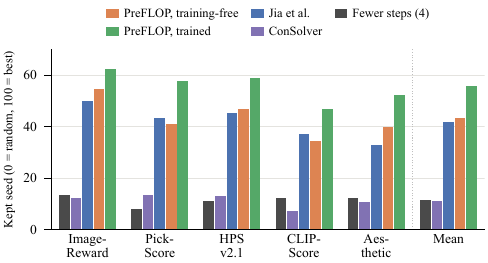}
  \caption{\textbf{Best of eight seeds on FLUX.1-dev, picked by a preview at $7\times$.} Full-image score of the
  picked seed: 0 is a random seed, 100 the best of the eight.}
  \label{fig:best8}
\end{figure}

%% file: tables/human/human.tex
\begin{table}[t]
  \centering
  \footnotesize
  \setlength{\tabcolsep}{3pt}
  \begin{tabular}{@{}ll ccc@{\hspace{2.8pt}}!{\color{black!30}\vrule width 0.4pt}@{\hspace{2.8pt}}ccc@{}}
    \toprule
    & & \multicolumn{3}{c}{Training-free} & \multicolumn{3}{c}{Trained} \\
    \cmidrule(l{0.5em}r{6pt}){3-5} \cmidrule(l{0pt}){6-8}
    & Speed & Ours & \makecell{Best\\baseline} & ID
          & Ours & \makecell{Best\\baseline} & ID \\
    \midrule
    \multirow{3}{*}{Images} & 3$\times$  & \textbf{55\%} & 39\% & 87\% & 41\% & \textbf{49\%} & 93\% \\
           & 5$\times$  & \textbf{79\%} & 12\% & 94\% & \textbf{65\%} & 22\% & 88\% \\
           & 10$\times$ & \textbf{73\%} & 10\% & 54\% & \textbf{73\%} & 7\% & 67\% \\
    \midrule
    \multirow{3}{*}{Video}  & 3$\times$  & \textbf{64\%} & 29\% & 100\% & \textbf{72\%} & 22\% & 100\% \\
           & 5$\times$  & \textbf{89\%} & 7\% & 98\% & \textbf{96\%} & 4\% & 100\% \\
           & 10$\times$ & \textbf{73\%} & 15\% & 70\% & \textbf{88\%} & 4\% & 88\% \\
    \bottomrule
  \end{tabular}
  \caption{\textbf{Human evaluation.} Comparison: share of choices for our preview and for the strongest baseline at
  the same cost. Identification (ID): share of trials in which raters matched our preview to its own full output among
  four seeds.}
  \label{tab:human}
\end{table}

%% file: tables/ablations/ablations.tex
\begin{table}[t]
  \centering
  \small
  \setlength{\tabcolsep}{2.5pt}
  \begin{tabular}{@{}l cc cc@{}}
    \toprule
    & \multicolumn{2}{c}{DreamSim\,$\downarrow$} & \multicolumn{2}{c}{Velocity comm.\,$\downarrow$} \\
    \cmidrule(lr){2-3} \cmidrule(l){4-5}
    Operator, time & $n_1{=}3$ & $n_1{=}6$ & $n_1{=}3$ & $n_1{=}6$ \\
    \midrule
    DCT, $\tau$ (PreFLOP) & \textbf{0.093} & \textbf{0.067} & \textbf{63.8} & \textbf{73.2} \\
    Mean-pooling, $\tau$ & \underline{0.097} & \underline{0.071} & \underline{65.5} & \underline{73.8} \\
    DCT, $t$ & 0.105 & 0.078 & 110.3 & 91.8 \\
    Min.-commutator pixel, $t$~\cite{jeong2026previews} & 0.151 & 0.113 & 135.7 & 125.7 \\
    Random pixel, $t$ & 0.163 & 0.121 & 153.5 & 135.9 \\
    \bottomrule
  \end{tabular}
  \caption{\textbf{Downsampling operators on FLUX.1-dev.} DreamSim to the full image and instantaneous velocity
  commutator of $512^2$ previews that hand over once.}
  \label{tab:ablations}
\end{table}

%% file: sections/discussion.tex
\section{Spectral Consistency}
\label{sec:discussion}

Ideally, a preview would be the downsampled full output, generated using only steps at the preview resolution. A
model can do that if it is spectrally consistent: its state can be downsampled at any timestep and continued at a
lower resolution to produce the downsampled full-resolution output. This property holds when downsampling commutes
with each of the three components of the model: the decoder that maps the final latent to an image, the velocity
field that moves the intermediate states, and the prior that draws the initial noise. We define a commutator for
each, whose magnitude measures how far the model is from consistency.

\noindent\emph{Decoder.} The decoder commutator
\begin{equation}
  [D,D',\mathrm{Dec}](z_0)=D'\bigl(\mathrm{Dec}(z_0)\bigr)-\mathrm{Dec}\bigl(D(z_0)\bigr)
  \label{eq:decoder-commutator}
\end{equation}
is the gap between the downsampled image and the image decoded from the downsampled latent, where $D'$ downsamples in
pixel space and $D$ in latent space. When it is zero, downsampling the latent with $D$ behaves as downsampling the
image with $D'$. Here $D'(\mathrm{Dec}(z_0))$ is the downsampled full-resolution output of the definition above, and
thus the image a spectrally consistent model produces at the lower resolution. $D'$ should therefore preserve the
image, as standard resampling filters such as area, bilinear and bicubic do. Autoencoders
with a scale-equivariance loss are trained to satisfy a version of
it~\cite{kouzelis2025eqvae,skorokhodov2025diffusability}.

\noindent\emph{Velocity.} The velocity commutator
\begin{equation}
  [D,v_\theta](z_t,t)=D\bigl(v_\theta(z_t,t,c)\bigr)-v_\theta\bigl(D(z_t),t,c\bigr)
  \label{eq:commutator}
\end{equation}
is the gap between the downsampled velocity and the velocity predicted from the downsampled state at the same
time~\cite{jeong2026previews}. When it is zero, the trajectory from $D(z_t)$ is the downsampled full trajectory.
Implicitly, a zero commutator requires the model to predict the low band of its velocity from the low band of its
state. This form assumes no timestep alignment, which a zero prior commutator makes unnecessary, unlike in
\cref{eq:vstar}.

\noindent\emph{Prior.} The prior commutator is the gap between the distribution of downsampled noise and the model's prior at
the lower resolution, $p_1^{\mathrm{low}}$. When it is zero, a downsampled noise sample is a typical sample of that
prior,
\begin{equation}
  D(\epsilon)\sim p_1^{\mathrm{low}},\quad\epsilon\sim p_1.
  \label{eq:prior-commutator}
\end{equation}
With $p_1=\mathcal N(0,I)$, the prior at the lower resolution is therefore not $\mathcal N(0,I)$ but
$\mathcal N(0,DD^{\top})$, the distribution of $D(\epsilon)$, which is $\mathcal N(0,I/R^2)$ for ideal downsampling
by $R$.
Current models use $\mathcal N(0,I)$ everywhere, which is why noise schedules are shifted with
resolution~\cite{hoogeboom2023simple,esser2024sd3} and why we renormalize and relabel the state (\cref{sec:downsampling}).

\input{figures/tradeoff/tradeoff}

Current models satisfy none of the three exactly. Our method realigns the state for the prior commutator, and relies
on the velocity and decoder commutators being small, which depends on when the state is downsampled. Too early, the
velocity commutator accumulates over the low-resolution steps and the preview drifts; too late, the preview ends
near $D(z_0)$, which the decoder commutator degrades (\cref{fig:tradeoff}). For each target speed-up, the schedule
search of \cref{sec:schedules} selects the downsampling time that best balances the two. A spectrally consistent
model has no such trade-off: its low-resolution trajectory is the downsampled full-resolution trajectory. It could
therefore move almost seamlessly between resolutions, which would benefit multi-resolution sampling in
general~\cite{xiao2026speed,jeong2025mixedres,tian2025bottleneck,zheng2026mrflow}. Spectral consistency is also
trainable and measurable: the prior is a design choice, and the decoder and velocity commutators can be minimized as
training losses. Training such a model, however, is beyond the scope of this work.

%% file: figures/tradeoff/tradeoff.tex
\begin{figure}[t]
  \centering
  \includegraphics[width=\linewidth]{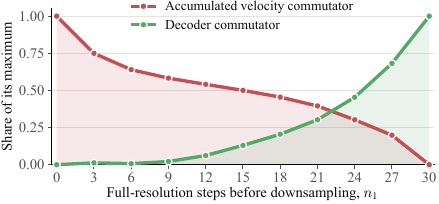}
  \caption{\textbf{The commutator trade-off} on FLUX.1-dev ($1024^2\to512^2$). Velocity: the commutator of
  \cref{eq:vstar}, summed over the low-resolution steps. Decoder: the preview's component along the decoder commutator
  of \cref{eq:decoder-commutator}, in DreamSim space.}
  \label{fig:tradeoff}
\end{figure}

%% file: sections/conclusion.tex
\section{Conclusions}
\label{sec:conclusion}

We presented PreFLOP, a training-free method for previewing flow models, grounded in sampling theory. It pushes the
frontier of speed-up against fidelity beyond the tested baselines, on both images and video. In addition, human raters
choose its previews far more often than any baseline as the speed-up grows, and reward models rank outputs by their
previews more reliably. Finally, we formulated the limits of this approach as three spectral consistency conditions,
and suggested how future models could be trained to satisfy them. Taking a wider view, creating an image or video
usually means generating many drafts and discarding nearly all of them, yet most acceleration methods treat every
output as the one that will be kept. We hope this work encourages separating the two: aggressive speed-ups for drafts,
and careful ones for the final output.

%% file: sections/supplementary.tex
\maketitlesupplementary
\setcounter{section}{0}\setcounter{figure}{0}\setcounter{table}{0}
\renewcommand{\thesection}{S\arabic{section}}\renewcommand{\theHsection}{S\arabic{section}}
\renewcommand{\thefigure}{S\arabic{figure}}\renewcommand{\theHfigure}{S\arabic{figure}}
\renewcommand{\thetable}{S\arabic{table}}\renewcommand{\theHtable}{S\arabic{table}}

This supplement follows the order of the main paper. It first examines what downsampling discards
(\cref{sec:supp-discard}), then derives the target of the commutator loss (\cref{sec:supp-target}), gives the implementation details of every experiment and the schedules
(\cref{sec:supp-impl}), the seed order per scorer (\cref{sec:supp-order}) and the full human evaluation
(\cref{sec:supp-human}). It then shows examples of the downsampling ablation, adds
ablations of fixed schedules and of the window of the video LoRA (\cref{sec:supp-ablations}), shows previews for
every downsampling time (\cref{sec:supp-n1}) and ends with qualitative examples (\cref{sec:supp-qual}).
\ifdefined\ArxivVersion\else %
Works cited in the main paper are named without a citation, and the reference list holds only works new to
this supplement.
\fi

\noindent\textbf{More visual examples are \ifdefined\ArxivVersion on the project page\else in \texttt{visual\_examples.html}\fi.}

\section{What Downsampling Discards}
\label{sec:supp-discard}

A downsampling with $R=2$ keeps the band $i,j<64$ of the orthonormal DCT of each $128\times128$ latent channel and
discards the band $i\geq64$ or $j\geq64$. We examine both bands on FLUX.1-dev and SD3.5-Large at $1024^2$.

\paragraph{Cosine similarity to the seed.}
For each trajectory, we compute the cosine similarity between a band of the state $z_t$ and the same band of its
initial noise $\epsilon$ (\cref{fig:supp-band-cos}). The retained band departs from the noise early, as the model forms
the image in it. The discarded band stays above 0.998 for the first 15 of 30 steps on both models, so up to that
point it is still the noise of the seed.

\input{figures/supp_band_cos/supp_band_cos}

\paragraph{Replacing the discarded band.}
To test whether this band affects the final image, we replace it after 15 of 30 steps by fresh Gaussian noise at the same noise level, and finish
the generation at full resolution.
The final image moves by a DreamSim of 0.031 on FLUX.1-dev and 0.032 on SD3.5-Large, against 0.188 and 0.234 for
another seed of the same prompt, and stays closer than another seed on all 50 prompts. Replacing the retained band
instead destroys the image (\cref{fig:supp-band-swap}).

\paragraph{Summary.}
These experiments extend the motivation of Sec.~4.1 of the main paper from the initial noise to the states along the
generation. Even halfway through it, the band that downsampling discards is still dominated by the noise of the seed,
and replacing it with other noise has little effect on the final image.

\input{figures/supp_band_swap/supp_band_swap}

\section{Target of the Commutator Loss}
\label{sec:supp-target}

We derive the target $\dot{\tilde z}_{\tau\dn{R}}$ of the commutator loss.
With
\begin{equation}
  q(t)=R-(R-1)\,t,
\end{equation}
the alignment of Eq.~(4) of the main paper reads
\begin{equation}
  \lambda=\frac{R}{q(t)},\qquad \tau=\frac{t}{q(t)},\qquad \tilde z_{\tau\dn{R}}=\lambda\,z_{t\dn{R}}.
\end{equation}
By the chain rule,
\begin{equation}
  \dot{\tilde z}_{\tau\dn{R}}=\frac{\mathrm d\tilde z_{\tau\dn{R}}}{\mathrm d\tau}
  =\frac{\mathrm d\tilde z_{\tau\dn{R}}}{\mathrm dt}\,\frac{\mathrm dt}{\mathrm d\tau}.
\end{equation}
With $v_{t\dn{R}}=D_R(v_t)$ and $q'(t)=-(R-1)$, the product rule gives
\begin{equation}
  \frac{\mathrm d\tilde z_{\tau\dn{R}}}{\mathrm dt}=\frac{R(R-1)}{q(t)^2}\,z_{t\dn{R}}+\frac{R}{q(t)}\,v_{t\dn{R}}.
\end{equation}
Since $\mathrm d\tau/\mathrm dt=R/q(t)^2>0$,
\begin{equation}
  \frac{\mathrm dt}{\mathrm d\tau}=\Bigl(\frac{\mathrm d\tau}{\mathrm dt}\Bigr)^{-1}=\frac{q(t)^2}{R}.
\end{equation}
Multiplying the two gives Eq.~(9) of the main paper,
\begin{equation}
  \dot{\tilde z}_{\tau\dn{R}}=(R-1)\,z_{t\dn{R}}+\bigl[R-(R-1)\,t\bigr]\,v_{t\dn{R}}.
\end{equation}
On a straight path, $v_t=\epsilon-z_0$ and $D_R(\epsilon)=\epsilon_{\dn{R}}/R$, so the target reduces to
$\epsilon_{\dn{R}}-z_{0\dn{R}}$, the velocity of the aligned path.

\section{Implementation Details}
\label{sec:supp-impl}
\label{sec:supp-metrics}
\label{sec:supp-schedules}

\subsection{Setup}
\label{sec:supp-common}

\paragraph{Models and prompts.}
\Cref{tab:supp-models} lists the settings of FLUX.1-dev, SD3.5-Large and
Wan2.1, and their prompts. Prompts are drawn at random from the PixArt-30K evaluation
set and MovieGen Bench~\cite{polyak2024moviegen}, and each seed is derived from its prompt's
index.
\input{tables/supp_models/supp_models}

\paragraph{Metrics and timing.}
Every similarity compares a preview with the full output of the same prompt and seed, both area-downscaled to the same
size: $512^2$ for images, $416\times240$ for Wan2.1-1.3B and $832\times480$ for Wan2.1-14B. The speed-up is the time of
the full generation divided by that of the preview, both including the VAE decode.

\subsection{Schedules}
\label{sec:supp-impl-schedules}

Each model samples on a grid that is uniform in an unshifted time $u\in[0,1]$ and mapped to the time $t$ of the main
paper by its timestep shift, $t=a\,u/(1+(a-1)\,u)$. The shift $a$ is fixed for SD3.5-Large and Wan2.1, and grows with
the number of latent tokens for FLUX.1-dev (\cref{tab:supp-models}). Every resolution therefore has its own map
between $u$ and $t$.

A preview takes the first $n_1$ steps of the model's own $N$-step grid at full resolution, so it coincides with the
full generation up to the first downsampling. The downsampling moves the state from time $t$ to $\tau$ (Sec.~4.2 of
the main paper). The receiving resolution converts $\tau$ to its own unshifted time and spaces all the steps left in
the preview, $n_2+n_3+\dots$, evenly in $u$ from there to 0 (\cref{fig:supp-respacing}). It runs the first $n_2$ of
them, each mapped back to $t$ through its own shift. A further downsampling repeats the rule from its own $\tau$ with
the steps still left, so the last resolution ends at $t=0$. Without a full-resolution step, $\tau=1$, and a preview at
a single lower resolution runs that resolution's own $n_2$-step grid.
In the hand-over, the steps of Wan2.1-1.3B use its own shift.

The search follows SMAC3, with the cost of a schedule computed from measured step and decode
times.
\Cref{tab:supp-schedules} lists the schedules.
\input{figures/supp_respacing/supp_respacing}
\input{tables/supp_schedules/supp_schedules}

\subsection{Baselines}
\label{sec:supp-baselines}

Every baseline is calibrated for each target speed-up.
\begin{itemize}[nosep,leftmargin=*]
  \item \emph{Fewer steps} runs the model's own schedule with fewer steps, and the \emph{clean-output prediction}
  decodes $\hat z_0$ after the same number of steps of the full run. Both take the step count whose speed-up is
  closest to the target.
  \item The \emph{downsampled seed} keeps the DCT low band of the seed, rescaled to unit variance, and runs all of the
  model's steps at the preview size. The size is the one whose speed-up is closest to the target.
  \item \emph{SeaCache} uses its official gate. On images and video alike, we pick the threshold $\delta$ that performs
  best while running on par with or slower than our preview at the same target speed-up.
  \item \emph{Jeong et al.} release no code, so we implement their Algorithm~1. In their own
  setting, one downsampling after 10 of 30 steps, our implementation reaches DreamSim 0.078 on
  FLUX.1-dev and 0.118 on SD3.5-Large, where they report 0.068 and 0.135. As they do for larger speed-ups, we add
  TaylorSeer. Its settings and the downsampling step come from a grid search for the
  lowest DreamSim on the search prompts. Jeong et al.\ report no quantitative results on video, so we report none for
  their method on video either.
  \item \emph{Jia et al.} and \emph{ConSolver} (best of eight)
  release no weights for this setting, so we train both as Jia et al.\ describe. Over the five scorers, our trainings
  reach 41.7 and 11.3 on the scale of the main paper, where Jia et al.\ report 39.6 and 9.6. Jia et al.\ report no
  training time. Their recipe, timed on the GPU of our training, takes 28.0 GPU hours of training and 18.5 to render
  its 36{,}864 references, about 47 in all. This is at $512^2$, the resolution at which Jia et al.\ train. Training at
  $1024^2$, the resolution of our other image experiments, would take much longer and is not guaranteed to succeed.
\end{itemize}

\subsection{Fine-tuning}
\label{sec:supp-impl-images}
\label{sec:supp-impl-video}

Each image model has one LoRA and one map $r_\psi$, trained together with the path loss, every rollout drawing one of
the searched schedules. The map is a small convolutional network conditioned on the noise level and the downsampling
ratio. The video previews use no map: their LoRA is trained and run with the DCT downsampling and alignment of Sec.~4.2 of the
main paper. It is trained with the commutator loss on downsampled states and velocities of 50-step $480$p generations,
first at $240$p and then cycling through $384$p, $320$p and $240$p, and is active only on the first half of the
low-resolution steps (\cref{sec:supp-lora-window}). Every LoRA adapts the attention and MLP projections (ranks in \cref{tab:supp-models}).
All LoRAs are trained with AdamW, a warm-up and a cosine decay, and the map with a learning rate of $5\cdot10^{-5}$.

\subsection{Seed selection}
\label{sec:supp-seed}

For seed order, FLUX.1-dev and SD3.5-Large each generate 125 GenEval prompts with 16 seeds each. Previews and full images are scored by ImageReward,
PickScore, HPS~v2.1, CLIPScore,
VQAScore, the LAION aesthetic predictor and GenEval. Ties of the full
images are dropped, and the percentage is averaged over prompts. For video, Wan2.1-1.3B generates eight seeds
for 82 held-out prompts, scored by six VBench measures.

For best of eight, FLUX.1-dev generates $512^2$ images in 28 steps for 512 COCO val2017 captions,
with eight seeds each, as in Jia et al. Our schedules are searched on other captions at the cost of a batch of eight
seeds. The training-free preview takes 1 step at $512^2$ and 7 at $384^2$, and the trained one 3 and 3, with a LoRA and
a map trained on 1{,}024 other captions. At $384^2$, FLUX's 512 T5 text tokens, most of them padding, are almost half
of the tokens of a step, so the $384^2$ steps of both previews drop the padding beyond the first 64 text tokens.

\section{Seed Order per Scorer}
\label{sec:supp-order}

\Cref{fig:supp-order} breaks the seed order of the main paper down by image scorer and video measure, for our
training-free and trained previews and the five baselines of Fig.~5 of the main paper, Jeong et al.\ + TaylorSeer
on images only.

\input{figures/supp_order/supp_order}

\section{Human Evaluation}
\label{sec:supp-human}

\paragraph{Sessions.} Raters opened a single link and were assigned a session of 25 image or 19 video trials,
including one catch trial whose reference is itself among the candidates. Participation was anonymous. The study was approved by the ethics committee of the authors' institution\ifdefined\ArxivVersion\else{}
(name withheld for anonymous review)\fi, and raters took part voluntarily and gave their consent. Sessions that failed
the catch trial were excluded.

\paragraph{Design.} Each session covers 3$\times$, 5$\times$ and 10$\times$, with the same number of trials at each
speed-up: eight for images and six for video. Low speed-ups have more comparison trials and high speed-ups more
identification trials. Our training-free and trained previews alternate within each speed-up and task. A comparison
trial shows our preview and three baselines at the same speed-up:
\begin{itemize}[nosep,leftmargin=*]
  \item images at 3$\times$ and 5$\times$: SeaCache, Jeong et al.\ with TaylorSeer, and the clean-output prediction or
  the fewer-step run, in alternation.
  \item images at 10$\times$ and video: SeaCache, the clean-output prediction and the fewer-step run.
\end{itemize}
The prompts (48 for images and 38 for video, one seed each) rotate over the sessions, so that each prompt appears with
every speed-up, task and preview. Trial order and candidate positions are random. The share of each candidate is
averaged first within a prompt and then over prompts, with 95\% bootstrap intervals over prompts.

\Cref{fig:supp-human} gives the full results.

\input{figures/supp_human/supp_human}

\section{Ablations}
\label{sec:supp-ablations}

\subsection{Downsampling operators}
\label{sec:supp-handover}

\Cref{fig:supp-ablation} shows the previews of every operator of the main paper's downsampling ablation for one
held-out prompt, after $n_1=3$ and $n_1=6$ full-resolution steps. Both pixel selections are farther from the full
image than the DCT and mean-pooling operators, and more so after three steps. \ifdefined\ArxivVersion The project page\else \texttt{visual\_examples.html}\fi{} shows more prompts in the
same layout.

\input{tables/supp_schedule_ablation/supp_schedule_ablation}
\input{tables/supp_lora_window/supp_lora_window}

\subsection{Searched versus fixed schedules}
\label{sec:supp-schedule-ablation}

We replace each searched schedule of Tab.~1 of the main paper (\cref{tab:supp-schedules}) by a fixed one within the
same budget $T_{\mathrm{full}}/\rho$ and change nothing else. Both fixed schedules downsample once, from $1024^2$ to
$512^2$:
\begin{itemize}
  \item \emph{30 steps} keeps the model's 30 steps and runs as many at $1024^2$ as the budget allows: $(1,0,29)$ at
  3$\times$, the single downsampling of Tab.~5 of the main paper. It does not fit
  the 5$\times$ budget.
  \item \emph{Equal time} spends half of the budget at each resolution, \eg $(5,0,16)$ on FLUX.1-dev at 3$\times$. We
  chose it after seeing the searched schedules.
\end{itemize}
The results (\cref{tab:supp-schedule-ablation}) show that departing from the model's 30 steps brings the previews
much closer to the full images, and that searching the schedule brings them closer still, with 95\% intervals over
prompts that exclude zero.

\subsection{Video LoRA window}
\label{sec:supp-lora-window}

The video LoRA is active only on the first half of the low-resolution steps. \Cref{tab:supp-lora-window} compares this
window with no LoRA and with the LoRA on every low-resolution step, all with the same schedules. The first half is the
closest under LPIPS at every speed-up and under VJ-patch at 3$\times$ and 5$\times$. At 10$\times$, the LoRA on every
step is 0.009 VJ-patch closer.

\input{figures/supp_ablation/supp_ablation}

\section{Downsampling Time}
\label{sec:supp-n1}
\Cref{fig:supp-n1-sweep} shows FLUX.1-dev previews for a downsampling from $1024^2$ to $512^2$ after $n_1$ of 30 steps,
from $n_1=0$ to $n_1=30$, with enlarged crops, for the runs of the commutator trade-off in Sec.~6 of the main paper.

\input{figures/supp_n1_sweep/supp_n1_sweep}

\section{Qualitative Examples}
\label{sec:supp-qual}
\Cref{fig:supp-grid-flux1,fig:supp-grid-p0223,fig:supp-video-wan13-5x,fig:supp-video-wan13-10x,fig:supp-video-wan14-8x,fig:supp-video-wan14-15x} compare our previews with the
baselines and the full image or video. Many more examples, for images and video, are \ifdefined\ArxivVersion on the project page\else in \texttt{visual\_examples.html}\fi.

\input{figures/supp_grid_flux1/supp_grid_flux1}
\input{figures/supp_grid_p0223/supp_grid_p0223}
\input{figures/supp_video_wan13_5x/supp_video_wan13_5x}
\input{figures/supp_video_wan13_10x/supp_video_wan13_10x}
\input{figures/supp_video_wan14_8x/supp_video_wan14_8x}
\input{figures/supp_video_wan14_15x/supp_video_wan14_15x}

%% file: figures/supp_band_cos/supp_band_cos.tex
\begin{figure}[t]
  \centering
  \includegraphics[width=\linewidth]{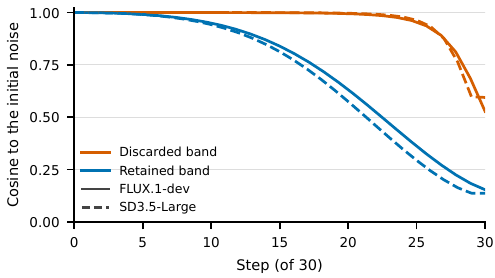}
  \caption{\textbf{Cosine similarity of the state to its initial noise.} Per band of a $1024^2\to512^2$ downsampling.}
  \label{fig:supp-band-cos}
\end{figure}

%% file: figures/supp_band_swap/supp_band_swap.tex
\begin{figure*}[t]
  \centering
  \includegraphics[width=\textwidth]{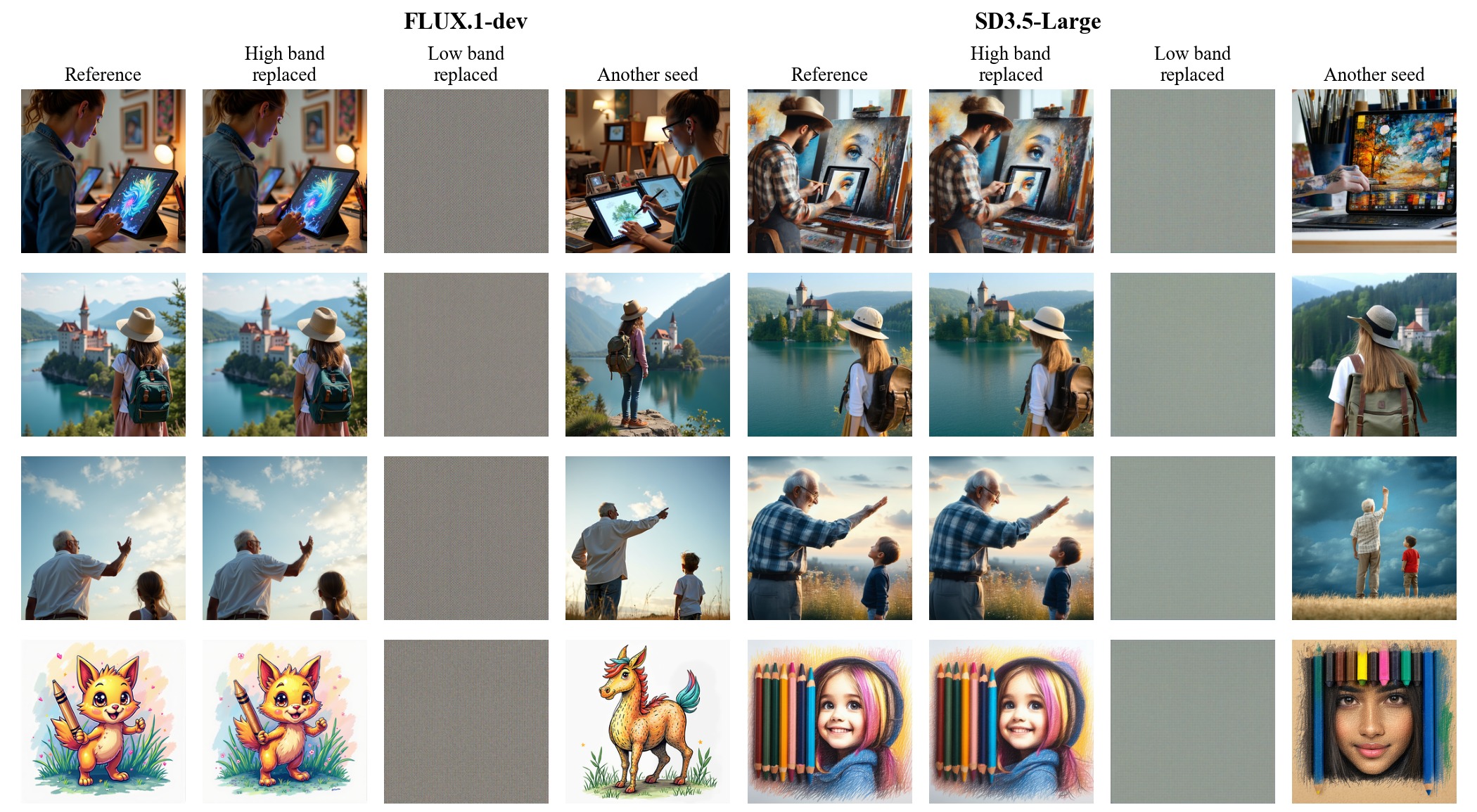}
  \caption{\textbf{Replacing a band of the state with noise.} Per model, four prompts: the final image; the final
  image when, after 15 of 30 steps, the high or low band of $z_t$ is replaced by fresh noise at the same noise level;
  and the final image of another seed.}
  \label{fig:supp-band-swap}
\end{figure*}

%% file: tables/supp_models/supp_models.tex
\begin{table}[t]
  \centering
  \small
  \setlength{\tabcolsep}{2pt}
  \begin{tabular}{@{}l cccc@{}}
    \toprule
    & \makecell{FLUX.1-\\dev} & \makecell{SD3.5-\\Large} & \makecell{Wan2.1-\\1.3B} & \makecell{Wan2.1-\\14B} \\
    \midrule
    Output       & $1024^2$ & $1024^2$ & $832\times480$ & $1248\times720$ \\
    Frames       & 1 & 1 & 81 & 81 \\
    Steps        & 30 & 30 & 50 & 50 \\
    Guidance     & 3.5 & 3.5 & 6 & 5 \\
    Time shift   & dynamic & 3 & 8 & 5 \\
    \midrule
    Prompt set   & \multicolumn{2}{c}{PixArt-30K} & \multicolumn{2}{c}{MovieGen Bench} \\
    Test         & 500 & 500 & 300 & 100 \\
    \midrule
    \makecell[l]{Searched\\resolutions} & \multicolumn{2}{c}{\makecell{$1024^2$, $768^2$, $512^2$,\\$384^2$, $256^2$}}
                 & \makecell{480p, 384p,\\320p, 240p} & \makecell{720p, 480p,\\then 1.3B\\at 480p--240p} \\
    \midrule
    Loss         & path & path & commutator & -- \\
    LoRA rank    & 16 & 32 & 32 & -- \\
    \makecell[l]{Training\\examples} & 1{,}024 & 1{,}024 & 800 & -- \\
    Learning rate & $10^{-4}$ & $10^{-4}$ & $10^{-4}$ & -- \\
    Batch        & 4 & 4 & 16 & -- \\
    Updates      & 600 & 1{,}000 & 6{,}200 & -- \\
    \bottomrule
  \end{tabular}
  \caption{Models, prompts, search and fine-tuning. Search, training and test prompts are disjoint.}
  \label{tab:supp-models}
\end{table}

%% file: figures/supp_respacing/supp_respacing.tex
\begin{figure}[t]
  \centering
  \definecolor{cfull}{HTML}{4C72B0}
  \definecolor{clow}{HTML}{DD8452}
  \ifdefined\tikzexternaldisable\tikzexternaldisable\fi
  \begin{tikzpicture}[x=\linewidth,y=1cm,font=\scriptsize]
    \def\xL{0.13}\def\xW{0.8}
    \pgfmathsetmacro{\aH}{exp(1.15)}\pgfmathsetmacro{\aL}{exp(0.63)}
    \pgfmathsetmacro{\sone}{1-10/30}
    \pgfmathsetmacro{\tone}{\aH*\sone/(1+(\aH-1)*\sone)}
    \pgfmathsetmacro{\ttau}{\tone/(2-\tone)}
    \pgfmathsetmacro{\stau}{\ttau/(\aL-(\aL-1)*\ttau)}
    \foreach \j in {0,...,29}{
      \pgfmathsetmacro{\ua}{1-\j/30}\pgfmathsetmacro{\ub}{1-(\j+1)/30}
      \ifodd\j \def\bcol{white}\else\ifnum\j<10 \def\bcol{cfull!35}\else\def\bcol{black!15}\fi\fi
      \fill[\bcol] (\xL+\xW-\ua*\xW,2.1) rectangle (\xL+\xW-\ub*\xW,2.24);}
    \draw[black!40] (\xL,2.24) rectangle (\xL+\xW,2.1);
    \draw[black!40] (\xL,1.45) -- (\xL+\xW,1.45);
    \foreach \j in {0,...,30}{
      \pgfmathsetmacro{\sv}{1-\j/30}
      \pgfmathsetmacro{\tv}{\aH*\sv/(1+(\aH-1)*\sv)}
      \ifnum\j<11 \def\col{cfull}\else\def\col{black!20}\fi
      \draw[\col,line width=0.4pt] (\xL+\xW-\sv*\xW,2.1) -- (\xL+\xW-\tv*\xW,1.45);}
    \node[anchor=east] at (\xL-0.01,2.17) {$u$};
    \node[anchor=east] at (\xL-0.01,1.45) {$t$};
    \node[anchor=east,text=cfull] at (\xL-0.04,1.775) {$1024^2$};
    \draw[clow,line width=0.8pt,-{Latex[length=3pt,width=3pt]}] (\xL+\xW-\tone*\xW,1.45) -- (\xL+\xW-\ttau*\xW,0.8)
      node[midway,right,text=black] {$\downarrow$, $t\to\tau$};
    \draw[black!40] (\xL,0.8) -- (\xL+\xW,0.8);
    \foreach \j in {0,...,7}{
      \pgfmathsetmacro{\ua}{\stau*(1-\j/8)}\pgfmathsetmacro{\ub}{\stau*(1-(\j+1)/8)}
      \ifodd\j \def\bcol{white}\else\def\bcol{clow!35}\fi
      \fill[\bcol] (\xL+\xW-\ua*\xW,0.01) rectangle (\xL+\xW-\ub*\xW,0.15);}
    \draw[black!40] (\xL,0.01) rectangle (\xL+\xW,0.15);
    \foreach \j in {0,...,8}{
      \pgfmathsetmacro{\sv}{\stau*(1-\j/8)}
      \pgfmathsetmacro{\tv}{\aL*\sv/(1+(\aL-1)*\sv)}
      \draw[clow,line width=0.4pt] (\xL+\xW-\sv*\xW,0.15) -- (\xL+\xW-\tv*\xW,0.8);}
    \node[anchor=east] at (\xL-0.01,0.8) {$t$};
    \node[anchor=east] at (\xL-0.01,0.08) {$u$};
    \node[anchor=east,text=clow] at (\xL-0.04,0.475) {$512^2$};
    \node[anchor=north] at (\xL,-0.02) {1};
    \node[anchor=north] at (\xL+\xW,-0.02) {0};
  \end{tikzpicture}
  \ifdefined\tikzexternalenable\tikzexternalenable\fi
  \caption{\textbf{Respacing after a downsampling} (FLUX.1-dev, 10 of 30 steps at $1024^2$, then 8 at $512^2$). Each
  resolution spaces its steps evenly in its unshifted time $u$ (shaded bands of equal width) and maps them to $t$
  through its own timestep shift. After the downsampling, the 8 steps are evenly spaced in $u$ between $\tau$ and 0.}
  \label{fig:supp-respacing}
\end{figure}
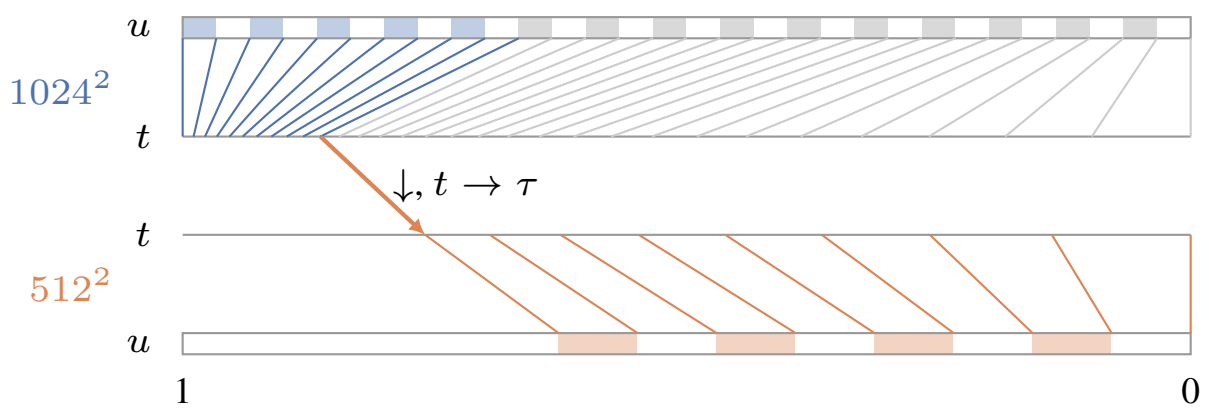

%% file: tables/supp_schedules/supp_schedules.tex
\begin{table}[t]
  \centering
  \small
  \setlength{\tabcolsep}{3pt}
  \begin{tabular}{@{}l cccc@{}}
    \toprule
    & \makecell{FLUX.1-\\dev} & \makecell{SD3.5-\\Large} & \makecell{Wan2.1-\\1.3B} & \makecell{Wan2.1-14B\\to 1.3B} \\
    & \makecell{\scriptsize $1024^2$, $768^2$,\\[-1pt] \scriptsize $512^2$} & \makecell{\scriptsize $1024^2$, $768^2$,\\[-1pt] \scriptsize $512^2$}
      & \makecell{\scriptsize 480p, 384p,\\[-1pt] \scriptsize 320p, 240p} & \makecell{\scriptsize 720p, 480p,\\[-1pt] \scriptsize 480p} \\
    \midrule
    3$\times$   & (5, 8, 0) & (7, 5, 0) & (13, 6, 0, 0) & -- \\
    4$\times$   & (4, 5, 0) & (5, 4, 0) & (9, 5, 0, 0)  & -- \\
    5$\times$   & (3, 1, 8) & (3, 3, 4) & (6, 3, 6, 0)  & -- \\
    6$\times$   & (3, 0, 6) & (2, 2, 6) & (6, 1, 4, 0)  & -- \\
    8$\times$   & (1, 0, 9) & (1, 3, 3) & (4, 2, 0, 6)  & (3, 10, 7) \\
    10$\times$  & (1, 0, 6) & (0, 3, 4) & (3, 1, 1, 6)  & -- \\
    15$\times$  & --        & --        & --            & (1, 4, 19) \\
    \bottomrule
  \end{tabular}
  \caption{Schedules: the steps $(n_1,n_2,\dots)$ at the resolutions under each model, per target speed-up. The
  hand-over runs Wan2.1-14B at 720p and 480p, then Wan2.1-1.3B at 480p.}
  \label{tab:supp-schedules}
\end{table}

%% file: figures/supp_order/supp_order.tex
\begin{figure*}[t]
  \centering
  \includegraphics[width=\textwidth]{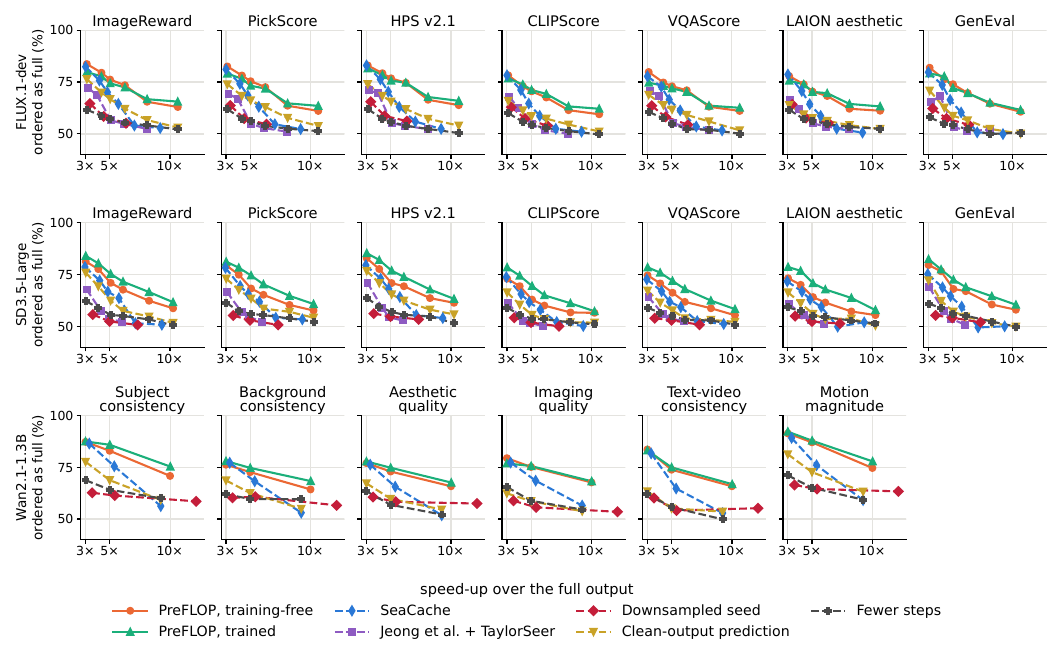}
  \caption{\textbf{Seed order per scorer.} Percentage of seed pairs of a prompt that a preview orders as their full
  outputs are ordered under the same scorer (50 is chance), against speed-up, for each image scorer and video measure.}
  \label{fig:supp-order}
\end{figure*}

%% file: figures/supp_human/supp_human.tex
\begin{figure*}[t]
  \centering
  \includegraphics[width=0.9\textwidth]{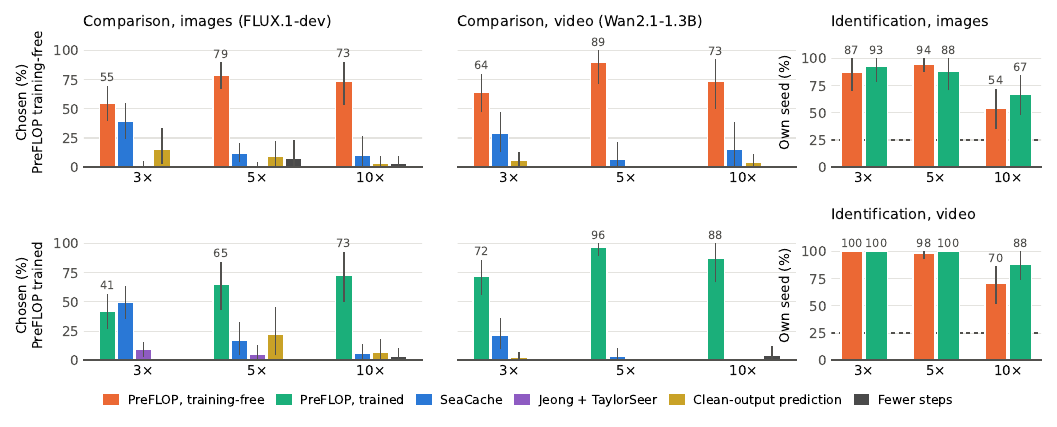}
  \caption{\textbf{Human evaluation.} Comparison: share of trials (\%) in which each look was chosen as closest to the
  full output, over the trials that showed it. Identification: share of trials in which raters matched our preview to
  its full output among four seeds (dashed: chance).}
  \label{fig:supp-human}
\end{figure*}

%% file: tables/supp_schedule_ablation/supp_schedule_ablation.tex
\begin{table}[!htb]
  \centering
  \small
  \setlength{\tabcolsep}{2.5pt}
  \begin{tabular}{@{}cl ccccc@{}}
    \toprule
    & Schedule & \makecell{Speed\\$\uparrow$} & \makecell{Image\\Reward\\$\uparrow$} & \makecell{DreamSim\\$\downarrow$}
      & \makecell{DiffSim\\$\uparrow$} & \makecell{LPIPS\\$\downarrow$} \\
    \midrule
    \multicolumn{7}{@{}l}{\emph{FLUX.1-dev}} \\
    3$\times$ & Searched & 3.11$\times$ & \textbf{0.87} & \textbf{0.049} & \textbf{0.945} & \textbf{0.205} \\
    & 30 steps & 3.11$\times$ & 0.86 & 0.120 & 0.871 & 0.355 \\
    & Equal time & 3.08$\times$ & 0.86 & 0.071 & 0.925 & 0.258 \\
    5$\times$ & Searched & 5.04$\times$ & \textbf{0.85} & \textbf{0.089} & \textbf{0.906} & \textbf{0.296} \\
    & Equal time & 5.01$\times$ & \textbf{0.85} & 0.096 & 0.901 & 0.307 \\
    \midrule
    \multicolumn{7}{@{}l}{\emph{SD3.5-Large}} \\
    3$\times$ & Searched & 3.02$\times$ & \textbf{0.77} & \textbf{0.069} & \textbf{0.923} & \textbf{0.246} \\
    & 30 steps & 3.17$\times$ & 0.55 & 0.240 & 0.715 & 0.564 \\
    & Equal time & 3.02$\times$ & 0.74 & 0.140 & 0.830 & 0.397 \\
    5$\times$ & Searched & 5.08$\times$ & \textbf{0.66} & \textbf{0.135} & \textbf{0.827} & \textbf{0.410} \\
    & Equal time & 5.06$\times$ & 0.65 & 0.183 & 0.777 & 0.470 \\
    \bottomrule
  \end{tabular}
  \caption{\textbf{Searched versus fixed schedules.} Scored at $512^2$ on the 500 held-out prompts of Tab.~1 of the
  main paper.}
  \label{tab:supp-schedule-ablation}
\end{table}

%% file: tables/supp_lora_window/supp_lora_window.tex
\begin{table}[!hb]
  \centering
  \footnotesize
  \setlength{\tabcolsep}{2pt}
  \begin{tabular}{@{}l cc cc cc@{}}
    \toprule
    & \multicolumn{2}{c}{3$\times$} & \multicolumn{2}{c}{5$\times$} & \multicolumn{2}{c}{10$\times$} \\
    \cmidrule(lr){2-3} \cmidrule(lr){4-5} \cmidrule(l){6-7}
    LoRA on & \makecell{VJ-patch\\$\uparrow$} & \makecell{LPIPS\\$\downarrow$} & \makecell{VJ-patch\\$\uparrow$}
      & \makecell{LPIPS\\$\downarrow$} & \makecell{VJ-patch\\$\uparrow$} & \makecell{LPIPS\\$\downarrow$} \\
    \midrule
    None & 0.598 & 0.162 & 0.533 & 0.257 & 0.467 & 0.383 \\
    First half & \textbf{0.605} & \textbf{0.161} & \textbf{0.557} & \textbf{0.221} & 0.492 & \textbf{0.335} \\
    All & 0.597 & 0.194 & 0.552 & 0.255 & \textbf{0.501} & 0.371 \\
    \bottomrule
  \end{tabular}
  \caption{\textbf{LoRA window of the video previews.} Wan2.1-1.3B, 32 held-out videos, LoRA on none, the first half
  (ours) or all of the low-resolution steps.}
  \label{tab:supp-lora-window}
\end{table}

%% file: figures/supp_ablation/supp_ablation.tex
\input{figures/supp_ablation/labels}
\begin{figure*}[t]
  \centering
  \newcommand{\gpanelw}{0.148}%
  \newcommand{\grefw}{0.2}%
  \newcommand{\gcolsep}{0.8pt}%
  \newcommand{\grefsep}{6pt}%
  \newcommand{\gheadsep}{1.5pt}%
  \newcommand{\gpromptsep}{1.5pt}%
  \newcommand{\gstatsep}{0.3pt}%
  \newcommand{\growsep}{1.5pt}%
  \newcommand{\gblocksep}{9pt}%
  \setlength{\tabcolsep}{\gcolsep}\renewcommand{\arraystretch}{0}%
  \newcommand{\ghead}[2]{\vtop{\hbox to \gpanelw\linewidth{\hss\strut#1\hss}\hbox to \gpanelw\linewidth{\hss\strut#2\hss}}}%
  \newcommand{\glab}[1]{\rotatebox{90}{\hbox to \gpanelw\linewidth{\hss\strut#1\hss}}}%
  \newcommand{\gtile}[2]{\vtop{\hbox{\includegraphics[width=\gpanelw\linewidth]{#1}}
    \vspace{\gstatsep}\hbox to \gpanelw\linewidth{\hss\strut#2\hss}}}%
  \newcommand{\gcell}[2]{\gtile{figures/supp_ablation/#1/#2.jpg}{\csname sa@#1@#2\endcsname}}%
  \newcommand{\gref}[1]{\makebox[\grefw\linewidth]{\smash{\raisebox{\dimexpr\gpanelw\linewidth-\height\relax}{\includegraphics[width=\grefw\linewidth]{figures/supp_ablation/#1/ref.jpg}}}}}%
  \newcommand{\gprompt}[1]{\multicolumn{7}{@{}l@{}}{\parbox[b]{\dimexpr\grefw\linewidth+5\dimexpr\gpanelw\linewidth\relax+0.03\linewidth\relax}{\gridprompt{\csname sa@#1@prompt\endcsname}}} \\[\gpromptsep]}%
  \newcommand{\gblock}[1]{\gprompt{#1}
    \gref{#1} & \glab{$n_1{=}3$} & \gcell{#1}{D3} & \gcell{#1}{A3} & \gcell{#1}{Dr3} & \gcell{#1}{C3} & \gcell{#1}{R3} \\[\growsep]
     & \glab{$n_1{=}6$} & \gcell{#1}{D6} & \gcell{#1}{A6} & \gcell{#1}{Dr6} & \gcell{#1}{C6} & \gcell{#1}{R6} \\}%
  {\scriptsize
  \begin{tabular}{@{}c@{\hspace{\grefsep}}c@{\hspace{2pt}}ccccc@{}}
    \hbox to \grefw\linewidth{\hss\strut Full image\hss} & & \ghead{DCT, $\tau$}{(PreFLOP)} & \ghead{Mean-pooling,}{$\tau$}
      & \ghead{DCT,}{$t$} & \ghead{Min.-commutator}{pixel, $t$} & \ghead{Random}{pixel, $t$} \\[\gheadsep]
    \gblock{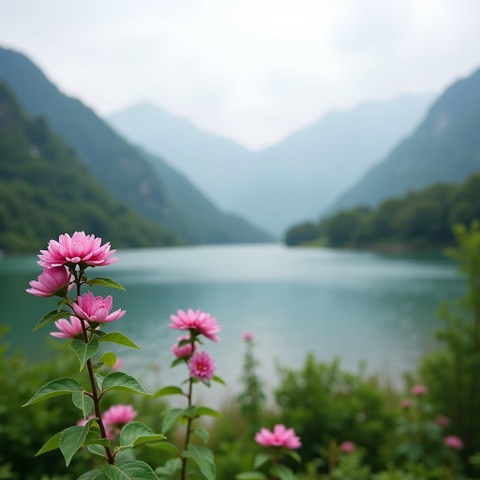}
  \end{tabular}}
  \caption{\textbf{Downsampling operators after $n_1=3$ and $n_1=6$ full-resolution steps.} FLUX.1-dev previews at
  $512^2$ that hand over once and finish with the model's own $512^2$ steps, one column per operator of the main
  paper's downsampling ablation, for one held-out prompt. The number under a tile is its DreamSim distance to the
  full image.}
  \label{fig:supp-ablation}
\end{figure*}

%% file: figures/supp_ablation/labels.tex
\expandafter\def\csname sa@p0483@D3\endcsname{0.056}
\expandafter\def\csname sa@p0483@A3\endcsname{0.053}
\expandafter\def\csname sa@p0483@Dr3\endcsname{0.047}
\expandafter\def\csname sa@p0483@C3\endcsname{0.077}
\expandafter\def\csname sa@p0483@R3\endcsname{0.145}
\expandafter\def\csname sa@p0483@D6\endcsname{0.048}
\expandafter\def\csname sa@p0483@A6\endcsname{0.051}
\expandafter\def\csname sa@p0483@Dr6\endcsname{0.044}
\expandafter\def\csname sa@p0483@C6\endcsname{0.045}
\expandafter\def\csname sa@p0483@R6\endcsname{0.056}
\expandafter\def\csname sa@p0483@prompt\endcsname{In the image, there are two women standing against a gray wall. The woman on the left is dressed in a beige coat and a green hijab, while the woman \ldots}
\expandafter\def\csname sa@p0484@D3\endcsname{0.066}
\expandafter\def\csname sa@p0484@A3\endcsname{0.074}
\expandafter\def\csname sa@p0484@Dr3\endcsname{0.060}
\expandafter\def\csname sa@p0484@C3\endcsname{0.144}
\expandafter\def\csname sa@p0484@R3\endcsname{0.162}
\expandafter\def\csname sa@p0484@D6\endcsname{0.076}
\expandafter\def\csname sa@p0484@A6\endcsname{0.081}
\expandafter\def\csname sa@p0484@Dr6\endcsname{0.088}
\expandafter\def\csname sa@p0484@C6\endcsname{0.068}
\expandafter\def\csname sa@p0484@R6\endcsname{0.119}
\expandafter\def\csname sa@p0484@prompt\endcsname{Lin Daiyu is a classical beauty standing next to a rice paper table, writing poetry with a brush in hand, Apricot eyes slightly covered, cheeks with \ldots}
\expandafter\def\csname sa@p0487@D3\endcsname{0.040}
\expandafter\def\csname sa@p0487@A3\endcsname{0.039}
\expandafter\def\csname sa@p0487@Dr3\endcsname{0.047}
\expandafter\def\csname sa@p0487@C3\endcsname{0.121}
\expandafter\def\csname sa@p0487@R3\endcsname{0.132}
\expandafter\def\csname sa@p0487@D6\endcsname{0.033}
\expandafter\def\csname sa@p0487@A6\endcsname{0.038}
\expandafter\def\csname sa@p0487@Dr6\endcsname{0.040}
\expandafter\def\csname sa@p0487@C6\endcsname{0.071}
\expandafter\def\csname sa@p0487@R6\endcsname{0.075}
\expandafter\def\csname sa@p0487@prompt\endcsname{The image captures a serene scene of nature. In the foreground, a cluster of pink flowers with green leaves is prominently featured. The flowers are \ldots}
\expandafter\def\csname sa@p0488@D3\endcsname{0.088}
\expandafter\def\csname sa@p0488@A3\endcsname{0.088}
\expandafter\def\csname sa@p0488@Dr3\endcsname{0.096}
\expandafter\def\csname sa@p0488@C3\endcsname{0.308}
\expandafter\def\csname sa@p0488@R3\endcsname{0.404}
\expandafter\def\csname sa@p0488@D6\endcsname{0.072}
\expandafter\def\csname sa@p0488@A6\endcsname{0.089}
\expandafter\def\csname sa@p0488@Dr6\endcsname{0.071}
\expandafter\def\csname sa@p0488@C6\endcsname{0.160}
\expandafter\def\csname sa@p0488@R6\endcsname{0.160}
\expandafter\def\csname sa@p0488@prompt\endcsname{The portable mind, psychedelic, alien ships, surrealism}

%% file: figures/supp_n1_sweep/supp_n1_sweep.tex
\begin{figure*}[t]
  \centering
  \includegraphics[width=\textwidth]{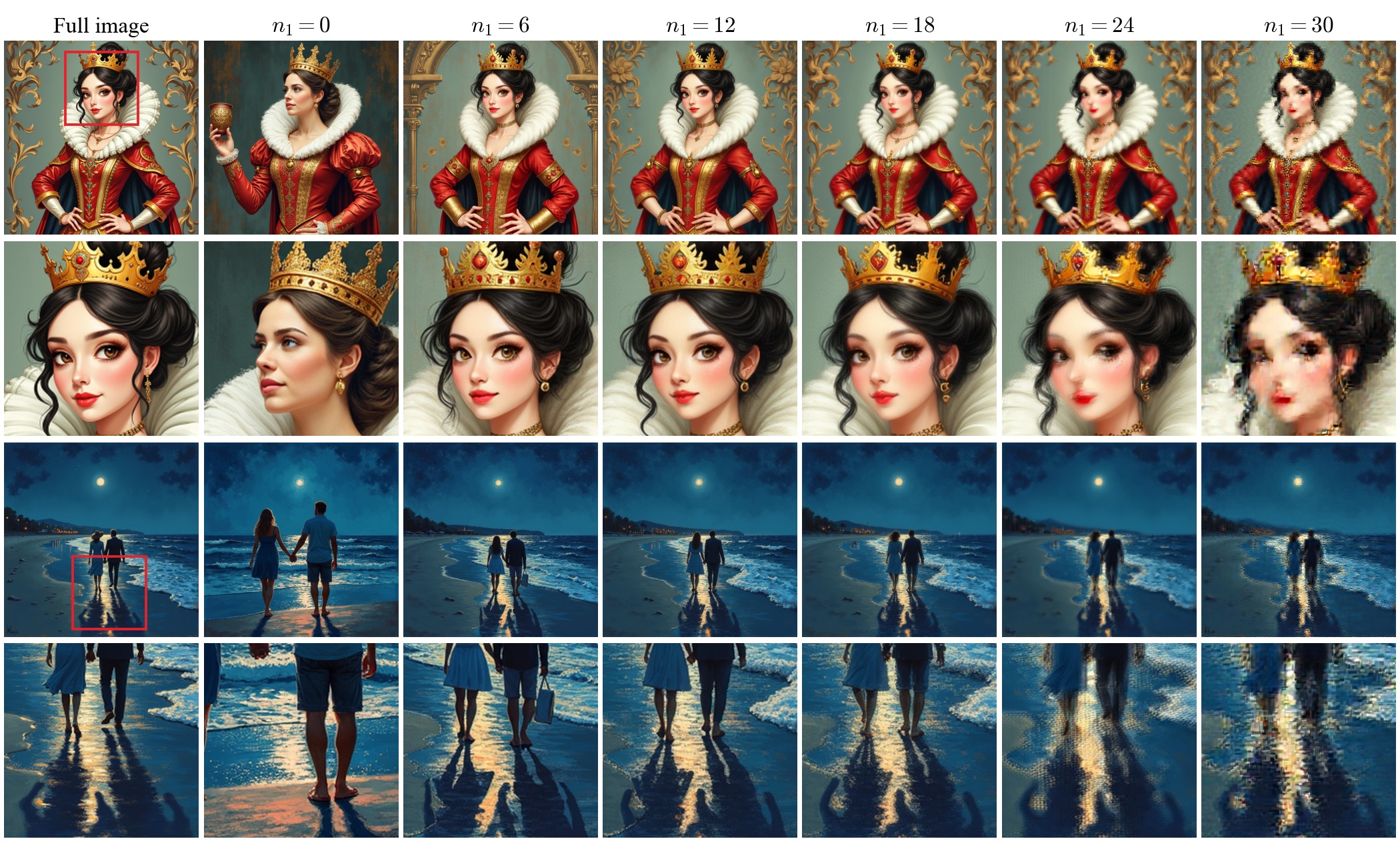}
  \caption{\textbf{Previews by downsampling time.} FLUX.1-dev, downsampled from $1024^2$ to $512^2$ after $n_1$ of 30
  steps and finished at $512^2$. At $n_1=30$, the downsampled final latent is decoded. Below each row, the region boxed
  in the full image, enlarged.}
  \label{fig:supp-n1-sweep}
\end{figure*}

%% file: figures/supp_grid_flux1/supp_grid_flux1.tex
\begin{figure*}[t]
  \centering
  \newcommand{\gpanelw}{0.112}%
  \newcommand{\grefw}{0.14}%
  \newcommand{\glabw}{0.022}%
  \newcommand{\gcolsep}{0.8pt}%
  \newcommand{\grefsep}{6pt}%
  \newcommand{\gheadsep}{1.5pt}%
  \newcommand{\gpromptsep}{1.5pt}%
  \newcommand{\growsep}{1.5pt}%
  \newcommand{\gblocksep}{4pt}%
  \setlength{\tabcolsep}{\gcolsep}\renewcommand{\arraystretch}{0}%
  \newcommand{\ghead}[2]{\vtop{\hbox to \gpanelw\linewidth{\hss\strut#1\hss}\hbox to \gpanelw\linewidth{\hss\strut#2\hss}}}%
  \newcommand{\glab}[1]{\rotatebox{90}{\hbox to \gpanelw\linewidth{\hss\strut#1\hss}}}%
  \newcommand{\gtile}[1]{\includegraphics[width=\gpanelw\linewidth]{#1}}%
  \newcommand{\gtiletext}[2]{\rlap{\gtile{#1}}\raisebox{0.47\dimexpr\gpanelw\linewidth\relax}{\hbox to \gpanelw\linewidth{\hss\strut#2\hss}}}%
  \newcommand{\gcell}[3]{\gtile{figures/supp_grid_flux1/#1/#2-#3.jpg}}%
  \newcommand{\gna}{\gtiletext{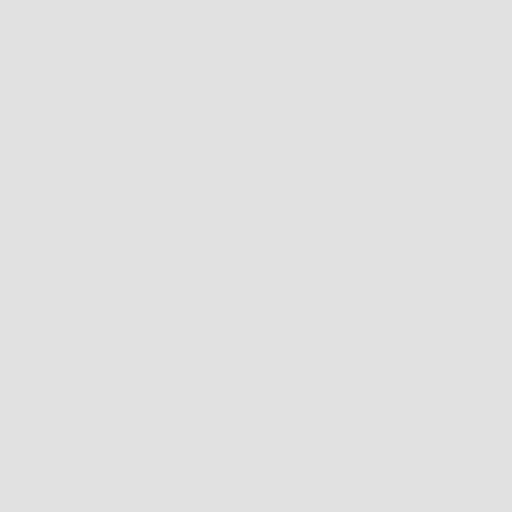}{cannot reach}}%
  \newcommand{\gref}[1]{\makebox[\grefw\linewidth]{\smash{\raisebox{\dimexpr\gpanelw\linewidth-\height\relax}{\includegraphics[width=\grefw\linewidth]{#1}}}}}%
  {\scriptsize
  \begin{tabular}{@{}c@{\hspace{\grefsep}}c@{\hspace{2pt}}ccccccc@{}}
    \hbox to \grefw\linewidth{\hss\strut Full image\hss} & & \ghead{Ours,}{trained} & \ghead{Ours,}{training-free} & \ghead{}{SeaCache} & \ghead{Jeong et al. +}{TaylorSeer} & \ghead{Downsampled}{seed} & \ghead{Fewer}{steps} & \ghead{Clean-output}{prediction} \\[\gheadsep]
    \multicolumn{9}{@{}l@{}}{\gridprompt{a photo of three giraffes}} \\[\gpromptsep]
    \gref{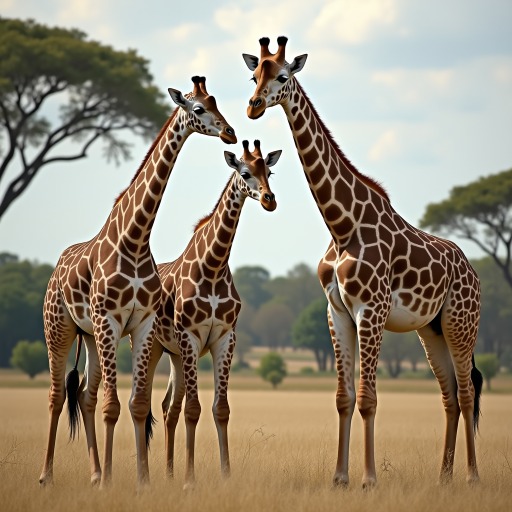} & \glab{$\approx$3$\times$} & \gcell{g00227}{trained}{3} & \gcell{g00227}{ours}{3} & \gcell{g00227}{sc}{3} & \gcell{g00227}{jts}{3} & \gcell{g00227}{k0}{3} & \gcell{g00227}{fs}{3} & \gcell{g00227}{x0}{3} \\[\growsep]
     & \glab{$\approx$5$\times$} & \gcell{g00227}{trained}{5} & \gcell{g00227}{ours}{5} & \gcell{g00227}{sc}{5} & \gcell{g00227}{jts}{5} & \gcell{g00227}{k0}{5} & \gcell{g00227}{fs}{5} & \gcell{g00227}{x0}{5} \\[\growsep]
     & \glab{$\approx$10$\times$} & \gcell{g00227}{trained}{10} & \gcell{g00227}{ours}{10} & \gcell{g00227}{sc}{10} & \gna & \gna & \gcell{g00227}{fs}{10} & \gcell{g00227}{x0}{10} \\
  \end{tabular}}
  \caption{\textbf{FLUX.1-dev previews at about 3$\times$, 5$\times$ and 10$\times$, with every image baseline.}}
  \label{fig:supp-grid-flux1}
\end{figure*}

%% file: figures/supp_grid_p0223/supp_grid_p0223.tex
\begin{figure*}[t]
  \centering
  \newcommand{\gpanelw}{0.112}%
  \newcommand{\grefw}{0.14}%
  \newcommand{\gcolsep}{0.8pt}%
  \newcommand{\grefsep}{6pt}%
  \newcommand{\gheadsep}{1.5pt}%
  \newcommand{\gpromptsep}{1.5pt}%
  \newcommand{\growsep}{1.5pt}%
  \setlength{\tabcolsep}{\gcolsep}\renewcommand{\arraystretch}{0}%
  \newcommand{\ghead}[2]{\vtop{\hbox to \gpanelw\linewidth{\hss\strut#1\hss}\hbox to \gpanelw\linewidth{\hss\strut#2\hss}}}%
  \newcommand{\glab}[1]{\rotatebox{90}{\hbox to \gpanelw\linewidth{\hss\strut#1\hss}}}%
  \newcommand{\gtile}[1]{\includegraphics[width=\gpanelw\linewidth]{#1}}%
  \newcommand{\gtiletext}[2]{\rlap{\gtile{#1}}\raisebox{0.47\dimexpr\gpanelw\linewidth\relax}{\hbox to \gpanelw\linewidth{\hss\strut#2\hss}}}%
  \newcommand{\gcell}[2]{\gtile{figures/supp_grid_p0223/#1-#2.jpg}}%
  \newcommand{\gna}{\gtiletext{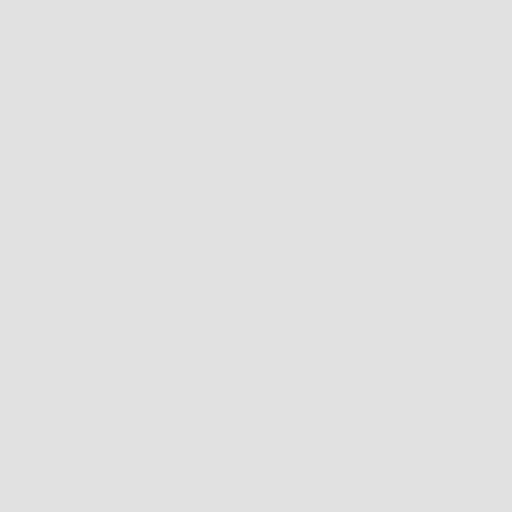}{cannot reach}}%
  \newcommand{\gref}[1]{\makebox[\grefw\linewidth]{\smash{\raisebox{\dimexpr\gpanelw\linewidth-\height\relax}{\includegraphics[width=\grefw\linewidth]{#1}}}}}%
  {\scriptsize
  \begin{tabular}{@{}c@{\hspace{\grefsep}}c@{\hspace{2pt}}ccccccc@{}}
    \hbox to \grefw\linewidth{\hss\strut Full image\hss} & & \ghead{Ours,}{trained} & \ghead{Ours,}{training-free} & \ghead{}{SeaCache} & \ghead{Jeong et al. +}{TaylorSeer} & \ghead{Downsampled}{seed} & \ghead{Fewer}{steps} & \ghead{Clean-output}{prediction} \\[\gheadsep]
    \multicolumn{9}{@{}l@{}}{\gridprompt{a whimsical illustration of a fairy riding a unicorn through a magical forest, capture photography, disney style, high detail}} \\[\gpromptsep]
    \gref{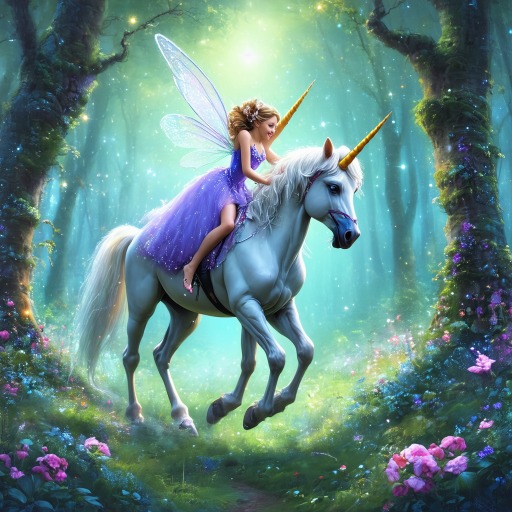} & \glab{$\approx$3$\times$} & \gcell{trained}{3} & \gcell{free}{3} & \gcell{sc}{3} & \gcell{jts}{3} & \gcell{k0}{3} & \gcell{fewer}{3} & \gcell{x0}{3} \\[\growsep]
     & \glab{$\approx$5$\times$} & \gcell{trained}{5} & \gcell{free}{5} & \gcell{sc}{5} & \gcell{jts}{5} & \gcell{k0}{5} & \gcell{fewer}{5} & \gcell{x0}{5} \\[\growsep]
     & \glab{$\approx$10$\times$} & \gcell{trained}{10} & \gcell{free}{10} & \gcell{sc}{10} & \gna & \gna & \gcell{fewer}{10} & \gcell{x0}{10} \\
  \end{tabular}}
  \caption{\textbf{SD3.5-Large previews at about 3$\times$, 5$\times$ and 10$\times$, with every image baseline.}}
  \label{fig:supp-grid-p0223}
\end{figure*}

%% file: figures/supp_video_wan13_5x/supp_video_wan13_5x.tex
\begin{figure*}[t]
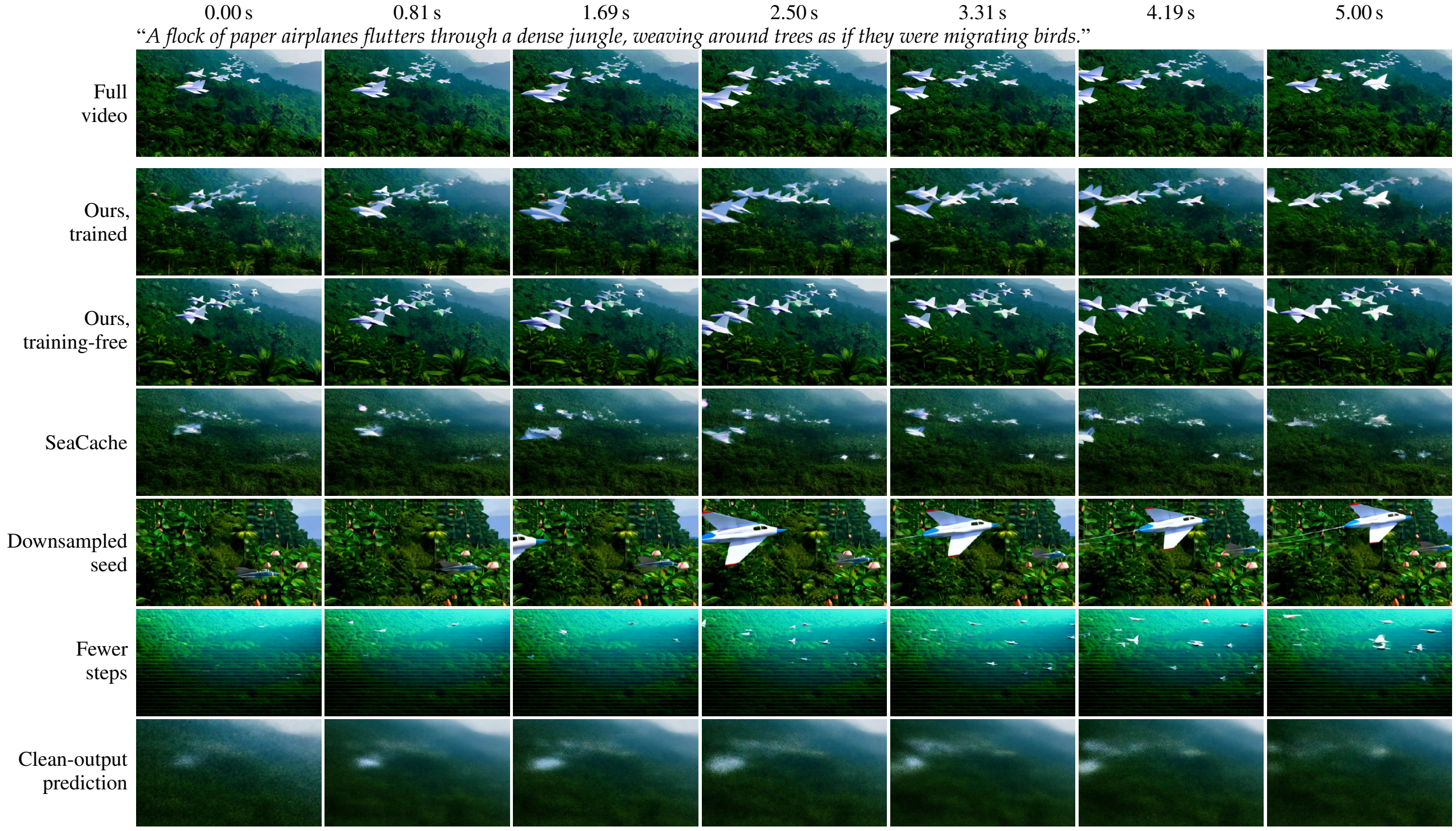

  \centering
  \newcommand{\gpanelw}{0.1245}%
  \newcommand{\glabw}{0.098}%
  \newcommand{\gcolsep}{0.6pt}%
  \newcommand{\glabsep}{3pt}%
  \newcommand{\gheadsep}{1.5pt}%
  \newcommand{\gpromptsep}{1.5pt}%
  \newcommand{\grefsep}{4pt}%
  \newcommand{\growsep}{1.2pt}%
  \setlength{\tabcolsep}{\gcolsep}\renewcommand{\arraystretch}{0}%
  \newcommand{\ghead}[1]{\hbox to \gpanelw\linewidth{\hss\strut#1\hss}}%
  \newcommand{\glab}[1]{\parbox[b][\dimexpr\gpanelw\linewidth*240/416\relax][c]{\glabw\linewidth}{\raggedleft#1}}%
  \newcommand{\gcell}[2]{\includegraphics[width=\gpanelw\linewidth]{figures/supp_video_wan13_5x/#1-f#2.jpg}}%
  {\scriptsize
  \begin{tabular}{@{}r@{\hspace{\glabsep}}ccccccc@{}}
     & \ghead{0.00\,s} & \ghead{0.81\,s} & \ghead{1.69\,s} & \ghead{2.50\,s} & \ghead{3.31\,s} & \ghead{4.19\,s} & \ghead{5.00\,s} \\[\gheadsep]
     & \multicolumn{7}{@{}l@{}}{\gridprompt{A flock of paper airplanes flutters through a dense jungle, weaving around trees as if they were migrating birds.}} \\[\gpromptsep]
    \glab{Full\\video} & \gcell{ref}{00} & \gcell{ref}{13} & \gcell{ref}{27} & \gcell{ref}{40} & \gcell{ref}{53} & \gcell{ref}{67} & \gcell{ref}{80} \\[\grefsep]
    \glab{Ours,\\trained} & \gcell{trained-5}{00} & \gcell{trained-5}{13} & \gcell{trained-5}{27} & \gcell{trained-5}{40} & \gcell{trained-5}{53} & \gcell{trained-5}{67} & \gcell{trained-5}{80} \\[\growsep]
    \glab{Ours,\\training-free} & \gcell{ours-5}{00} & \gcell{ours-5}{13} & \gcell{ours-5}{27} & \gcell{ours-5}{40} & \gcell{ours-5}{53} & \gcell{ours-5}{67} & \gcell{ours-5}{80} \\[\growsep]
    \glab{SeaCache} & \gcell{sc-5}{00} & \gcell{sc-5}{13} & \gcell{sc-5}{27} & \gcell{sc-5}{40} & \gcell{sc-5}{53} & \gcell{sc-5}{67} & \gcell{sc-5}{80} \\[\growsep]
    \glab{Downsampled\\seed} & \gcell{k0-5}{00} & \gcell{k0-5}{13} & \gcell{k0-5}{27} & \gcell{k0-5}{40} & \gcell{k0-5}{53} & \gcell{k0-5}{67} & \gcell{k0-5}{80} \\[\growsep]
    \glab{Fewer\\steps} & \gcell{fs-5}{00} & \gcell{fs-5}{13} & \gcell{fs-5}{27} & \gcell{fs-5}{40} & \gcell{fs-5}{53} & \gcell{fs-5}{67} & \gcell{fs-5}{80} \\[\growsep]
    \glab{Clean-output\\prediction} & \gcell{x0-5}{00} & \gcell{x0-5}{13} & \gcell{x0-5}{27} & \gcell{x0-5}{40} & \gcell{x0-5}{53} & \gcell{x0-5}{67} & \gcell{x0-5}{80} \\
  \end{tabular}}
  \caption{\textbf{Wan2.1-1.3B previews at about 5$\times$, with every video baseline.}}
  \label{fig:supp-video-wan13-5x}
\end{figure*}

%% file: figures/supp_video_wan13_10x/supp_video_wan13_10x.tex
\begin{figure*}[t]
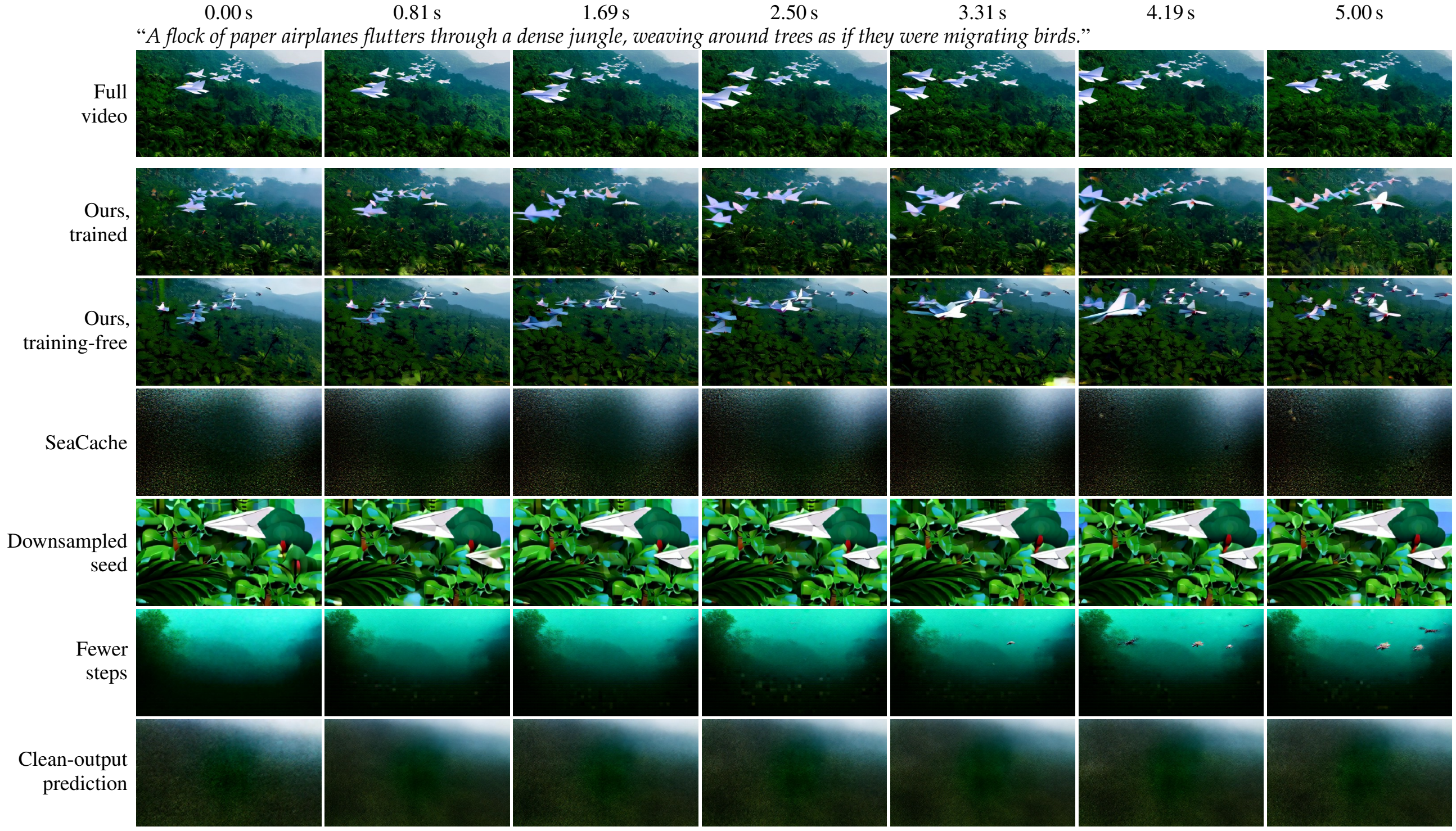

  \centering
  \newcommand{\gpanelw}{0.1245}%
  \newcommand{\glabw}{0.098}%
  \newcommand{\gcolsep}{0.6pt}%
  \newcommand{\glabsep}{3pt}%
  \newcommand{\gheadsep}{1.5pt}%
  \newcommand{\gpromptsep}{1.5pt}%
  \newcommand{\grefsep}{4pt}%
  \newcommand{\growsep}{1.2pt}%
  \setlength{\tabcolsep}{\gcolsep}\renewcommand{\arraystretch}{0}%
  \newcommand{\ghead}[1]{\hbox to \gpanelw\linewidth{\hss\strut#1\hss}}%
  \newcommand{\glab}[1]{\parbox[b][\dimexpr\gpanelw\linewidth*240/416\relax][c]{\glabw\linewidth}{\raggedleft#1}}%
  \newcommand{\gcell}[2]{\includegraphics[width=\gpanelw\linewidth]{figures/supp_video_wan13_10x/#1-f#2.jpg}}%
  {\scriptsize
  \begin{tabular}{@{}r@{\hspace{\glabsep}}ccccccc@{}}
     & \ghead{0.00\,s} & \ghead{0.81\,s} & \ghead{1.69\,s} & \ghead{2.50\,s} & \ghead{3.31\,s} & \ghead{4.19\,s} & \ghead{5.00\,s} \\[\gheadsep]
     & \multicolumn{7}{@{}l@{}}{\gridprompt{A flock of paper airplanes flutters through a dense jungle, weaving around trees as if they were migrating birds.}} \\[\gpromptsep]
    \glab{Full\\video} & \gcell{ref}{00} & \gcell{ref}{13} & \gcell{ref}{27} & \gcell{ref}{40} & \gcell{ref}{53} & \gcell{ref}{67} & \gcell{ref}{80} \\[\grefsep]
    \glab{Ours,\\trained} & \gcell{trained-10}{00} & \gcell{trained-10}{13} & \gcell{trained-10}{27} & \gcell{trained-10}{40} & \gcell{trained-10}{53} & \gcell{trained-10}{67} & \gcell{trained-10}{80} \\[\growsep]
    \glab{Ours,\\training-free} & \gcell{ours-10}{00} & \gcell{ours-10}{13} & \gcell{ours-10}{27} & \gcell{ours-10}{40} & \gcell{ours-10}{53} & \gcell{ours-10}{67} & \gcell{ours-10}{80} \\[\growsep]
    \glab{SeaCache} & \gcell{sc-10}{00} & \gcell{sc-10}{13} & \gcell{sc-10}{27} & \gcell{sc-10}{40} & \gcell{sc-10}{53} & \gcell{sc-10}{67} & \gcell{sc-10}{80} \\[\growsep]
    \glab{Downsampled\\seed} & \gcell{k0-10}{00} & \gcell{k0-10}{13} & \gcell{k0-10}{27} & \gcell{k0-10}{40} & \gcell{k0-10}{53} & \gcell{k0-10}{67} & \gcell{k0-10}{80} \\[\growsep]
    \glab{Fewer\\steps} & \gcell{fs-10}{00} & \gcell{fs-10}{13} & \gcell{fs-10}{27} & \gcell{fs-10}{40} & \gcell{fs-10}{53} & \gcell{fs-10}{67} & \gcell{fs-10}{80} \\[\growsep]
    \glab{Clean-output\\prediction} & \gcell{x0-10}{00} & \gcell{x0-10}{13} & \gcell{x0-10}{27} & \gcell{x0-10}{40} & \gcell{x0-10}{53} & \gcell{x0-10}{67} & \gcell{x0-10}{80} \\
  \end{tabular}}
  \caption{\textbf{Wan2.1-1.3B previews at about 10$\times$, with every video baseline.}}
  \label{fig:supp-video-wan13-10x}
\end{figure*}

%% file: figures/supp_video_wan14_8x/supp_video_wan14_8x.tex
\begin{figure*}[t]
  \centering
  \newcommand{\gpanelw}{0.1245}%
  \newcommand{\glabw}{0.098}%
  \newcommand{\gcolsep}{0.6pt}%
  \newcommand{\glabsep}{3pt}%
  \newcommand{\gheadsep}{1.5pt}%
  \newcommand{\gpromptsep}{1.5pt}%
  \newcommand{\grefsep}{4pt}%
  \newcommand{\growsep}{1.2pt}%
  \setlength{\tabcolsep}{\gcolsep}\renewcommand{\arraystretch}{0}%
  \newcommand{\ghead}[1]{\hbox to \gpanelw\linewidth{\hss\strut#1\hss}}%
  \newcommand{\glab}[1]{\parbox[b][\dimexpr\gpanelw\linewidth*480/832\relax][c]{\glabw\linewidth}{\raggedleft#1}}%
  \newcommand{\gcell}[2]{\includegraphics[width=\gpanelw\linewidth]{figures/supp_video_wan14_8x/#1-f#2.jpg}}%
  {\scriptsize
  \begin{tabular}{@{}r@{\hspace{\glabsep}}ccccccc@{}}
     & \ghead{0.00\,s} & \ghead{0.81\,s} & \ghead{1.69\,s} & \ghead{2.50\,s} & \ghead{3.31\,s} & \ghead{4.19\,s} & \ghead{5.00\,s} \\[\gheadsep]
     & \multicolumn{7}{@{}l@{}}{\gridprompt{Photorealistic closeup video of two pirate ships battling each other as they sail inside a cup of coffee.}} \\[\gpromptsep]
    \glab{Full\\video} & \gcell{ref}{00} & \gcell{ref}{13} & \gcell{ref}{27} & \gcell{ref}{40} & \gcell{ref}{53} & \gcell{ref}{67} & \gcell{ref}{80} \\[\grefsep]
    \glab{Ours} & \gcell{ours-8}{00} & \gcell{ours-8}{13} & \gcell{ours-8}{27} & \gcell{ours-8}{40} & \gcell{ours-8}{53} & \gcell{ours-8}{67} & \gcell{ours-8}{80} \\[\growsep]
    \glab{SeaCache} & \gcell{sc-8}{00} & \gcell{sc-8}{13} & \gcell{sc-8}{27} & \gcell{sc-8}{40} & \gcell{sc-8}{53} & \gcell{sc-8}{67} & \gcell{sc-8}{80} \\[\growsep]
    \glab{Downsampled\\seed} & \gcell{k0-8}{00} & \gcell{k0-8}{13} & \gcell{k0-8}{27} & \gcell{k0-8}{40} & \gcell{k0-8}{53} & \gcell{k0-8}{67} & \gcell{k0-8}{80} \\[\growsep]
    \glab{Fewer\\steps} & \gcell{fs-8}{00} & \gcell{fs-8}{13} & \gcell{fs-8}{27} & \gcell{fs-8}{40} & \gcell{fs-8}{53} & \gcell{fs-8}{67} & \gcell{fs-8}{80} \\[\growsep]
    \glab{Clean-output\\prediction} & \gcell{x0-8}{00} & \gcell{x0-8}{13} & \gcell{x0-8}{27} & \gcell{x0-8}{40} & \gcell{x0-8}{53} & \gcell{x0-8}{67} & \gcell{x0-8}{80} \\
  \end{tabular}}
  \caption{\textbf{Wan2.1-14B previews at about 8$\times$, with every video baseline.} Our previews hand over from Wan2.1-14B to Wan2.1-1.3B, as in the main paper's hand-over table.}
  \label{fig:supp-video-wan14-8x}
\end{figure*}

%% file: figures/supp_video_wan14_15x/supp_video_wan14_15x.tex
\begin{figure*}[t]
  \centering
  \newcommand{\gpanelw}{0.1245}%
  \newcommand{\glabw}{0.098}%
  \newcommand{\gcolsep}{0.6pt}%
  \newcommand{\glabsep}{3pt}%
  \newcommand{\gheadsep}{1.5pt}%
  \newcommand{\gpromptsep}{1.5pt}%
  \newcommand{\grefsep}{4pt}%
  \newcommand{\growsep}{1.2pt}%
  \setlength{\tabcolsep}{\gcolsep}\renewcommand{\arraystretch}{0}%
  \newcommand{\ghead}[1]{\hbox to \gpanelw\linewidth{\hss\strut#1\hss}}%
  \newcommand{\glab}[1]{\parbox[b][\dimexpr\gpanelw\linewidth*480/832\relax][c]{\glabw\linewidth}{\raggedleft#1}}%
  \newcommand{\gcell}[2]{\includegraphics[width=\gpanelw\linewidth]{figures/supp_video_wan14_15x/#1-f#2.jpg}}%
  {\scriptsize
  \begin{tabular}{@{}r@{\hspace{\glabsep}}ccccccc@{}}
     & \ghead{0.00\,s} & \ghead{0.81\,s} & \ghead{1.69\,s} & \ghead{2.50\,s} & \ghead{3.31\,s} & \ghead{4.19\,s} & \ghead{5.00\,s} \\[\gheadsep]
     & \multicolumn{7}{@{}l@{}}{\gridprompt{Photorealistic closeup video of two pirate ships battling each other as they sail inside a cup of coffee.}} \\[\gpromptsep]
    \glab{Full\\video} & \gcell{ref}{00} & \gcell{ref}{13} & \gcell{ref}{27} & \gcell{ref}{40} & \gcell{ref}{53} & \gcell{ref}{67} & \gcell{ref}{80} \\[\grefsep]
    \glab{Ours} & \gcell{ours-15}{00} & \gcell{ours-15}{13} & \gcell{ours-15}{27} & \gcell{ours-15}{40} & \gcell{ours-15}{53} & \gcell{ours-15}{67} & \gcell{ours-15}{80} \\[\growsep]
    \glab{SeaCache} & \gcell{sc-15}{00} & \gcell{sc-15}{13} & \gcell{sc-15}{27} & \gcell{sc-15}{40} & \gcell{sc-15}{53} & \gcell{sc-15}{67} & \gcell{sc-15}{80} \\[\growsep]
    \glab{Downsampled\\seed} & \gcell{k0-15}{00} & \gcell{k0-15}{13} & \gcell{k0-15}{27} & \gcell{k0-15}{40} & \gcell{k0-15}{53} & \gcell{k0-15}{67} & \gcell{k0-15}{80} \\[\growsep]
    \glab{Fewer\\steps} & \gcell{fs-15}{00} & \gcell{fs-15}{13} & \gcell{fs-15}{27} & \gcell{fs-15}{40} & \gcell{fs-15}{53} & \gcell{fs-15}{67} & \gcell{fs-15}{80} \\[\growsep]
    \glab{Clean-output\\prediction} & \gcell{x0-15}{00} & \gcell{x0-15}{13} & \gcell{x0-15}{27} & \gcell{x0-15}{40} & \gcell{x0-15}{53} & \gcell{x0-15}{67} & \gcell{x0-15}{80} \\
  \end{tabular}}
  \caption{\textbf{Wan2.1-14B previews at about 15$\times$, with every video baseline.} Our previews hand over from Wan2.1-14B to Wan2.1-1.3B, as in the main paper's hand-over table.}
  \label{fig:supp-video-wan14-15x}
\end{figure*}

%% file: arxiv.bbl
\begin{thebibliography}{66}
\providecommand{\natexlab}[1]{#1}
\providecommand{\url}[1]{\texttt{#1}}
\expandafter\ifx\csname urlstyle\endcsname\relax
  \providecommand{\doi}[1]{doi: #1}\else
  \providecommand{\doi}{doi: \begingroup \urlstyle{rm}\Url}\fi

\bibitem[Ahmed et~al.(1974)Ahmed, Natarajan, and Rao]{ahmed1974dct}
Nasir Ahmed, T. Natarajan, and Kamisetty~R. Rao.
\newblock {Discrete Cosine Transform}.
\newblock \emph{IEEE Transactions on Computers}, C-23\penalty0 (1):\penalty0
  90--93, 1974.

\bibitem[Ahn et~al.(2024)Ahn, Cho, Min, Jang, Kim, Kim, Park, Jin, and
  Kim]{ahn2024pag}
Donghoon Ahn, Hyoungwon Cho, Jaewon Min, Wooseok Jang, Jungwoo Kim, SeonHwa
  Kim, Hyun~Hee Park, Kyong~Hwan Jin, and Seungryong Kim.
\newblock {Self-Rectifying Diffusion Sampling with Perturbed-Attention
  Guidance}.
\newblock In \emph{European Conference on Computer Vision (ECCV)}, 2024.

\bibitem[Assran et~al.(2025)Assran, Bardes, Fan, Garrido, Howes, Komeili,
  Muckley, Rizvi, Roberts, Sinha, Zholus, Arnaud, Gejji, Martin, Robert~Hogan,
  Dugas, Bojanowski, Khalidov, Labatut, Massa, Szafraniec, Krishnakumar, Li,
  Ma, Chandar, Meier, LeCun, Rabbat, and Ballas]{assran2025vjepa2}
Mido Assran, Adrien Bardes, David Fan, Quentin Garrido, Russell Howes, Mojtaba
  Komeili, Matthew Muckley, Ammar Rizvi, Claire Roberts, Koustuv Sinha, Artem
  Zholus, Sergio Arnaud, Abha Gejji, Ada Martin, Francois Robert~Hogan, Daniel
  Dugas, Piotr Bojanowski, Vasil Khalidov, Patrick Labatut, Francisco Massa,
  Marc Szafraniec, Kapil Krishnakumar, Yong Li, Xiaodong Ma, Sarath Chandar,
  Franziska Meier, Yann LeCun, Michael Rabbat, and Nicolas Ballas.
\newblock {V-JEPA 2: Self-Supervised Video Models Enable Understanding,
  Prediction and Planning}.
\newblock arXiv preprint arXiv:2506.09985, 2025.

\bibitem[{Black Forest Labs}(2024)]{bfl2024flux}
{Black Forest Labs}.
\newblock {FLUX.1}.
\newblock \url{https://github.com/black-forest-labs/flux}, 2024.

\bibitem[Chen et~al.(2024)Chen, Ge, Xie, Wu, Yao, Ren, Wang, Luo, Lu, and
  Li]{chen2024pixartsigma}
Junsong Chen, Chongjian Ge, Enze Xie, Yue Wu, Lewei Yao, Xiaozhe Ren, Zhongdao
  Wang, Ping Luo, Huchuan Lu, and Zhenguo Li.
\newblock {PixArt-$\Sigma$: Weak-to-Strong Training of Diffusion Transformer
  for 4K Text-to-Image Generation}.
\newblock In \emph{European Conference on Computer Vision (ECCV)}, pages
  74--91, 2024.

\bibitem[Cheng et~al.(2025)Cheng, Wei, Diao, Liu, Chen, Huang, Liu, Yu, Du,
  Lin, and You]{cheng2025srdiffusion}
Shenggan Cheng, Yuanxin Wei, Lansong Diao, Yong Liu, Bujiao Chen, Lianghua
  Huang, Yu Liu, Wenyuan Yu, Jiangsu Du, Wei Lin, and Yang You.
\newblock {SRDiffusion: Accelerate Video Diffusion Inference via
  Sketching-Rendering Cooperation}.
\newblock arXiv preprint arXiv:2505.19151, 2025.

\bibitem[Chung et~al.(2026)Chung, Hyun, Lee, Han, Cha, Wee, Hong, and
  Heo]{chung2026seacache}
Jiwoo Chung, Sangeek Hyun, MinKyu Lee, Byeongju Han, Geonho Cha, Dongyoon Wee,
  Youngjun Hong, and Jae-Pil Heo.
\newblock {SeaCache: Spectral-Evolution-Aware Cache for Accelerating Diffusion
  Models}.
\newblock In \emph{IEEE/CVF Conference on Computer Vision and Pattern
  Recognition (CVPR)}, pages 14283--14294, 2026.

\bibitem[Cui et~al.(2026)Cui, Huang, Ye, Dong, Chen, Xue, Yang, Huang, Tang,
  and Hong]{cui2026diffusionprobe}
Benlei Cui, Bukun Huang, Zhizeng Ye, Xuemei Dong, Tuo Chen, Hui Xue, Dingkang
  Yang, Longtao Huang, Jingqun Tang, and Haiwen Hong.
\newblock {Diffusion Probe: Generated Image Result Prediction Using CNN
  Probes}.
\newblock In \emph{IEEE/CVF Conference on Computer Vision and Pattern
  Recognition (CVPR)}, pages 35926--35935, 2026.

\bibitem[Esser et~al.(2024)Esser, Kulal, Blattmann, Entezari, M{\"u}ller,
  Saini, Levi, Lorenz, Sauer, Boesel, et~al.]{esser2024sd3}
Patrick Esser, Sumith Kulal, Andreas Blattmann, Rahim Entezari, Jonas
  M{\"u}ller, Harry Saini, Yam Levi, Dominik Lorenz, Axel Sauer, Frederic
  Boesel, et~al.
\newblock {Scaling Rectified Flow Transformers for High-Resolution Image
  Synthesis}.
\newblock In \emph{International Conference on Machine Learning (ICML)}, pages
  12606--12633. PMLR, 2024.

\bibitem[Fan et~al.(2025)Fan, Zheng, Yeh, and Liu]{fan2025cfgzero}
Weichen Fan, Amber~Yijia Zheng, Raymond~A. Yeh, and Ziwei Liu.
\newblock {CFG-Zero*: Improved Classifier-Free Guidance for Flow Matching
  Models}.
\newblock arXiv preprint arXiv:2503.18886, 2025.

\bibitem[Fu et~al.(2023)Fu, Tamir, Sundaram, Chai, Zhang, Dekel, and
  Isola]{fu2023dreamsim}
Stephanie Fu, Netanel Tamir, Shobhita Sundaram, Lucy Chai, Richard Zhang, Tali
  Dekel, and Phillip Isola.
\newblock {DreamSim: Learning New Dimensions of Human Visual Similarity using
  Synthetic Data}.
\newblock In \emph{Advances in Neural Information Processing Systems
  (NeurIPS)}, pages 50742--50768, 2023.

\bibitem[Ghosh et~al.(2023)Ghosh, Hajishirzi, and Schmidt]{ghosh2023geneval}
Dhruba Ghosh, Hannaneh Hajishirzi, and Ludwig Schmidt.
\newblock {GenEval: An Object-Focused Framework for Evaluating Text-to-Image
  Alignment}.
\newblock In \emph{Advances in Neural Information Processing Systems (NeurIPS),
  Datasets and Benchmarks Track}, 2023.

\bibitem[Guimar{\~a}es and Perona(2026)]{guimaraes2026psp}
Rog{\'e}rio Guimar{\~a}es and Pietro Perona.
\newblock {Inference-Time Scaling of Diffusion Models via Progressive Seed
  Pruning}.
\newblock In \emph{European Conference on Computer Vision (ECCV)}, 2026.

\bibitem[Guo et~al.(2024)Guo, Liu, Cui, Li, Yang, and Huang]{guo2024initno}
Xiefan Guo, Jinlin Liu, Miaomiao Cui, Jiankai Li, Hongyu Yang, and Di Huang.
\newblock {InitNO: Boosting Text-to-Image Diffusion Models via Initial Noise
  Optimization}.
\newblock In \emph{IEEE/CVF Conference on Computer Vision and Pattern
  Recognition (CVPR)}, pages 9380--9389, 2024.

\bibitem[Hao et~al.(2023)Hao, Chi, Dong, and Wei]{hao2023promptist}
Yaru Hao, Zewen Chi, Li Dong, and Furu Wei.
\newblock {Optimizing Prompts for Text-to-Image Generation}.
\newblock In \emph{Advances in Neural Information Processing Systems
  (NeurIPS)}, pages 66923--66939, 2023.

\bibitem[Hessel et~al.(2021)Hessel, Holtzman, Forbes, Le~Bras, and
  Choi]{hessel2021clipscore}
Jack Hessel, Ari Holtzman, Maxwell Forbes, Ronan Le~Bras, and Yejin Choi.
\newblock {CLIPScore: A Reference-free Evaluation Metric for Image Captioning}.
\newblock In \emph{Conference on Empirical Methods in Natural Language
  Processing (EMNLP)}, 2021.

\bibitem[Ho and Salimans(2021)]{ho2022cfg}
Jonathan Ho and Tim Salimans.
\newblock {Classifier-Free Diffusion Guidance}.
\newblock In \emph{NeurIPS Workshop on Deep Generative Models and Downstream
  Applications}, 2021.

\bibitem[Hoogeboom et~al.(2023)Hoogeboom, Heek, and
  Salimans]{hoogeboom2023simple}
Emiel Hoogeboom, Jonathan Heek, and Tim Salimans.
\newblock {simple diffusion: End-to-end diffusion for high resolution images}.
\newblock In \emph{International Conference on Machine Learning (ICML)}, pages
  13213--13232. PMLR, 2023.

\bibitem[Hu et~al.(2022)Hu, Shen, Wallis, Allen-Zhu, Li, Wang, Wang, and
  Chen]{hu2022lora}
Edward~J. Hu, Yelong Shen, Phillip Wallis, Zeyuan Allen-Zhu, Yuanzhi Li, Shean
  Wang, Lu Wang, and Weizhu Chen.
\newblock {LoRA: Low-Rank Adaptation of Large Language Models}.
\newblock In \emph{International Conference on Learning Representations
  (ICLR)}, 2022.

\bibitem[Huang et~al.(2024)Huang, He, Yu, Zhang, Si, Jiang, Zhang, Wu, Jin,
  Chanpaisit, et~al.]{huang2024vbench}
Ziqi Huang, Yinan He, Jiashuo Yu, Fan Zhang, Chenyang Si, Yuming Jiang, Yuanhan
  Zhang, Tianxing Wu, Qingyang Jin, Nattapol Chanpaisit, et~al.
\newblock {VBench: Comprehensive Benchmark Suite for Video Generative Models}.
\newblock In \emph{IEEE/CVF Conference on Computer Vision and Pattern
  Recognition (CVPR)}, pages 21807--21818, 2024.

\bibitem[Jeong et~al.(2026{\natexlab{a}})Jeong, Lee, Seo, and
  Chun]{jeong2025mixedres}
Wongi Jeong, Kyungryeol Lee, Hoigi Seo, and Se~Young Chun.
\newblock {Training-free Mixed-Resolution Latent Upsampling for Spatially
  Accelerated Diffusion Transformers}.
\newblock In \emph{IEEE/CVF Conference on Computer Vision and Pattern
  Recognition (CVPR)}, pages 18673--18682, 2026{\natexlab{a}}.

\bibitem[Jeong et~al.(2026{\natexlab{b}})Jeong, Seo, and
  Chun]{jeong2026previews}
Wongi Jeong, Hoigi Seo, and Se~Young Chun.
\newblock {Training-free, Perceptually Consistent Low-Resolution Previews with
  High-Resolution Image for Efficient Workflows of Diffusion Models}.
\newblock In \emph{IEEE/CVF Conference on Computer Vision and Pattern
  Recognition (CVPR)}, pages 4547--4557, 2026{\natexlab{b}}.

\bibitem[Jia et~al.(2026)Jia, Liu, and Wang]{jia2026faithfulpreviews}
Jing Jia, Sifan Liu, and Guanyang Wang.
\newblock {Unlocking Few-Step Diffusion for Faithful Previews}.
\newblock arXiv preprint arXiv:2609.34406, 2026.

\bibitem[Karras et~al.(2024)Karras, Aittala, Kynk{\"a}{\"a}nniemi, Lehtinen,
  Aila, and Laine]{karras2024autoguidance}
Tero Karras, Miika Aittala, Tuomas Kynk{\"a}{\"a}nniemi, Jaakko Lehtinen, Timo
  Aila, and Samuli Laine.
\newblock {Guiding a Diffusion Model with a Bad Version of Itself}.
\newblock In \emph{Advances in Neural Information Processing Systems
  (NeurIPS)}, 2024.

\bibitem[Kirstain et~al.(2023)Kirstain, Polyak, Singer, Matiana, Penna, and
  Levy]{kirstain2023pickapic}
Yuval Kirstain, Adam Polyak, Uriel Singer, Shahbuland Matiana, Joe Penna, and
  Omer Levy.
\newblock {Pick-a-Pic: An Open Dataset of User Preferences for Text-to-Image
  Generation}.
\newblock In \emph{Advances in Neural Information Processing Systems
  (NeurIPS)}, pages 36652--36663, 2023.

\bibitem[Kouzelis et~al.(2025)Kouzelis, Kakogeorgiou, Gidaris, and
  Komodakis]{kouzelis2025eqvae}
Theodoros Kouzelis, Ioannis Kakogeorgiou, Spyros Gidaris, and Nikos Komodakis.
\newblock {EQ-VAE: Equivariance Regularized Latent Space for Improved
  Generative Image Modeling}.
\newblock In \emph{International Conference on Machine Learning (ICML)}, 2025.

\bibitem[Kwon et~al.(2026)Kwon, Lim, Nam, Moon, and Oh]{kwon2026earlyfailure}
Byung-Ki Kwon, Sohwi Lim, Hyeon-Woo Nam, Ye-Bin Moon, and Tae-Hyun Oh.
\newblock {Early Failure Detection and Intervention in Video Diffusion Models}.
\newblock In \emph{Advances in Neural Information Processing Systems
  (NeurIPS)}, 2026.

\bibitem[Lin et~al.(2014)Lin, Maire, Belongie, Hays, Perona, Ramanan,
  Doll{\'a}r, and Zitnick]{lin2014coco}
Tsung-Yi Lin, Michael Maire, Serge Belongie, James Hays, Pietro Perona, Deva
  Ramanan, Piotr Doll{\'a}r, and C.~Lawrence Zitnick.
\newblock {Microsoft COCO}: Common objects in context.
\newblock In \emph{European Conference on Computer Vision (ECCV)}, pages
  740--755, 2014.

\bibitem[Lin et~al.(2024)Lin, Pathak, Li, Li, Xia, Neubig, Zhang, and
  Ramanan]{lin2024vqascore}
Zhiqiu Lin, Deepak Pathak, Baiqi Li, Jiayao Li, Xide Xia, Graham Neubig,
  Pengchuan Zhang, and Deva Ramanan.
\newblock {Evaluating Text-to-Visual Generation with Image-to-Text Generation}.
\newblock In \emph{European Conference on Computer Vision (ECCV)}, 2024.

\bibitem[Lindauer et~al.(2022)Lindauer, Eggensperger, Feurer, Biedenkapp, Deng,
  Benjamins, Ruhkopf, Sass, and Hutter]{lindauer2022smac3}
Marius Lindauer, Katharina Eggensperger, Matthias Feurer, Andr\'e Biedenkapp,
  Difan Deng, Carolin Benjamins, Tim Ruhkopf, Ren\'e Sass, and Frank Hutter.
\newblock {SMAC3: A Versatile Bayesian Optimization Package for Hyperparameter
  Optimization}.
\newblock \emph{Journal of Machine Learning Research}, 23\penalty0
  (54):\penalty0 1--9, 2022.

\bibitem[Lipman et~al.(2023)Lipman, Chen, Ben-Hamu, Nickel, and
  Le]{lipman2023flowmatching}
Yaron Lipman, Ricky T.~Q. Chen, Heli Ben-Hamu, Maximilian Nickel, and Matt Le.
\newblock {Flow Matching for Generative Modeling}.
\newblock In \emph{International Conference on Learning Representations
  (ICLR)}, 2023.

\bibitem[Liu et~al.(2025{\natexlab{a}})Liu, Zhang, Wang, Wei, Qiu, Zhao, Zhang,
  Ye, and Wan]{liu2025teacache}
Feng Liu, Shiwei Zhang, Xiaofeng Wang, Yujie Wei, Haonan Qiu, Yuzhong Zhao,
  Yingya Zhang, Qixiang Ye, and Fang Wan.
\newblock {Timestep Embedding Tells: It's Time to Cache for Video Diffusion
  Model}.
\newblock In \emph{IEEE/CVF Conference on Computer Vision and Pattern
  Recognition (CVPR)}, pages 7353--7363, 2025{\natexlab{a}}.

\bibitem[Liu et~al.(2025{\natexlab{b}})Liu, Liu, Liang, Yuan, Liu, Zheng, Wu,
  Wang, Xia, Wang, et~al.]{liu2025videoreward}
Jie Liu, Gongye Liu, Jiajun Liang, Ziyang Yuan, Xiaokun Liu, Mingwu Zheng,
  Xiele Wu, Qiulin Wang, Menghan Xia, Xintao Wang, et~al.
\newblock {Improving Video Generation with Human Feedback}.
\newblock In \emph{Advances in Neural Information Processing Systems
  (NeurIPS)}, 2025{\natexlab{b}}.

\bibitem[Liu et~al.(2025{\natexlab{c}})Liu, Zou, Lyu, Chen, and
  Zhang]{liu2025taylorseer}
Jiacheng Liu, Chang Zou, Yuanhuiyi Lyu, Junjie Chen, and Linfeng Zhang.
\newblock {From Reusing to Forecasting: Accelerating Diffusion Models with
  TaylorSeers}.
\newblock In \emph{IEEE/CVF International Conference on Computer Vision
  (ICCV)}, pages 15853--15863, 2025{\natexlab{c}}.

\bibitem[Liu and Chilton(2022)]{liu2022promptguidelines}
Vivian Liu and Lydia~B. Chilton.
\newblock {Design Guidelines for Prompt Engineering Text-to-Image Generative
  Models}.
\newblock In \emph{Proceedings of the CHI Conference on Human Factors in
  Computing Systems (CHI)}, pages 1--23, 2022.

\bibitem[Liu et~al.(2023)Liu, Gong, and Liu]{liu2023rectifiedflow}
Xingchao Liu, Chengyue Gong, and Qiang Liu.
\newblock {Flow Straight and Fast: Learning to Generate and Transfer Data with
  Rectified Flow}.
\newblock In \emph{International Conference on Learning Representations
  (ICLR)}, 2023.

\bibitem[Luo et~al.(2023)Luo, Tan, Huang, Li, and Zhao]{luo2023lcm}
Simian Luo, Yiqin Tan, Longbo Huang, Jian Li, and Hang Zhao.
\newblock {Latent Consistency Models: Synthesizing High-Resolution Images with
  Few-Step Inference}.
\newblock arXiv preprint arXiv:2310.04378, 2023.

\bibitem[Ma et~al.(2024)Ma, Fang, and Wang]{ma2024deepcache}
Xinyin Ma, Gongfan Fang, and Xinchao Wang.
\newblock {DeepCache: Accelerating Diffusion Models for Free}.
\newblock In \emph{IEEE/CVF Conference on Computer Vision and Pattern
  Recognition (CVPR)}, pages 15762--15772, 2024.

\bibitem[Malnick et~al.(2026)Malnick, Orenstein, Fried, and
  Avidan]{malnick2026softcache}
Shimon Malnick, Nadav Orenstein, Ohad Fried, and Shai Avidan.
\newblock {SoftCache: Adaptive Step Size Caching for Diffusion Transformers}.
\newblock arXiv preprint arXiv:2610.22369, 2026.

\bibitem[Munir et~al.(2026)Munir, Vu, Misra, Jena, Norouzi, Taghibakhshi,
  Ahmad, Patney, Molchanov, and Tajbakhsh]{munir2026track}
Mustafa Munir, Huy Vu, Shreyas Misra, Rohit Jena, Sajad Norouzi, Ali
  Taghibakhshi, Anis Ahmad, Anjul Patney, Pavlo Molchanov, and Nima Tajbakhsh.
\newblock {Accelerating Video Diffusion via Training-Free Trajectory Routing}.
\newblock arXiv preprint arXiv:2609.30096, 2026.

\bibitem[Oppenheim and Schafer(2010)]{oppenheim2010dtsp}
Alan~V. Oppenheim and Ronald~W. Schafer.
\newblock \emph{{Discrete-Time Signal Processing}}.
\newblock Prentice Hall, 3 edition, 2010.

\bibitem[Pan et~al.(2025)Pan, Zhuang, Huang, Nie, Yu, Xiao, Cai, and
  Anandkumar]{pan2025tstitch}
Zizheng Pan, Bohan Zhuang, De-An Huang, Weili Nie, Zhiding Yu, Chaowei Xiao,
  Jianfei Cai, and Anima Anandkumar.
\newblock {T-Stitch: Accelerating Sampling in Pre-Trained Diffusion Models with
  Trajectory Stitching}.
\newblock In \emph{International Conference on Learning Representations
  (ICLR)}, 2025.

\bibitem[Polyak et~al.(2024)Polyak, Zohar, Brown, Tjandra, Sinha, Lee, Vyas,
  Shi, Ma, Chuang, et~al.]{polyak2024moviegen}
Adam Polyak, Amit Zohar, Andrew Brown, Andros Tjandra, Animesh Sinha, Ann Lee,
  Apoorv Vyas, Bowen Shi, Chih-Yao Ma, Ching-Yao Chuang, et~al.
\newblock {Movie Gen: A Cast of Media Foundation Models}.
\newblock arXiv preprint arXiv:2410.13720, 2024.

\bibitem[Rawal et~al.(2026)Rawal, Shirkavand, Paul, Wen, Huang, Chen,
  Goldstein, and Somepalli]{rawal2026flashbon}
Ruchit Rawal, Reza Shirkavand, Sayak Paul, Yuxin Wen, Heng Huang, Yizheng Chen,
  Tom Goldstein, and Gowthami Somepalli.
\newblock {Flash-BoN: Instant Drafts for Inference-Time Scaling in Diffusion
  Models}.
\newblock In \emph{European Conference on Computer Vision (ECCV)}, 2026.

\bibitem[Schuhmann et~al.(2022)Schuhmann, Beaumont, Vencu, Gordon, Wightman,
  Cherti, Coombes, Katta, Mullis, Wortsman, et~al.]{schuhmann2022laion5b}
Christoph Schuhmann, Romain Beaumont, Richard Vencu, Cade Gordon, Ross
  Wightman, Mehdi Cherti, Theo Coombes, Aarush Katta, Clayton Mullis, Mitchell
  Wortsman, et~al.
\newblock {LAION-5B: An open large-scale dataset for training next generation
  image-text models}.
\newblock In \emph{Advances in Neural Information Processing Systems (NeurIPS),
  Datasets and Benchmarks Track}, pages 25278--25294, 2022.

\bibitem[Skorokhodov et~al.(2025)Skorokhodov, Girish, Hu, Menapace, Li, Abdal,
  Tulyakov, and Siarohin]{skorokhodov2025diffusability}
Ivan Skorokhodov, Sharath Girish, Benran Hu, Willi Menapace, Yanyu Li, Rameen
  Abdal, Sergey Tulyakov, and Aliaksandr Siarohin.
\newblock {Improving the Diffusability of Autoencoders}.
\newblock In \emph{International Conference on Machine Learning (ICML)}, 2025.

\bibitem[Song et~al.(2023)Song, Dhariwal, Chen, and
  Sutskever]{song2023consistency}
Yang Song, Prafulla Dhariwal, Mark Chen, and Ilya Sutskever.
\newblock {Consistency Models}.
\newblock In \emph{International Conference on Machine Learning (ICML)}, pages
  32211--32252. PMLR, 2023.

\bibitem[Song et~al.(2025)Song, Liu, and Shou]{song2025diffsim}
Yiren Song, Xiaokang Liu, and Mike~Zheng Shou.
\newblock {DiffSim: Taming Diffusion Models for Evaluating Visual Similarity}.
\newblock In \emph{IEEE/CVF International Conference on Computer Vision
  (ICCV)}, pages 16904--16915, 2025.

\bibitem[Subramaniam et~al.(2026)Subramaniam, Katz, Cheung, Li, Pfister, and
  Song]{subramaniam2026previewdiff}
Vighnesh Subramaniam, Boris Katz, Brian Cheung, Chun-Liang Li, Tomas Pfister,
  and Yale Song.
\newblock {PreviewDiff: Multimodal Critic-Guided Search over Diffusion
  Latents}.
\newblock arXiv preprint arXiv:2609.36199, 2026.

\bibitem[Tian et~al.(2025)Tian, Xia, Ren, Lin, Wang, Xiao, Tong, Yang, and
  Cui]{tian2025bottleneck}
Ye Tian, Xin Xia, Yuxi Ren, Shanchuan Lin, Xing Wang, Xuefeng Xiao, Yunhai
  Tong, Ling Yang, and Bin Cui.
\newblock {Training-free Diffusion Acceleration with Bottleneck Sampling}.
\newblock arXiv preprint arXiv:2503.18940, 2025.

\bibitem[Tong et~al.(2022)Tong, Song, Wang, and Wang]{tong2022videomae}
Zhan Tong, Yibing Song, Jue Wang, and Limin Wang.
\newblock {VideoMAE}: Masked autoencoders are data-efficient learners for
  self-supervised video pre-training.
\newblock In \emph{Advances in Neural Information Processing Systems}, pages
  10078--10093, 2022.

\bibitem[Wan et~al.(2025)Wan, Wang, Ai, Wen, Mao, Xie, Chen, Yu, Zhao, Yang,
  et~al.]{wan2025wan}
Team Wan, Ang Wang, Baole Ai, Bin Wen, Chaojie Mao, Chen-Wei Xie, Di Chen,
  Feiwu Yu, Haiming Zhao, Jianxiao Yang, et~al.
\newblock {Wan: Open and Advanced Large-Scale Video Generative Models}.
\newblock arXiv preprint arXiv:2503.20314, 2025.

\bibitem[Wang and Vastola(2023)]{wang2023painters}
Binxu Wang and John~J. Vastola.
\newblock {Diffusion Models Generate Images Like Painters: an Analytical Theory
  of Outline First, Details Later}.
\newblock arXiv preprint arXiv:2303.02490, 2023.

\bibitem[Wang et~al.(2026)Wang, Zhou, Yuan, Woo, Gong, Han, Yang, Zhang, Zhu,
  Liu, and Zhao]{wang2026consolver}
Fu-Yun Wang, Hao Zhou, Liangzhe Yuan, Sanghyun Woo, Boqing Gong, Bohyung Han,
  Ming-Hsuan Yang, Han Zhang, Yukun Zhu, Ting Liu, and Long Zhao.
\newblock {Image Diffusion Preview with Consistency Solver}.
\newblock In \emph{IEEE/CVF Conference on Computer Vision and Pattern
  Recognition (CVPR)}, pages 43271--43280, 2026.

\bibitem[Wang et~al.(2025)Wang, Wan, Belkin, and Wang]{wang2025seeds}
Qingsong Wang, Zhengchao Wan, Mikhail Belkin, and Yusu Wang.
\newblock {Seeds of Structure: Patch PCA Reveals Universal Compositional Cues
  in Diffusion Models}.
\newblock In \emph{Advances in Neural Information Processing Systems
  (NeurIPS)}, 2025.

\bibitem[Wu et~al.(2023)Wu, Hao, Sun, Chen, Zhu, Zhao, and Li]{wu2023hpsv2}
Xiaoshi Wu, Yiming Hao, Keqiang Sun, Yixiong Chen, Feng Zhu, Rui Zhao, and
  Hongsheng Li.
\newblock {Human Preference Score v2: A Solid Benchmark for Evaluating Human
  Preferences of Text-to-Image Synthesis}.
\newblock arXiv preprint arXiv:2306.09341, 2023.

\bibitem[Xiao et~al.(2026)Xiao, Chao, Yariv, and Wetzstein]{xiao2026speed}
Howard Xiao, Brian Chao, Lior Yariv, and Gordon Wetzstein.
\newblock {Spectral Progressive Diffusion for Efficient Image and Video
  Generation}.
\newblock In \emph{Advances in Neural Information Processing Systems
  (NeurIPS)}, 2026.

\bibitem[Xie et~al.(2023)Xie, Pan, Ma, Jie, and Mei]{xie2023promptlog}
Yutong Xie, Zhaoying Pan, Jinge Ma, Luo Jie, and Qiaozhu Mei.
\newblock {A Prompt Log Analysis of Text-to-Image Generation Systems}.
\newblock In \emph{Proceedings of the ACM Web Conference (WWW)}, pages
  3892--3902, 2023.

\bibitem[Xu et~al.(2023)Xu, Liu, Wu, Tong, Li, Ding, Tang, and
  Dong]{xu2023imagereward}
Jiazheng Xu, Xiao Liu, Yuchen Wu, Yuxuan Tong, Qinkai Li, Ming Ding, Jie Tang,
  and Yuxiao Dong.
\newblock {ImageReward: Learning and Evaluating Human Preferences for
  Text-to-Image Generation}.
\newblock In \emph{Advances in Neural Information Processing Systems
  (NeurIPS)}, pages 15903--15935, 2023.

\bibitem[Xu et~al.(2025)Xu, Zhang, and Shi]{xu2025goodseed}
Katherine Xu, Lingzhi Zhang, and Jianbo Shi.
\newblock {Good Seed Makes a Good Crop: Discovering Secret Seeds in
  Text-to-Image Diffusion Models}.
\newblock In \emph{IEEE/CVF Winter Conference on Applications of Computer
  Vision (WACV)}, pages 3024--3034, 2025.

\bibitem[Yin et~al.(2024)Yin, Gharbi, Zhang, Shechtman, Durand, Freeman, and
  Park]{yin2024dmd}
Tianwei Yin, Micha{\"e}l Gharbi, Richard Zhang, Eli Shechtman, Fr{\'e}do
  Durand, William~T. Freeman, and Taesung Park.
\newblock {One-step Diffusion with Distribution Matching Distillation}.
\newblock In \emph{IEEE/CVF Conference on Computer Vision and Pattern
  Recognition (CVPR)}, pages 6613--6623, 2024.

\bibitem[Z.~Cohen et~al.(2026)Z.~Cohen, Abramovich, and Shamir]{Z_Cohen_2026}
Nadav Z.~Cohen, Ofir Abramovich, and Ariel Shamir.
\newblock {Colorful-Noise: Training-Free Low-Frequency Noise Manipulation for
  Color-Based Conditional Image Generation}.
\newblock In \emph{ACM SIGGRAPH Conference Papers}, pages 1--11, 2026.

\bibitem[Zhang et~al.(2018)Zhang, Isola, Efros, Shechtman, and
  Wang]{zhang2018lpips}
Richard Zhang, Phillip Isola, Alexei~A. Efros, Eli Shechtman, and Oliver Wang.
\newblock {The Unreasonable Effectiveness of Deep Features as a Perceptual
  Metric}.
\newblock In \emph{IEEE/CVF Conference on Computer Vision and Pattern
  Recognition (CVPR)}, pages 586--595, 2018.

\bibitem[Zhao et~al.(2026)Zhao, Liu, Cao, Gong, Zhang, Song, Deng, and
  Patras]{zhao2026latsearch}
Zengqun Zhao, Ziquan Liu, Yu Cao, Shaogang Gong, Zhensong Zhang, Jifei Song,
  Jiankang Deng, and Ioannis Patras.
\newblock {LatSearch: Latent Reward-Guided Search for Faster Inference-Time
  Scaling in Video Diffusion}.
\newblock In \emph{European Conference on Computer Vision (ECCV)}, 2026.

\bibitem[Zheng et~al.(2026)Zheng, Liu, Ding, Feng, Lin, Guo, and
  Qin]{zheng2026mrflow}
Xingyu Zheng, Xianglong Liu, Yifu Ding, Weilun Feng, Junqing Lin, Jinyang Guo,
  and Haotong Qin.
\newblock {Multi-Resolution Flow Matching: Training-Free Diffusion Acceleration
  via Staged Sampling}.
\newblock arXiv preprint arXiv:2607.01642, 2026.

\bibitem[Zhou et~al.(2025)Zhou, Shao, Bai, Zhang, Xu, Han, and
  Xie]{zhou2025goldennoise}
Zikai Zhou, Shitong Shao, Lichen Bai, Shufei Zhang, Zhiqiang Xu, Bo Han, and
  Zeke Xie.
\newblock {Golden Noise for Diffusion Models: A Learning Framework}.
\newblock In \emph{IEEE/CVF International Conference on Computer Vision
  (ICCV)}, pages 17688--17697, 2025.

\end{thebibliography}
